\documentclass[letterpaper, 10 pt, journal, twoside]{IEEEtran}
\IEEEoverridecommandlockouts
\usepackage{amsmath,amsfonts,amssymb}
\usepackage{algorithmic}
\usepackage{algorithm}
\usepackage{array}
\usepackage[caption=false,font=normalsize,labelfont=sf,textfont=sf]{subfig}
\usepackage{textcomp}
\usepackage{stfloats}
\usepackage{url}
\usepackage{verbatim}
\usepackage{graphicx}
\usepackage{cite}

\usepackage[linkcolor=red,citecolor=red,urlcolor=blue,colorlinks=true]{hyperref}
\usepackage{caption}
\usepackage{hypcap}
\usepackage{mathtools}
\usepackage{bm}
\usepackage{color}
\usepackage[table]{xcolor}
\usepackage{algorithm}
\usepackage{algorithmic}
\usepackage{tabulary}
\usepackage{booktabs}
\usepackage{multirow}
\usepackage{multicol}
\usepackage{gensymb}
\usepackage{float}
\usepackage{pifont}
\usepackage[normalem]{ulem}

\newcommand{\fullwidth}{\textwidth}
\newcommand{\halfwidth}{\columnwidth}

\newcommand{\CREPES}{CREPES-X}

\newcommand{\PAPERTITLEB}{\CREPES{}: Hierarchical Bearing-Distance-Inertial \\ Direct Cooperative Relative Pose Estimation System}

\newcommand{\goodmark}{\textcolor{green}{\ding{51}}}
\newcommand{\badmark}{\textcolor{red}{\ding{55}}}

\allowdisplaybreaks

\begin{document}

\title{\PAPERTITLEB{}}

\author{
    Zhehan Li\textsuperscript{1,2},
    Jiadong Lu\textsuperscript{$\dagger$,1,2},
    Zheng Wang\textsuperscript{$\dagger$,1,2,3},
    Qi Liu\textsuperscript{2,4},
    Zhiren Xun\textsuperscript{1,2},
    \\
    Yue Wang\textsuperscript{1},
    Fei Gao\textsuperscript{1},
    Chao Xu\textsuperscript{1,2},
    and
    Yanjun Cao\textsuperscript{1,2}
    \thanks{
        \textsuperscript{$\dagger$} \textbf{Equal contribution.}
    }
    \thanks{
        \textsuperscript{1} State Key Laboratory of Industrial Control Technology,
        Institute of Cyber-Systems and Control,
        Zhejiang University,
        Hangzhou, China.
    }
    \thanks{
        \textsuperscript{2} Huzhou Institute of Zhejiang University,
        Huzhou, China.
    }
    \thanks{
        \textsuperscript{3} University of Michigan,
        Ann Arbor, MI, USA.
    }
    \thanks{
        \textsuperscript{4} The Hong Kong Polytechnic University,
        Hong Kong, China.
    }
    \thanks{
        This work was supported by National Nature Science Foundation of China under Grant 62103368.
        The corresponding author is Yanjun Cao.
    }
    \thanks{
        E-mails: \texttt{zhehanli@zju.edu.cn}, \texttt{yanjuncao@zju.edu.cn}.
    }
}

\maketitle

\begin{abstract}

    Relative localization is essential for cooperation in autonomous multi-robot systems.
    Existing approaches either rely on shared environmental features or inertial assumptions, or they degrade under pairwise non-line-of-sight conditions and outliers in complex environments.
    Robustly and efficiently fusing inter-robot bearings, distances, and inertial measurements for tens of robots remains challenging.
    We present \CREPES{} (Cooperative RElative Pose Estimation System with multiple eXtended features), a hierarchical relative localization framework that enhances speed, accuracy, and robustness under challenging conditions, without requiring any global information.
    The hardware packs infrared (IR) LEDs, an IR camera, an ultra-wideband module, and an IMU into a cube no larger than $6\,\text{cm}$ on each side.
    On this hardware, a two-stage hierarchical estimator meets different latency, accuracy, and robustness requirements.
    The single-frame estimator returns instantaneous relative poses from a closed-form solution with bearing outlier rejection.
    The multi-frame estimator then refines these poses with IMU pre-integration under robocentric relative kinematics, using loosely- and tightly-coupled optimization.
    Extensive simulations and real-world experiments validate the effectiveness of \CREPES{}, demonstrating robustness of up to $90\%$ bearing outliers, resilience in challenging conditions, and RMSE of $7.0\,\text{cm}$ and $2.2\degree$ in real-world datasets.

\end{abstract}


\section{Introduction}
\label{Sec: Introduction}

\IEEEPARstart{M}{ulti-robot} systems are increasingly deployed in fields such as search and rescue \cite{waharte2010supporting,tian2020search} and environmental exploration \cite{zhou2023racer}.
These applications rely on relative localization for cooperative tasks, including formation control \cite{guo2019ultra,quan2023robust}, target tracking \cite{riazuelo2014c2tam,bonatti2020autonomous}, cooperative navigation \cite{mueggler2014aerial,chen2024cost}, and environmental mapping \cite{xu2022d,tian2022kimera,zhu2024swarm}.
Faster, more accurate, and more robust localization leads to better collaboration.

\begin{figure}[t]
    \centering
    \includegraphics[width=\halfwidth]{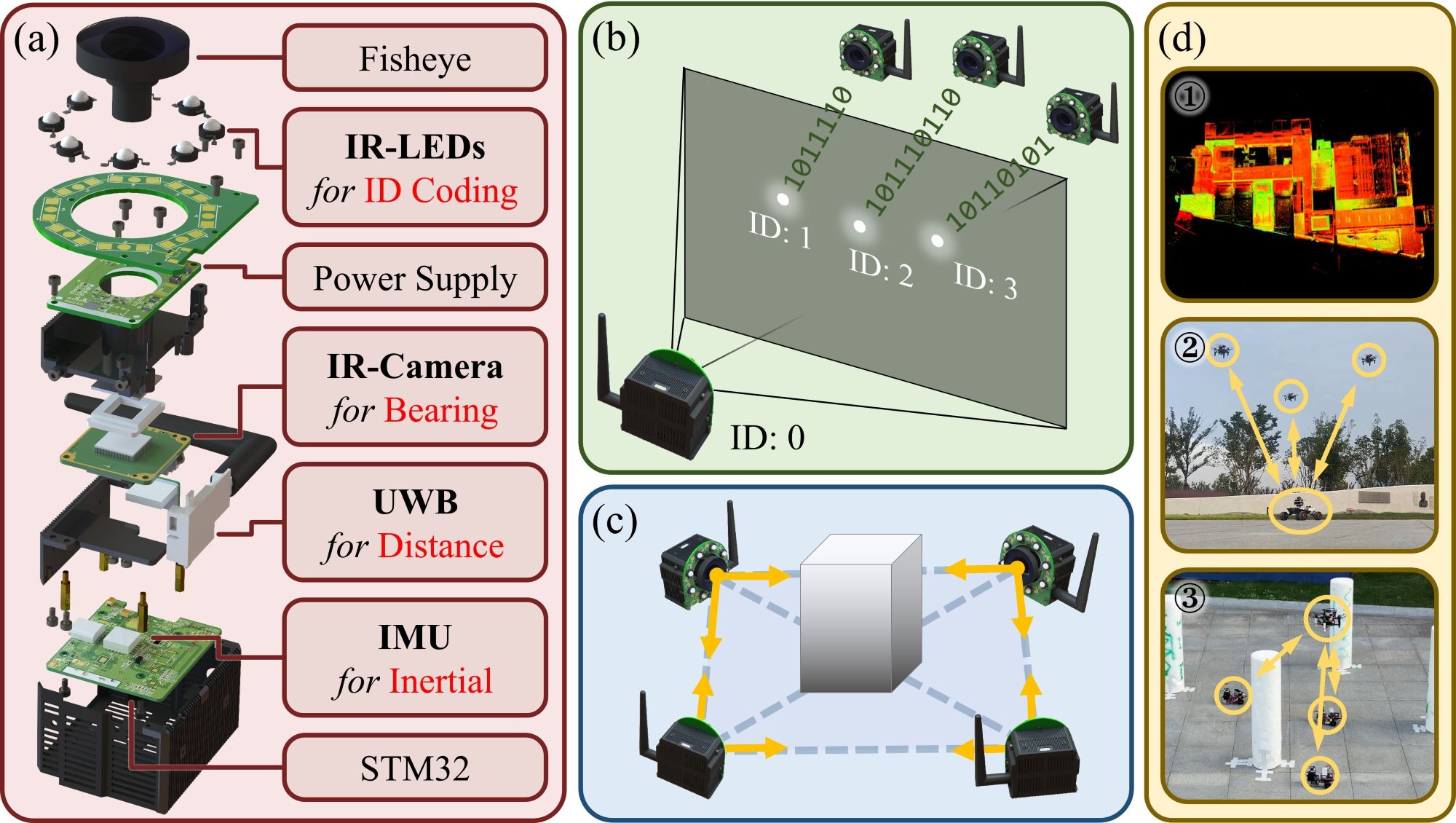}
    \caption{
        \CREPES{} works in a robocentric frame, independent of the environment, and provides accurate and robust relative state estimation in real-time.
        (a) The compact hardware design of \CREPES{}.
        (b) IR LEDs and an IR camera work as light-coded communication, providing bearings with ID.
        (c) Multiple \CREPES{} can overcome pairwise non-line-of-sight conditions via indirect observations.
        (d) \CREPES{} can be used in \ding{172} merging point clouds map from neighbors, \ding{173} direct controlling quadrotors in a vehicle's frame \cite{zhang2023coni}, and \ding{174} continuous sharing information between robots for navigation \cite{chen2024cost}.
    }
    \label{Fig: Top Figure 1}
\end{figure}

\begin{figure}[t]
    \centering
    \includegraphics[width=\halfwidth]{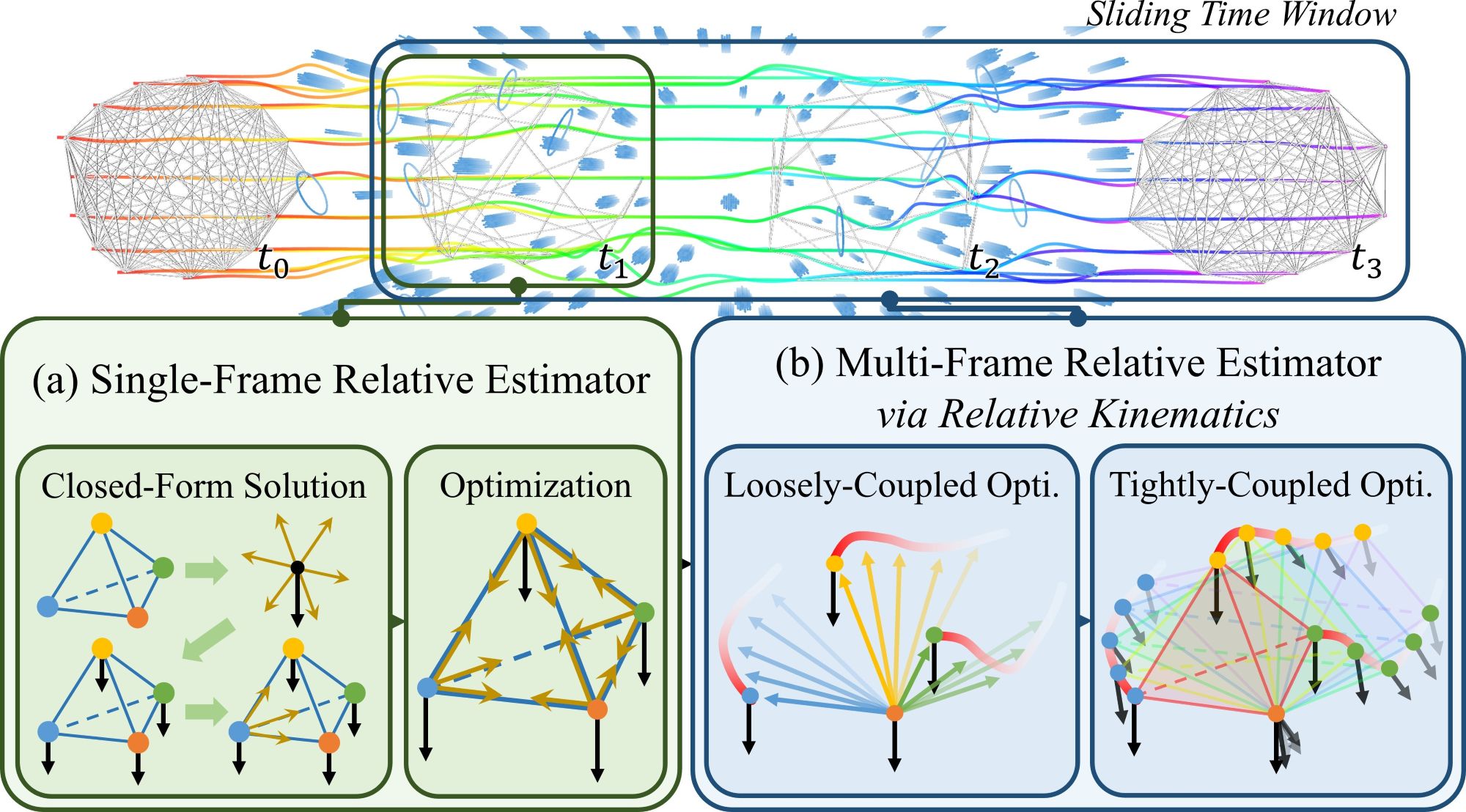}
    \caption{
        \CREPES{} estimates relative states using distance, bearing, inertial, and optionally gravity measurements.
        The rainbow-colored trajectories in the top panel visualize the multi-robot formation, with the color changing from earlier to later timestamps.
        The proposed two-stage hierarchical estimator is designed to satisfy different accuracy and latency requirements:
        (a) The single-frame estimator delivers instantaneous relative poses through multi-stage closed-form solutions followed by optimization.
        (b) The multi-frame estimator refines relative states over a time window using loosely- and tightly-coupled optimizations for improved accuracy and robustness.
    }
    \label{Fig: Top Figure 2}
\end{figure}

A common approach uses each robot's odometry in a global reference frame \cite{ledergerber2015robot,nguyen2016ultra,preiss2017crazyswarm}, so the relative state follows from the difference between the agents' global states.
However, these systems rely on pre-installed infrastructure or require time-consuming calibration, limiting their applicability in unknown environments.
Simultaneous Localization And Mapping (SLAM) provides each robot's odometry in its own frame \cite{qin2018vins,xu2021fast,rosinol2020kimera}, and relative transformations between robots can be estimated by matching common features in their maps \cite{xu2022d,tian2022kimera,zhu2024swarm} or aligning odometries with inner-robot observations \cite{montesano2005cooperative,stegagno2013relative,wang2022certifiably,wang2023bearing,wang2024certifiable,trawny20073d,zhou2008robot,guo2017ultra,nguyen2021flexible,jiang20193,li2020robot,nguyen2023relative,franchi2013mutual,zhou2012determining,xiong2024onboard,zhang2022agile,xu2022omni,chen2025relative}.
However, these methods need high computational resources and communication bandwidth and may fail to provide relative poses in featureless or non-inertial environments, where the global acceleration is unavailable or unreliable.
By equipping robots with specially designed structures, such as AprilTags \cite{olson2011apriltag} and LEDs \cite{faessler2014monocular,yan2019active,walter2018fast,cutler2013lightweight,dias2016board}, the relative pose can be estimated from direct inter-robot observations.
However, these systems are constrained by short detection ranges, strict viewpoint requirements, and sensitivity to ambient light, limiting their effectiveness in multi-robot applications.

\newcommand{\CHALLENGES}{\textit{ON\textsuperscript{3}} challenges}
Based on the above challenges, we conclude that a desired relative localization system that is applicable to multi-robot systems needs to be able to operate in a variety of environments, which we have summarized as the \CHALLENGES{}:
\begin{enumerate}
    \item \textbf{O}utlier-Existence (OE): The system should be robust to large numbers of outliers in the measurements.
    \item \textbf{N}on-Global-Information (NGI): The system should not rely on pre-installed infrastructures (e.g., Global Positioning System (GPS), Motion Capture System (MCS)).
    \item \textbf{N}on-Line-Of-Sight (NLOS): The system should be able to work in occluded scenarios through indirect observations, not limited by fully-connected direct observation.
    \item \textbf{N}on-Inertial (NI): The system should not rely on the inertial reference frame and can operate in NI scenarios.
\end{enumerate}

To overcome the \CHALLENGES{}, we introduce \CREPES{}, a fully improved cooperative relative pose estimation system with a hierarchical structure to produce multi-layered, accurate, robust relative states within inter-robot mutual observations, shown in \autoref{Fig: Top Figure 1}.
\CREPES{} integrates compact hardware with a complete software stack.
The hardware consists of four components: an Inertial Measurement Unit (IMU), an Ultra-WideBand (UWB) module, a set of InfraRed (IR) LEDs, and an IR camera.
This unique configuration provides complementary measurements, namely bearings and IDs from the LED-camera system, distances from the UWB, and inertial with gravity information from the IMU.
The measurements are shared among robots via Wi-Fi.
Built upon these inputs, the software implements a hierarchical estimator comprising (i) a single-frame relative estimator that rejects outliers by consistency and yields fast relative poses in closed-form, and (ii) a multi-frame relative estimator leveraging robocentric relative kinematics with IMU pre-integration to achieve accurate and robust relative states, as shown in \autoref{Fig: Top Figure 2}.

Compared with existing methods, \CREPES{} eliminates dependence on SLAM, ensuring drift-free estimation independent of environmental features.
By exploiting inter-robot observations, it remains effective under pairwise NLOS conditions, as long as the LOS visibility graph among agents remains connected.
With consensus-based validation, it robustly rejects bearing outliers in multi-robot settings.
Furthermore, by formulating estimation directly in the robocentric frame, \CREPES{} avoids global inertial assumptions, making it suitable for NI scenarios.

In summary, our main contributions are:
\begin{enumerate}
    \item We propose \CREPES{}, a complete and self-contained relative localization system addressing the \CHALLENGES{} via a compact hardware design and a novel hierarchical estimation framework.
    \item We derive a closed-form single-frame relative estimator that rejects bearing outliers and exploits inter-robot observations to maximize pose observability, which is scalable, computationally efficient, and provides instant relative poses with an optimization refinement.
    \item We exploit IMU pre-integration via robocentric relative kinematics over multiple frames and propose loosely- and tightly-coupled optimization methods that can provide accurate and robust relative states in challenging conditions such as NLOS or high dynamic motions.
    \item We validate \CREPES{} in extensive simulations and real-world experiments, and will release the source code and datasets for the benefit of the community (\url{https://github.com/FAST-FIRE/CREPES-X}).
\end{enumerate}

This manuscript significantly extends the work of Xun et al. \cite{xun2023crepes}, which met only NGI and NI of the \CHALLENGES{} using similar hardware.
Their two-robot system relied on an Error-State Kalman Filter (ESKF) and one-shot Pose Graph Optimization (PGO) but was prone to drift under OE and NLOS conditions.
These issues stem from decomposing the system into dual-robot subsystems, which need both to be visible to each other, thus limiting robustness and failing to fully utilize observations.
In contrast, we design both single-frame and multi-frame relative estimators entirely based on multi-robot scenarios while considering NLOS cases, enabling robust and accurate state estimation under \CHALLENGES{}.

\section{Related Works}
\label{Sec: Related Works}

Here, we review the state-of-the-art in multi-robot relative localization systems, focusing on relative state estimation and outlier rejection.
We exclude infrastructure-based localization systems (e.g., MCS and UWB with anchors) here, as they need pre-installed infrastructure in the environment, which is not suitable for autonomous multi-robot systems.
Existing relative localization systems can be broadly divided into indirect and direct methods, depending on whether the relative pose can be estimated instantaneously.
We summarize a selection of representative relative localization algorithms in \autoref{Tab: Relative Localization Algorithms}.
Motion measurements, including self-displacement, global pose, or velocity, are all categorized as odometry.

\begin{table}[ht]
    \centering
    \caption{Relative Localization Algorithms}
    \label{Tab: Relative Localization Algorithms}
    \resizebox{\halfwidth}{!}{
        \begin{tabular}{llllccc}
            \toprule
            Algorithm                             &
            Type                                  &
            Mutual Meas.                          &
            Motion Meas.                          &
            NGI                                   &
            NI                                    &
            NLOS                                    \\
            \midrule
            \cite{eriksson2016consensus,zhang2017distributed,briales2018certifiably}
                                                  &
            Indirect                              &
            \multicolumn{2}{c}{Bundle Adjustment} &
            \badmark                              &
            \goodmark                             &
            \goodmark                               \\
            \cite{xu2022d,tian2022kimera,zhu2024swarm}
                                                  &
            Indirect                              &
            \multicolumn{2}{c}{Cooperative SLAM}  &
            \badmark                              &
            \badmark                              &
            \goodmark                               \\
            \midrule
            \cite{montesano2005cooperative,stegagno2013relative,wang2022certifiably,wang2023bearing,wang2024certifiable}
                                                  &
            Indirect                              &
            Bearing                               &
            Odometry                              &
            \badmark                              &
            \badmark                              &
            \goodmark                               \\
            \cite{trawny20073d,zhou2008robot,guo2017ultra,nguyen2021flexible,jiang20193,li2020robot,nguyen2023relative}
                                                  &
            Indirect                              &
            Distance                              &
            Odometry                              &
            \badmark                              &
            \badmark                              &
            \goodmark                               \\
            \cite{franchi2013mutual}
                                                  &
            Indirect                              &
            Position                              &
            Odometry                              &
            \badmark                              &
            \badmark                              &
            \goodmark                               \\
            \cite{zhou2012determining,xiong2024onboard}
                                                  &
            Indirect                              &
            Bearing, Distance                     &
            Odometry                              &
            \badmark                              &
            \badmark                              &
            \goodmark                               \\
            \cite{zhang2022agile}
                                                  &
            Indirect                              &
            Distance, Position                    &
            Odometry                              &
            \badmark                              &
            \badmark                              &
            \goodmark                               \\
            \cite{xu2022omni}
                                                  &
            Indirect                              &
            Distance, Pose                        &
            Odometry                              &
            \badmark                              &
            \badmark                              &
            \goodmark                               \\
            \cite{chen2025relative}
                                                  &
            Indirect                              &
            Angle                                 &
            Odometry                              &
            \badmark                              &
            \badmark                              &
            \goodmark                               \\
            \midrule
            \cite{cutler2013lightweight,faessler2014monocular,dias2016board,walter2018fast,yan2019active}
                                                  &
            Direct                                &
            Bearing                               &
            -                                     &
            \goodmark                             &
            \goodmark                             &
            \badmark                                \\
            \cite{fishberg2022multi,cossette2022optimal,wu2023distributed,fishberg2024murp,wu2024scalable}
                                                  &
            Direct                                &
            Distance                              &
            -                                     &
            \goodmark                             &
            \goodmark                             &
            \badmark                                \\
            \cite{pugh2006relative,de2012development,mao2013relative}
                                                  &
            Direct                                &
            Bearing, Distance                     &
            -                                     &
            \goodmark                             &
            \goodmark                             &
            \badmark                                \\
            \midrule
            \cite{armani2024accurately}
                                                  &
            Direct                                &
            Distance                              &
            Inertial                              &
            \goodmark                             &
            \badmark                              &
            \goodmark                               \\
            \cite{stegagno2016ground}
                                                  &
            Direct                                &
            Bearing                               &
            Inertial                              &
            \goodmark                             &
            \goodmark                             &
            \badmark                                \\
            \cite{shalaby2024multi}
                                                  &
            Direct                                &
            Distance                              &
            Inertial                              &
            \goodmark                             &
            \goodmark                             &
            \badmark                                \\
            \cite{lai2024dual,aghili2016robust}
                                                  &
            Direct                                &
            Position, Rotation                    &
            Inertial                              &
            \goodmark                             &
            \goodmark                             &
            \badmark                                \\
            Xun's \cite{xun2023crepes}
                                                  &
            Direct                                &
            Bearing, Distance                     &
            Inertial                              &
            \goodmark                             &
            \goodmark                             &
            \badmark                                \\
            \midrule
            Proposed
                                                  &
            Direct                                &
            Bearing, Distance                     &
            Inertial                              &
            \goodmark                             &
            \goodmark                             &
            \goodmark                               \\
            \bottomrule
        \end{tabular}
    }
\end{table}

\subsection{Indirect Relative State Estimation}

Indirect methods estimate relative poses by aligning each robot's local frame to a common reference frame, typically defined by one of the robots or an initial global world frame.
Bundle adjustment \cite{eriksson2016consensus,zhang2017distributed,briales2018certifiably} achieves this through global optimization across multiple views.
Cooperative SLAM \cite{xu2022d,tian2022kimera,lajoie2023swarm,zhu2024swarm} matches shared features and performs PGO in a common frame, relying on inter-robot loop closures derived from bundle adjustment or point cloud registration.
However, these approaches require continuous exchange of feature descriptors, and their accuracy degrades when descriptors are generated from significantly different perspectives.

Mutual observations, such as relative bearing or distance, are applied to help reduce the high dependency on the environment and inter-loop detection.
Relative bearings can be obtained by detecting markers \cite{olson2011apriltag} or using neural networks \cite{xu2022omni}, enabling their use in relative localization.
With bearings, the odometry of each robot can be transformed into a common frame, such as robocentric fusion for pairwise localization \cite{montesano2005cooperative}, using probability hypothesis density filter for multi-target tracking to recover UGVs' positions using UAV \cite{stegagno2013relative}, and optimization-based formulations that leverage bearing and trajectory data to achieve certifiably correct mutual localization through convex relaxation \cite{wang2022certifiably,wang2023bearing,wang2024certifiable}.
Distance measurements, especially via UWB, are popular in relative localization for their light weight and ease of use.
Early work by \cite{trawny20073d} algebraically estimated 6-DoF poses using ten ranges, while \cite{zhou2008robot} proved five distance constraints are sufficient for 3-DoF, and \cite{guo2017ultra} built a UWB-based 3-DoF system for UAV formation.
Later works fused distances with odometries using sliding window optimization \cite{nguyen2021flexible}, or solved the nonconvex optimization problem via semidefinite programming for global guarantees \cite{jiang20193,li2020robot,nguyen2023relative}.
The use of multiple sensor types for relative localization has also been extensively explored.
\cite{zhou2012determining} analyzed 14 minimal bearing-range configurations with closed-form 6-DoF solutions.
\cite{xiong2024onboard} combined vision-based drone detection, odometry, and UWB via distributed graph optimization and delay-aware filtering.
\cite{zhang2022agile} and \cite{xu2022omni} fused visual, inertial, and UWB data for pose refinement using optimization-based frameworks.
\cite{franchi2013mutual} addressed anonymous position measurements with probabilistic registration and particle filtering.
\cite{chen2025relative} unified bearing and distance into angle measurements for distributed 2D localization with theoretical analysis of localization and localizability.

While the use of odometry grants these methods NLOS resistance, their reliance on SLAM systems introduces drawbacks.
They may suffer from degeneration in featureless or NI environments, and are susceptible to drift and error accumulation over time, particularly in large-scale environments.

\subsection{Direct Relative State Estimation}

Although direct methods typically need customized hardware, the self-sufficiency, stability, efficiency, and accuracy have attracted considerable attention.
For bearing-only systems, IR-based marker designs paired with PnP algorithms enable pose estimation at short range \cite{cutler2013lightweight,faessler2014monocular}, with improvements like active markers \cite{dias2016board} or active LEDs coded board \cite{yan2019active} for multi-target identification and Ultraviolet LEDs \cite{walter2018fast} to extend range.
Since the utilization of the matching algorithm, these methods usually work at a short distance to keep the LED structures distinguishable in the image.
Distance-based systems using UWB have demonstrated accurate 2D and 3D pose estimation through multi-module configurations on one robot with least squares optimization.
\cite{fishberg2022multi,fishberg2024murp} used multiple UWB modules respectively, along with IMU data, to estimate 3-DoF poses.
\cite{cossette2022optimal} computed optimal formations for improving relative pose estimation with two UWB modules.
\cite{wu2023distributed,wu2024scalable} formulated the problem as a generalized graph realization and solved the multiconvex optimization with block coordinate descent for scalability.
These systems often require multiple UWB modules with large baselines to achieve good performance, which limits their application to small robots.

Various methods have been proposed to obtain both bearing and distance information, such as using IR receivers \cite{mao2013relative}, multiple IR receivers \cite{pugh2006relative}, or a combination of camera and acoustic sensors \cite{de2012development}.
These approaches often employ Kalman filtering for data fusion \cite{de2012development}.

To improve the accuracy and smoothness of the direct method, inertial measurements were introduced.
For bearing-based systems, \cite{stegagno2016ground} fused the anonymous bearing with inertial data via a particle filter and recovered the scale.
For distance-based systems, \cite{shalaby2024multi} proposed passive UWB ranging with IMU preintegration on the $SE_2(3)$ manifold within an EKF, which can be viewed as a particular instance of the more general Galilean-group framework of \cite{mahony2025galilean}; however, both approaches remain filter-based rather than optimization-based.
Meanwhile, \cite{armani2024accurately} fused UWB distances with inertial and magnetometer data to estimate absolute orientation.
Pose-level fusion has also been explored: \cite{aghili2016robust} used ICP and an adaptive Kalman filter with IMUs in spacecraft, \cite{xun2023crepes} used the directly solved pose as input to the ESKF update step, and \cite{lai2024dual} focused on dual IMUs' bias observability.
Without absolute orientation (e.g., from a magnetometer), \cite{aghili2016robust,lai2024dual,xun2023crepes} chose to derive the relative kinematics model in a robocentric way.
However, these methods are limited to two-robot systems, facing challenges in multi-IMU fusion as the robot count increases.
In contrast, \cite{stegagno2016ground,shalaby2024multi} achieve multi-robot fusion within the filter update step.
As the first optimization-based framework, \CREPES{} naturally accommodates multi-robot IMU fusion under relative kinematic constraints.

\subsection{Outlier Rejection and Robust Estimation}

Outliers are common in multi-robot systems due to NLOS conditions, communication errors, and sensor noise.
Robust estimation requires effective outlier rejection, typically categorized as consensus maximization or M-estimation.
Consensus maximization methods remove outliers before optimization.
RANSAC \cite{fischler1981random} iteratively samples minimal subsets to estimate models and counts inliers.
ADAPT \cite{tzoumas2019outlier} improves upon RANSAC with adaptive trimming and iteration guarantees.
Pairwise Consistent Maximization (PCM) \cite{mangelson2018pairwise} selects consistent loop closures via maximum cliques in a pairwise consistency graph, solvable efficiently with heuristics \cite{pattabiraman2015fast}.
M-estimation methods employ robust loss functions \cite{huber1992robust,holland1977robust,tukey1960survey,geman1985bayesian}, but cannot fully suppress outliers due to smoothness requirements.
Truncated Least Squares (TLS) \cite{enqvist2012robust} completely discards high-residual outliers, though non-convexity can cause local minima.
Graduated Non-Convexity (GNC) \cite{yang2020graduated} addresses this by enabling TLS-based optimization with convergence guarantees.
Yang et al. further developed certifiable methods for robust Wahba’s problem \cite{yang2019quaternion} and point cloud registration \cite{yang2020teaser} using TLS and convex relaxation.
Motivated by the state-free robustness of consensus maximization, we develop a novel adaptation of PCM specifically for bearing measurements, which forms our outlier rejection pipeline.

\section{System Overview}
\label{Sec: System Overview}

\begin{figure*}[t]
    \centering
    \includegraphics[width=\fullwidth]{Figures/System_Overview.jpg}
    \caption{
        System architecture of \CREPES{}.
        Time synchronization between different robots is provided by UWB.
        The IR-camera and IR-LEDs are alternately triggered by the synchronized clock.
        The camera captures images for ID extraction to obtain bearing-ID pairs, which, along with distance, inertial, and gravity data, are broadcast to neighbors.
        Received data are used in decentralized estimation.
        The Single-Frame Relative Estimator (SFRE) computes relative poses from a single time frame.
        It first applies the Single-Frame Closed-form solver (SFC) with outlier rejection, and then refines the solution using Single-Frame Optimization (SFO).
        The Multi-Frame Relative Estimator (MFRE) extends SFRE by fusing temporal information within a sliding window.
        It performs Multi-Frame Loosely-coupled Optimization (MFLO) to generate robust initial guesses, followed by Multi-Frame Tightly-coupled Optimization (MFTO).
        The dashed lines between MFLO and MFTO indicate that MFTO accesses the optimized keyframe estimates from MFLO only during initialization.
    }
    \label{Fig: System Overview}
\end{figure*}

The overview of our system in \autoref{Fig: System Overview} contains five sections: Hardware Design and Implementation (\autoref{Sec: Hardware Design and Implementation}), Problem Statement and Measurement Model (\autoref{Sec: Problem Statement and Measurement Model}), Single-Frame Relative Estimator (SFRE, \autoref{Sec: Single-Frame Relative Estimator}), Multi-Frame Relative Estimator (MFRE, \autoref{Sec: Multi-Frame Relative Estimator}), and Outlier Rejection (\autoref{Sec: Outlier Rejection}).
Our system is designed to satisfy a range of application requirements by providing four distinct outputs with different trade-offs in latency, accuracy, and robustness.

The hardware (\autoref{Sec: Hardware}) includes a set of IR LEDs, an IR camera, a UWB, and an IMU.
The implementation integrates several modules: Time Synchronization, Swarm Bridge, ID Extraction (\autoref{Sec: ID Extraction}).
The Time Synchronization module leverages UWB to align the clocks of all robots to a globally consistent time base.
The Swarm Bridge module broadcasts bearing, distance, inertial, and gravity measurements via Wi-Fi to neighboring robots.
The ID Extraction module (\autoref{Sec: ID Extraction}) processes IR camera images to retrieve bearing-ID pairs, which serve as inputs to the estimators.

Robots in a team can mount the proposed module, and then all the robots can acquire relative localization of neighbors instantly.
Notably, \CREPES{} operates in a relative frame, requiring the definition of a reference frame for estimation.
The decentralized framework supports arbitrary reference selection; in practice, each robot selects itself as the reference and estimates others' relative states in a robocentric manner.

The SFRE (\autoref{Sec: Single-Frame Relative Estimator}) estimates relative poses using only measurements from a single time frame, making it inherently drift-free.
It begins with a Single-Frame Closed-form solver (SFC), which computes relative poses using bearing and distance measurements, along with a bearing outlier rejection step based on measurement consistency.
It contains five stages: position estimation, outlier rejection, position chirality determination, rotation estimation, and relative pose extraction.
This process is designed to utilize measurements from all robots through a closed-form solution, thereby producing instant estimation within several milliseconds.
Then, a Single-Frame Optimization (SFO) refines the poses via bearing-distance-(gravity) optimization.
While SFRE provides instantaneous estimates, its reliance on single-frame observations makes it susceptible to degradation under severe NLOS conditions.

The MFRE (\autoref{Sec: Multi-Frame Relative Estimator}) extends SFRE by fusing IMU measurements within a sliding window under robocentric relative kinematics, a formulation that critically eliminates dependence on a global inertial frame and enables robust operation in non-inertial environments.
In the first stage, Multi-Frame Loosely-coupled Optimization (MFLO) estimates all relative states in the window using SFRE estimates, refined through a position-rotation-inertial-(gravity) optimization, which provides robust initial guesses for the final optimization.
In the second stage, Multi-Frame Tightly-coupled Optimization (MFTO) fuses all measurements through a bearing-distance-inertial-(gravity) optimization.
This holistic approach enhances robustness and accuracy, especially in challenging NLOS environments.

The Outlier Rejection (\autoref{Sec: Outlier Rejection}) focuses on identifying and removing outliers in bearing measurements before they are fed into the estimators.
It employs a modified Pairwise Consistent Maximization (PCM) algorithm, which effectively detects outliers without requiring state information, ensuring robust bearing data for subsequent estimation processes.

\section{Hardware Design and Implementation}
\label{Sec: Hardware Design and Implementation}

\subsection{Hardware}
\label{Sec: Hardware}

\begin{figure}[ht]
    \centering
    \includegraphics[width=\halfwidth]{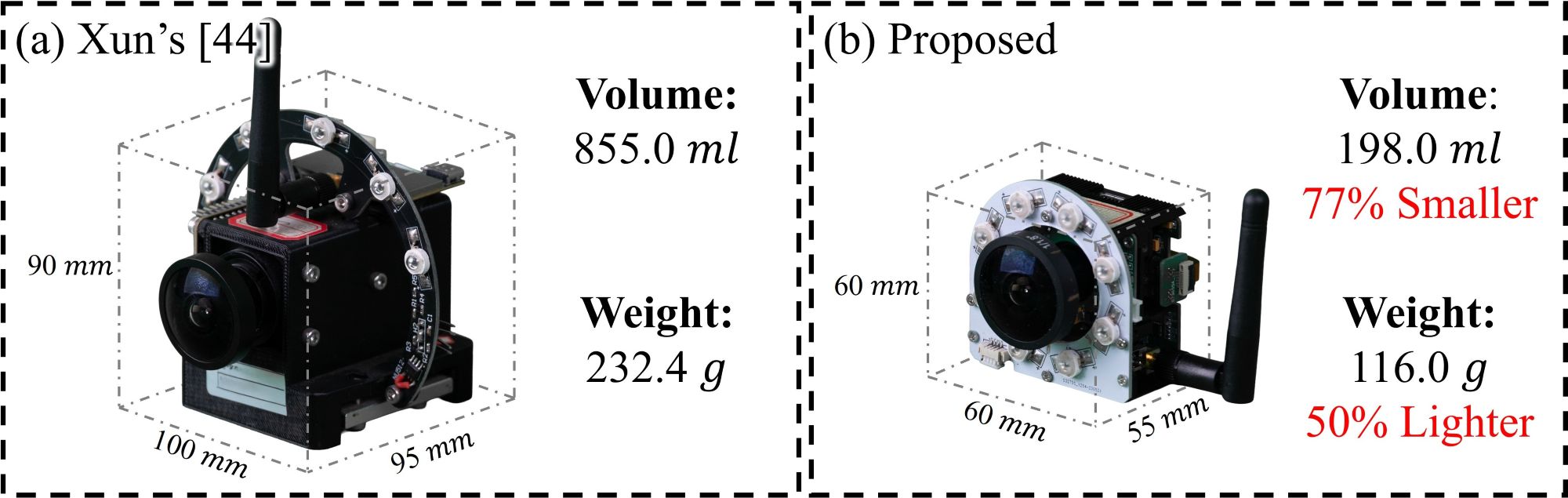}
    \caption{
        Comparison of our integrated module with Xun's \cite{xun2023crepes}.
        Both integrate the same modalities (camera, UWB, IMU, and LED markers); ours reduces volume by $77\,\%$ and weight by $50\,\%$, making it more suitable for smaller, weight-constrained platforms.
    }
    \label{Fig: Hardware Comparison}
\end{figure}

\autoref{Fig: Top Figure 1} (a) shows an overview of our hardware design, which integrates four complementary sensing modalities to achieve robust performance.
A circular board with eight $950\,\text{nm}$ IR LEDs is designed for ID encoding and bearing measurements.
An STM32 is used to receive signals from the computer and control the flickering patterns of the LEDs for encoding.
The IR camera, featuring a fisheye lens with a $185\degree$ field of view (FOV), is based on the MindVision MV-SUA133GM, which integrates a global shutter and a $950\,\text{nm}$ IR-pass filter.
The camera is triggered by the computer to ensure temporal alignment.
The LED-camera system can operate at a maximum of $245\,\text{Hz}$, and is set to $50\,\text{Hz}$ in practice for stability.
A DW1000-based UWB module from NoopLoop is used to provide mutual ranging, with a maximum range of $500\,\text{m}$.
A 6-DoF low-cost MEMS IMU module is used to provide linear accelerations and angular velocities at $100-200\,\text{Hz}$.

A key contribution of our work is the highly integrated and compact hardware.
Compared to the direct combination of off-the-shelf sensors in \cite{xun2023crepes}, our custom-designed solution is 77\% smaller and 50\% lighter (\autoref{Fig: Hardware Comparison}), expanding its applicability to smaller, weight-constrained robotic platforms.

\subsection{ID Extraction}
\label{Sec: ID Extraction}

\begin{figure}[ht]
    \centering
    \includegraphics[width=.75\halfwidth]{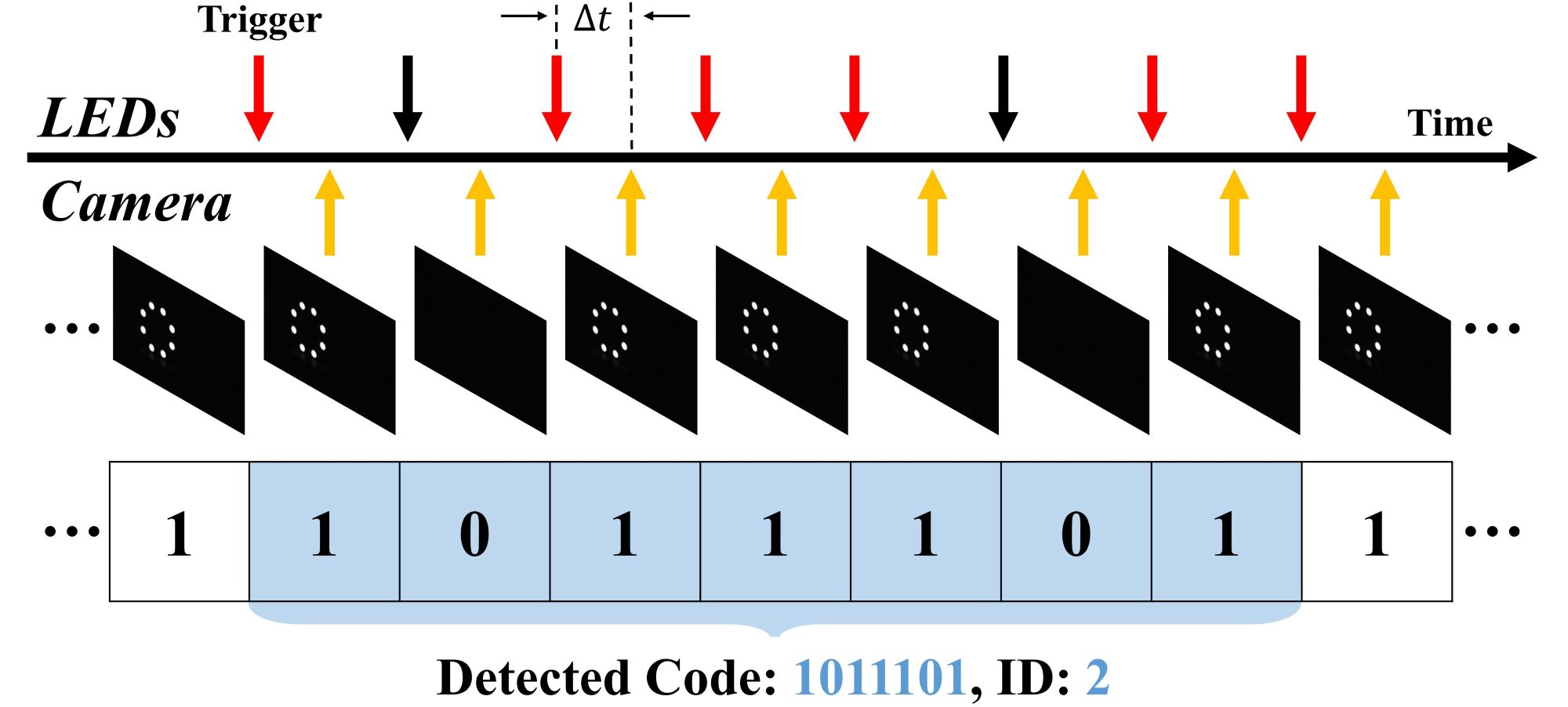}
    \caption{
        Synchronized LED-camera system with global clock triggering.
        Red arrows: LED on, black arrows: LED off, yellow arrows: camera capture.
        Sequential frames capture LED binary codes for ID extraction.
    }
    \label{Fig: ID Extraction}
\end{figure}

To improve detection speed, we propose a specially designed ID encoding and decoding scheme, as illustrated in \autoref{Fig: ID Extraction}.
All robots share a synchronized global clock, which alternately triggers the LEDs and the camera.
The triggering follows a simple division rule using the camera period, rounded to milliseconds.
The global clock, also in milliseconds, is divided by the camera period: triggers occur when the remainder is zero for the camera, and half the period for the LEDs.
When triggered, the camera captures an image, and the LEDs change state to encode the ID.
\autoref{Tab: ID Codebook} shows the codebook for 10 robots with 7-bit IDs, enabling ID extraction within 7 images while ensuring no more than 3 consecutive LED-off states, reducing the risk of tracking loss.
The codebook can be extended by increasing bits per ID to support more robots.

\begin{table}[ht]
    \centering
    \caption{ID Codebook}
    \label{Tab: ID Codebook}
    \begin{tabular}{|c|c|c|c|}
        \hline \rule{0pt}{8pt}
        \textbf{ID} & \textbf{Code} &
        \textbf{ID} & \textbf{Code}   \\
        \hline \rule{0pt}{8pt}
        0           & 1 0 1 1 1 1 1 &
        1           & 1 0 1 1 1 1 0   \\
        \hline \rule{0pt}{8pt}
        2           & 1 0 1 1 1 0 1 &
        3           & 1 0 1 1 0 1 0   \\
        \hline \rule{0pt}{8pt}
        4           & 1 0 1 1 1 0 0 &
        5           & 1 0 1 1 0 0 1   \\
        \hline \rule{0pt}{8pt}
        6           & 1 0 1 0 0 1 0 &
        7           & 1 0 0 1 1 1 0   \\
        \hline \rule{0pt}{8pt}
        8           & 1 0 0 1 1 1 1 &
        9           & 1 0 0 1 1 0 0   \\
        \hline
    \end{tabular}
\end{table}

This strategy ensures all cameras in different robots capture images at the same time, and all LEDs in robots flicker at the same time.
Moreover, the images are captured at the same time for all robots, which naturally synchronizes the bearing measurements.
The times of exposure and flicker need to be carefully designed to avoid LEDs' state changes during the exposure, which can result in false positives and incorrect ID extraction.
This makes the accuracy of time synchronization relevant to the frequency of the trigger signal.
To ensure proper staggering of LED and camera triggers across different robots, $\Delta t$ between triggers (equals to half-period of the camera) must exceed the synchronization error of the global clock.

Images captured by the camera are processed to extract bearing measurements and the corresponding peer IDs.
After binarization and edge extraction, we compute the minimum enclosing circles of the extracted edges to obtain candidate bright-spot centers $(u,v)$ in the image frame.
A distance-based clustering algorithm tracks these candidate bright spots within a temporal window by predicting their image locations from past observations and associating them with the nearest new detections.
The on/off sequence of each track is then recovered, and only tracks matching a valid LED code are decoded through the codebook and assigned peer IDs.
The latest coordinates $(u,v)$ of successfully decoded tracks are unprojected through the camera model to normalized bearing vectors $(x,y,z)$ in the camera frame.
The codebook avoids the pattern ``000'', so a valid LED should appear at least once in any three consecutive frames under normal visibility; in implementation, however, tracks are retained in a temporal buffer and removed only after a timeout without valid updates.
The camera is calibrated using Kalibr~\cite{furgale2013unified} with the double-sphere camera model~\cite{usenko2018double}.

\clearpage
\section{Problem Statement and Measurement Model}
\label{Sec: Problem Statement and Measurement Model}

\subsection{Notations}
\label{Sec: Notations}

\begin{table}[H]
    \centering
    \caption{
        \small
        Notations.
        We use $^{A}{X}_{B}^{T}$ to denote variable $X$ of coordinate frame $B$ expressed in coordinate frame $A$ at time $T$;
        e.g., ${}^{W}\mathbf{p}_{R_i}^{t_k}$ is the position of robot~$i$ in the world frame at time $t_k$.
    }
    \label{Tab: Notations}
    \begin{tabular}{c|l}
        \toprule
        \multicolumn{2}{l}{\textbf{Coordinate Frames}
        ( $A$ and $B$ of $^{A}{X}_{B}^{T}$ )}                                                                            \\
        \midrule
        $W$                        & The world frame                                                                     \\
        $R_r$                      & The body frame of robot itself, chosen as reference frame                           \\
        $F_u$                      & An unknown frame, whose transformation is unknown                                   \\
        $R_i$                      & The robot $i$'s frame, the same as its IMU frame                                    \\
        $B_i$                      & The robot $i$'s bearing frame, the same as its camera frame                         \\
        $D_i$                      & The robot $i$'s distance frame, the same as its UWB frame                           \\
        $M_i$                      & The robot $i$'s marker frame, the same as its LED frame                             \\
        $F_i \to F_j$              & The variable is from frame $F_i$ to frame $F_j$                                     \\
        \midrule
        \multicolumn{2}{l}{\textbf{Times}
        ( $T$ of ${}^{A}{X}_{B}^{T}$ )}                                                                                  \\
        \midrule
        $t_i$                      & The time $i$                                                                        \\
        $t_i \to t_j$              & The variable interval is from time $i$ to time $j$                                  \\
        \midrule
        \multicolumn{2}{l}{\textbf{Variables}
        ( $X$ of $^{A}{X}_{B}^{T}$ )}                                                                                    \\
        \midrule
        $\mathbf{p}$               & The $\mathbb{R}^{3}$ position                                                       \\
        $\mathbf{v}$               & The $\mathbb{R}^{3}$ linear velocity                                                \\
        $\mathbf{q}$               & The SO(3) rotation represented by Hamilton Quaternion                               \\
        $\mathbf{R}$               & The SO(3) rotation represented by $\mathbb{R}^{3\times3}$ matrix                    \\
        $\mathbf{G}$               & The O(3) $\mathbb{R}^{3\times3}$ matrix, rotation or reflection                     \\
        $\mathbf{a}$               & The $\mathbb{R}^{3}$ linear acceleration                                            \\
        $\bm{\omega}$              & The $\mathbb{R}^{3}$ angular velocity                                               \\
        $\mathbf{b}_{\mathbf{a}}$  & The $\mathbb{R}^{3}$ accelerometer bias                                             \\
        $\mathbf{b}_{\bm{\omega}}$ & The $\mathbb{R}^{3}$ gyroscope bias                                                 \\
        ${\hat{\mathbf{z}}_b}$     & The $\mathbb{R}^{3}$ bearing measurement, unit vector                               \\
        ${\hat{z}_d}$              & The $\mathbb{R}^{1}$ distance measurement                                           \\
        ${\hat{\mathbf{z}}_g}$     & The $\mathbb{R}^{3}$ gravity measurement, unit vector                               \\
        ${\hat{\mathbf{z}}_i}$     & The $\mathbb{R}^{6}$ inertial measurement, including $\mathbf{a}$ and $\bm{\omega}$ \\
        $\mathbf{n}$               & The noise of the measurement                                                        \\
        \midrule
        \multicolumn{2}{l}{\textbf{Operators}}                                                                           \\
        \midrule
        $\mathbf{R}\{\mathbf{q}\}$ & Convert quaternion $\mathbf{q}$ to rotation matrix $\mathbf{R} \in \text{SO}(3)$    \\
        $\mathbf{Log}(\mathbf{q})$ & Logarithmic map from $\text{SO}(3)$ (or unit quaternion) to $\mathbb{R}^{3}$        \\
        \bottomrule
    \end{tabular}
\end{table}

\subsection{Problem Statement}
\label{Sec: Problem Statement}

Assume $\mathcal{N} = \{0, 1, \ldots, n-1\}$ is the index set of $n$ robots in a system.
When a new image (or bearings) at time $t$ is captured, we generate a new frame $\mathfrak{F}^{t}$ including all measurements and relative states between robots at time $t$:
\begin{equation}
    \begin{aligned}
        \mathfrak{F}^{t} = \Bigl\{
         & \{ {\hat{\mathbf{z}}_b}{}^{t}_{B_j \to M_k} \}_{j \in \mathcal{N}, k \in \mathcal{B}^{t}_{j}}, \{ {\hat{z}_d}{}^{t}_{D_j \to D_k} \}_{j \in \mathcal{N}, k \in \mathcal{D}^{t}_{j}},
        \\
         & \{ {\hat{\mathbf{z}}_g}{}^{t}_{R_j} \}_{j \in \mathcal{N}}, \{ {\hat{\mathbf{z}}_i}{}^{t}_{R_j} \}_{j \in \mathcal{N}}, \mathfrak{X}^{t}_{R_{\mathcal N}}
        \Bigr\}
    \end{aligned}
\end{equation}
where $\mathcal{B}^{t}_{i}$ is the set of robots observed by robot $i$ using camera at time $t$, and $\mathcal{D}^{t}_{i}$ is the set of robots observed by robot $i$ using UWB at time $t$.
We define the state vector $\mathfrak{X}^{t}$ in $\mathfrak{F}^{t}$ as:
\begin{align}
    \mathfrak{X}^{t}_{R_{\mathcal{N}}} & = \left[ \mathfrak{X}^{t}_{R_0}, \mathfrak{X}^{t}_{R_1}, \cdots, \mathfrak{X}^{t}_{R_{n-1}}, {^{R_r}\mathbf{g}^{t}} \right],
    \\
    \mathfrak{X}^{t}_{R_j}             & = \left[ {^{R_r}\mathbf{p}_{R_j}^{t}}, {^{R_r}\mathbf{v}_{R_j}^{t}}, {^{R_r}\mathbf{q}_{R_j}^{t}},{^{R_j}\mathbf{b}_{\mathbf{a}}}_{R_j}^{t},{^{R_j}\mathbf{b}_{\bm{\omega}}}_{R_j}^{t} \right], j \in \mathcal{N},\notag
\end{align}
where the reference frame $R_r$ is the body frame of robot itself.
Noting that the states ${^{R_r}\mathbf{p}_{R_r}^{t_i}}$, ${^{R_r}\mathbf{v}_{R_r}^{t_i}}$, and ${^{R_r}\mathbf{q}_{R_r}^{t_i}}$ are always equal to the identity, we retain them in state vector for notational convenience in the subsequent formulations.

To alleviate the computational burden, we only use the frames $\mathfrak{F}^{t_i}$ in a sliding window, where $i \in \mathcal{M}, \mathcal{M} = \{0, 1, \ldots, m-1\}$ is the index set of the times in the window.
The full state vector $\mathfrak{X}^{t_{\mathcal{M}}}$ in the sliding window is:
\begin{align}
    \mathfrak{X}^{t_{\mathcal{M}}}_{R_{\mathcal{N}}} & = \left[ \mathfrak{X}^{t_0}_{R_{\mathcal{N}}}, \mathfrak{X}^{t_1}_{R_{\mathcal{N}}}, \cdots, \mathfrak{X}^{t_{m-1}}_{R_{\mathcal{N}}} \right],\notag
    \\
    \mathfrak{X}^{t_i}_{R_{\mathcal{N}}}             & = \left[ \mathfrak{X}^{t_i}_{R_0}, \mathfrak{X}^{t_i}_{R_1}, \cdots, \mathfrak{X}^{t_i}_{R_{n-1}}, {^{R_r}\mathbf{g}^{t_i}} \right], i \in \mathcal{M},
    \\
    \mathfrak{X}^{t_i}_{R_j}                         & = \left[ {^{R_r}\mathbf{p}_{R_j}^{t_i}}, {^{R_r}\mathbf{v}_{R_j}^{t_i}}, {^{R_r}\mathbf{q}_{R_j}^{t_i}},{^{R_j}\mathbf{b}_{\mathbf{a}}}_{R_j}^{t_i},{^{R_j}\mathbf{b}_{\bm{\omega}}}_{R_j}^{t_i} \right], j \in \mathcal{N}. \notag
\end{align}

We also select keyframes $\mathfrak{KF}$ from the sliding window to ensure optimization efficiency and accuracy.
Let $\mathcal{M}_k \subset \mathcal{M}$ denote the index set of keyframes, and define $\mathcal{M}_k' = \mathcal{M}_k \cup \{m-1\}$ as the set comprising all keyframe indices together with the latest frame.
It is important to note that during the sliding process, frames are removed until the first frame is ensured to be a keyframe, and the number of keyframes within the window remains constant.
Also, the latest frame is not necessarily a keyframe.
The optimized state vector is then:
\begin{equation}
    \mathfrak{X}^{t_{\mathcal{M}_k'}}_{R_{\mathcal{N}}} = \left[ \mathfrak{X}^{t_i}_{R_{\mathcal{N}}} \mid i \in \mathcal{M}_k' \right].
\end{equation}

\subsection{Distance Measurement}
\label{Sec: Distance Measurement}

Considering the extrinsic between the UWB and the IMU, the relative position of the distance frame $D_i,D_j$ of two robots in the reference frame can be calculated as:
\begin{align}
     & {^{R_r}\mathbf{p}_{D_i \to D_j}} =
    \\
     & \quad \underbrace{({^{R_r}\mathbf{p}_{R_j}}+{^{R_r}\mathbf{R}_{R_j}}{^{R_j}\mathbf{p}_{D_j}})}_{\text{j's absolute UWB position}} - \underbrace{({^{R_r}\mathbf{p}_{R_i}}+{^{R_r}\mathbf{R}_{R_i}}{^{R_i}\mathbf{p}_{D_i}})}_{\text{i's absolute UWB position}}.\notag
\end{align}

The UWB ranging model is defined as:
\begin{equation}
    {\hat{z}_d}{}_{D_i \to D_j} = \left\| {^{R_r}\mathbf{p}_{D_i \to D_j}} \right\|
    + n_{d},
    \label{Equ: Distance Measurements}
\end{equation}
where $n_{d} \sim \mathcal{N}(0, \sigma_{d}^2)$ is the line-of-sight (LOS) ranging noise.
The residuals can be calculated as:
\begin{equation}
    {r}_d ({\hat{z}_d}{}_{D_i \to D_j}, \mathfrak{X}_{R_{\mathcal{N}}}) = \left\| {^{R_r}\mathbf{p}_{D_i \to D_j}} \right\|
    - {\hat{z}_d}{}_{D_i \to D_j}.
    \label{Equ: Distance Residuals}
\end{equation}

This Gaussian model is accurate under LOS conditions.
Under NLOS conditions, however, multipath propagation and transmission through obstacles increase the effective time of flight.
In addition, antenna-orientation effects and out-of-plane antenna geometries can further degrade ranging accuracy.
Together, these effects produce a one-sided, heavy-tailed positive range bias, which drives the residual $r_d$ negative under our residual definition and cannot be properly captured by a symmetric cost~\cite{jiang2024ranging}.
To improve robustness, we adopt the half-Cauchy ranging model of~\cite{jiang2024ranging}, which weights $r_d$ with a Gaussian likelihood on the inlier side and a Cauchy heavy tail on the NLOS-biased side.
This yields an implicit M-estimator that reduces the influence of over-long ranges without requiring a hard LOS/NLOS decision for every measurement.
The robust loss function $\rho(r_d)$ of distance residual is calculated as:
\begin{equation}
    \begin{aligned}
        \mathcal{L}(r_d) & =
        \begin{cases}
            \dfrac{1}{2}\left(\dfrac{r_d}{\sigma_d}\right)^{2},     & r_d \geq 0, \\[6pt]
            \log\!\left(1+\dfrac{r_d^{\,2}}{\gamma_d^{\,2}}\right), & r_d < 0,
        \end{cases}
        \\
        \rho(r_d)        & = \mathrm{sign}(r_d)\sqrt{2\mathcal{L}(r_d)}
    \end{aligned}
    \label{Equ: Distance NLL Asymmetric}
\end{equation}
where $\gamma_d$ is the half-Cauchy scale (we use $\gamma_d=\sigma_d$).

\subsection{Bearing Measurement}
\label{Sec: Bearing Measurement}

Considering the extrinsic between the camera, marker, and the IMU, the relative position of the bearing frame $B_i$ and marker frame $M_j$ in the reference frame can be calculated as:
\begin{align}
     & {^{R_r}\mathbf{p}_{B_i \to M_j}} =
    \\
     & \quad \underbrace{({^{R_r}\mathbf{p}_{R_j}}+{^{R_r}\mathbf{R}_{R_j}}{^{R_j}\mathbf{p}_{M_j}})}_{\text{j's absolute marker position}} - \underbrace{({^{R_r}\mathbf{p}_{R_i}}+{^{R_r}\mathbf{R}_{R_i}}{^{R_i}\mathbf{p}_{B_i}})}_{\text{i's absolute camera position}}.\notag
\end{align}

The camera detection model is defined as:
\begin{equation}
    {\hat{\mathbf{z}}_b}{}_{B_i \to M_j} =
    (\underbrace{{^{R_r}\mathbf{R}_{R_i}} {^{R_i}\mathbf{R}_{B_i}}}_{\text{i's camera rotation}})^{-1}
    \frac{{^{R_r}\mathbf{p}_{B_i \to M_j}}}{\left\| {^{R_r}\mathbf{p}_{B_i \to M_j}} \right\|}
    + \mathbf{n}_{b},
    \label{Equ: Bearing Measurements}
\end{equation}
where $\mathbf{n}_{b} \sim \mathcal{N}(0, \bm{\Sigma}_{b}=\sigma_b^2 \mathbf{I})$ is the noise of the camera.
In particular, when $i$ and $j$ represent the same robot, the bearing is defined as a zero vector.
The residuals can be calculated as:
\begin{equation}
    \begin{aligned}
         & \mathbf{r}_b ({\hat{\mathbf{z}}_b}{}_{B_i \to M_j}, \mathfrak{X}_{R_{\mathcal{N}}}) =
        \\
         & \quad ({^{R_r}\mathbf{R}_{R_i}} {^{R_i}\mathbf{R}_{B_i}})^{-1}
        \frac{{^{R_r}\mathbf{p}_{B_i \to M_j}}}{\left\| {^{R_r}\mathbf{p}_{B_i \to M_j}} \right\|}
        - {\hat{\mathbf{z}}_b}{}_{B_i \to M_j}.
    \end{aligned}
    \label{Equ: Bearing Residuals}
\end{equation}
We use the difference of the two unit vectors in $\mathbb{R}^{3}$ rather than in $\mathbb{S}^{2}$, which is more efficient and nearly the same in practice.

\subsection{Gravity Measurement}
\label{Sec: Gravity Measurement}

Gravity can be estimated by an IMU using the complementary filter\cite{mahony2008nonlinear}, and it should be the same for all robots in the same reference frame.
The measurement is modeled as:
\begin{equation}
    {\hat{\mathbf{z}}_g}{}_{R_i} = (^{R_r}\mathbf{R}_{R_i})^{-1}{^{R_r}\mathbf{g}} + \mathbf{n}_{g},
    \label{Equ: Gravity Measurements}
\end{equation}
where $\mathbf{n}_{g} \sim \mathcal{N}(0,\bm{\Sigma}_{g}=\sigma_g^2\mathbf{I})$ is the noise of the IMU.
As we only focus on the direction of gravity, the gravity is represented by a unit vector.
The residuals can be calculated as:
\begin{equation}
    \mathbf{r}_g ({\hat{\mathbf{z}}_g}{}_{R_i}, \mathfrak{X}_{R_{\mathcal{N}}}, {^{R_r}\mathbf{g}})
    = (^{R_r}\mathbf{R}_{R_i})^{-1}{^{R_r}\mathbf{g}} - {\hat{\mathbf{z}}_g}{}_{R_i}.
    \label{Equ: Gravity Residuals}
\end{equation}

\textbf{Optional Gravity:}
In most real-world scenarios (e.g., indoor and outdoor), gravity can be reliably estimated.
With gravity measurements, both the consensus gravity direction and the relative yaw angles of other robots can be directly estimated.
However, in NI conditions, gravity measurements are unreliable; hence, only bearing and distance measurements are used for position and rotation estimation.
Without gravity measurements, additional bearings are required to constrain rotations, which reduces system observability.
To unify both cases, we introduce a binary weight $w_g$ to control the influence of gravity in the estimation:
\begin{equation}
    w_g =
    \begin{cases}
        1, & \text{gravity available},   \\
        0, & \text{gravity unavailable}.
    \end{cases}
\end{equation}

\subsection{IMU Measurement and Preintegration}
\label{Sec: IMU Measurement and Preintegration}

To fuse the linear acceleration and angular velocity measurements from the IMU, the preintegration method is needed in the inertial-based optimization.
We follow the method in \cite{qin2018vins} to build the IMU preintegration.
The raw IMU measurements, biases, and preintegrated quantities are expressed in the corresponding robot body frame.
For brevity, we omit the leading frame superscript in this subsection; for example, $\mathbf{a}_{R_i}^{t_k}$ and $\bm{\omega}_{R_i}^{t_k}$ should be understood as ${}^{R_i}\mathbf{a}_{R_i}^{t_k}$ and ${}^{R_i}\bm{\omega}_{R_i}^{t_k}$, respectively.
The IMU model of robot $i$ is:
\begin{equation}
    \begin{aligned}
        \hat{\mathbf{a}}_{R_i}  & = \mathbf{a}_{R_i}+{\mathbf{b}_{\mathbf{a}}}_{R_i}+ {^{R_i}\mathbf{R}_{W}}{^{W}\mathbf{g}}+\mathbf{n}_{a},
        \\
        \hat{\bm{\omega}}_{R_i} & = \bm{\omega}_{R_i}+{\mathbf{b}_{\bm{\omega}}}_{R_i}+\mathbf{n}_{\omega}.
    \end{aligned}
\end{equation}
The IMU preintegration of robot $i$ is calculated as:
\begin{equation}
    \begin{aligned}
        \bm{\alpha}^{t_0 \to t_1}_{R_i} & = \iint_{t_0}^{t_1} \mathbf{R}^{t_0 \to t}_{R_i}(\hat{\mathbf{a}}^{t}_{R_i}-{\mathbf{b}_{\mathbf{a}}}_{R_i}^{t})~dt^2,
        \\
        \bm{\beta}^{t_0 \to t_1}_{R_i}  & = \int_{t_0}^{t_1} \mathbf{R}^{t_0 \to t}_{R_i}(\hat{\mathbf{a}}^{t}_{R_i}-{\mathbf{b}_{\mathbf{a}}}_{R_i}^{t})~dt,
        \\
        \bm{\gamma}^{t_0 \to t_1}_{R_i} & = \int_{t_0}^{t_1} \frac{1}{2}\bm{\Omega}(\hat{\bm{\omega}}^{t}_{R_i}-{\mathbf{b}_{\bm{\omega}}}_{R_i}^{t})\bm{\gamma}^{t_0 \to t}_{R_i}~dt.
    \end{aligned}
\end{equation}

\section{Single-Frame Relative Estimator}
\label{Sec: Single-Frame Relative Estimator}

\begin{figure}[ht]
    \centering
    \includegraphics[width=\halfwidth]{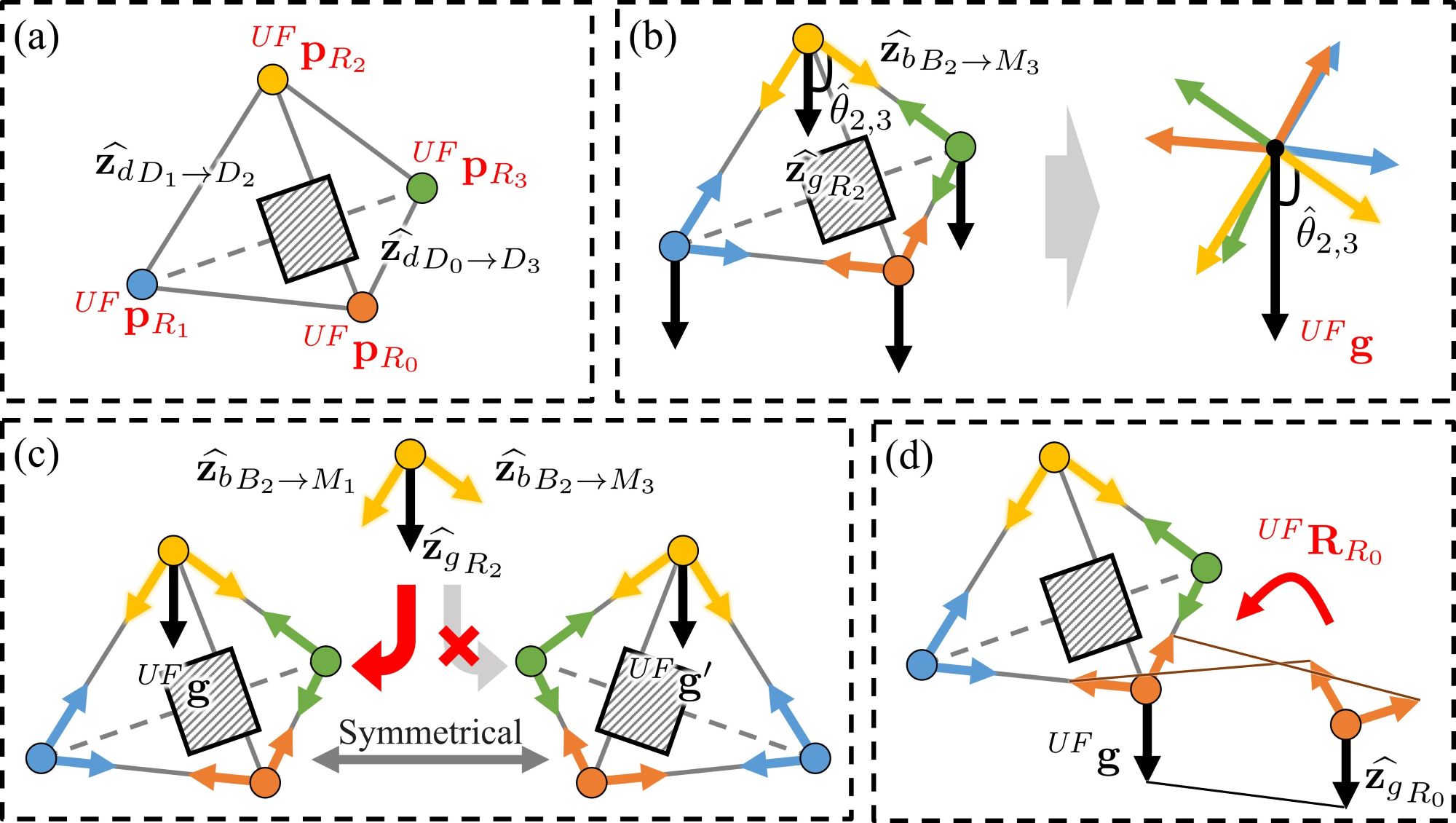}
    \caption{
        Relative pose estimation pipeline. The estimated state variable in each stage is in red.
        (a) Position ${^{F_u}\mathbf{p}_{R_i}}$ estimated from distances ${\hat{z}_d}{}_{D_i \to D_j}$.
        (b) Gravity ${^{F_u}\mathbf{g}}$ estimated via bearing-gravity angles $\hat{\theta}_{i,j}$.
        (c) Chirality determined by checking bearing (and gravity) consistency.
        (d) Rotation estimated by aligning ${\hat{\mathbf{z}}_b}{}_{B_i \to M_j}$ with ${^{F_u}\mathbf{p}_{R_i \to R_j}}$ (and ${\hat{\mathbf{z}}_g}{}_{R_i}$ with ${^{F_u}\mathbf{g}}$).
    }
    \label{Fig: SFC: Pose Estimation}
\end{figure}

Using one-shot bearing and distance measurements, we can estimate the relative poses between the robots, shown in \autoref{Fig: SFC: Pose Estimation}.
As this estimation only concerns a single frame, inertial measurements are not involved, and only the pose variables $ [{^{R_r}\mathbf{p}_{R_j}^{t}}, {^{R_r}\mathbf{q}_{R_j}^{t}}], j \in \mathcal{N}$ are estimated.
For simplicity, we omit the time index $t$ in the following variables in this section.
In this part, we illustrate the estimation pipeline without considering outliers, and the method to handle outliers is explicitly explained in \autoref{Sec: Outlier Rejection}.

\subsection{Single-Frame Closed-Form Solver (SFC)}
\label{Sec: Single-Frame Closed-Form Solver}

We propose a closed-form algorithm to compute the relative poses while considering NLOS conditions and maximizing the usage of all measurements.
Here, only rotational extrinsics between sensors are considered, as translational extrinsics are small compared to translations between robots in most cases.
This simplification can be refined in subsequent optimization.

\subsubsection{Position Estimation}
\label{Sec: SFC: Position Estimation}

The positions are estimated using mutual distance measurements by formulating a classical Multi-Dimensional Scaling (MDS) problem \cite{torgerson1952multidimensional}.
The solution can be subjected to arbitrary transformations, so its reference frame is an unknown reference frame, denoted as $F_u$.

Let $\mathbf{X} = [{^{F_u}\mathbf{p}_{R_0}}, {^{F_u}\mathbf{p}_{R_1}}, \cdots, {^{F_u}\mathbf{p}_{R_{n-1}}}]^T\in\mathbb{R}^{n\times3}$ denote the positions of $n$ robots.
The distance matrix $\mathbf{D} \in \mathbb{R}^{n \times n}$ (not to be confused with distance frame $D_i$) is constructed from pairwise UWB measurements:
\begin{equation}
    D_{ij} = {\hat{z}_d}{}^{t}_{D_i \to D_j},~ i \in \mathcal{N},~ j \in \mathcal{D}_i,
\end{equation}
where $\mathcal{D}_i$ is the set of robots observed by robot $i$.
The MDS formulation seeks to minimize the error of the distance matrix:
\begin{equation}
    \min_{\mathbf{X}} \sum_{i<j} \left( D_{ij} - \|{^{F_u}\mathbf{p}_{R_i}} - {^{F_u}\mathbf{p}_{R_j}}\| \right)^2,
    \label{Equ: SFC: MDS Problem}
\end{equation}
which admits a closed-form solution:
\begin{equation}
    \mathbf{H} = \mathbf{I} - \frac{1}{n} \mathbf{J},
    \mathbf{B} = -\frac{1}{2} \mathbf{H} \mathbf{D}^{(2)} \mathbf{H} = \mathbf{V} \mathbf{\Lambda} \mathbf{V}^T,
    \mathbf{X} = \mathbf{V} \mathbf{\Lambda}^{\frac{1}{2}},
    \label{Equ: SFC: MDS Solution}
\end{equation}
where $\mathbf{J} \in \mathbb{R}^{n \times n}$ is a matrix with all entries equal to 1.

\subsubsection{Gravity Estimation}
\label{Sec: SFC: Gravity Estimation}

If constant gravity assumption holds, the global gravity ${^{F_u}\mathbf{g}}$ can be estimated from each robot $i$'s gravity measurements ${\hat{\mathbf{z}}_g}{}_{R_i}$.
As ${\hat{\mathbf{z}}_g}{}_{R_i}$ and ${^{F_u}\mathbf{g}}$ are defined in different frames, their alignment needs bearing information.
To fully exploit the gravity and bearing measurements from all robots, we estimate ${^{F_u}\mathbf{g}}$ by minimizing the angular discrepancy between the bearings and the gravity.

Considering one bearing from robot $i$ to robot $j$, the angle ${\hat{\theta}_{i,j}}$ between bearing ${{\hat{\mathbf{z}}_b}{}_{B_i \to M_j}}$ and gravity ${{\hat{\mathbf{z}}_g}{}_{R_i}}$ is:
\begin{equation}
    {\hat{\theta}_{i,j}} = \cos^{-1} \left( {^{R_i}\mathbf{R}_{B_i}} {{\hat{\mathbf{z}}_b}{}_{B_i \to M_j}} \cdot {{\hat{\mathbf{z}}_g}{}_{R_i}}\right).
\end{equation}
We minimize the error of the angles between bearings and gravity to estimate the global gravity ${^{F_u}\mathbf{g}}$:
\begin{equation}
    {^{F_u}\tilde{\mathbf{p}}_{R_i \to R_j}} = {^{F_u}\mathbf{p}_{R_i \to R_j}} / \| {^{F_u}\mathbf{p}_{R_i \to R_j}} \|,
\end{equation}
\begin{equation}
    \min_{^{F_u}\mathbf{g} \in \mathbb{S}^2} \sum_{i \in \mathcal{N}} \sum_{j \in \mathcal{B}_i} \left\| \cos^{-1}({^{F_u}\tilde{\mathbf{p}}_{R_i \to R_j}} \cdot {^{F_u}\mathbf{g}}) - \hat{\theta}_{i,j} \right\|_{\sigma_\theta^2}^2,
\end{equation}
where $\mathcal{B}_i$ is the set of robots observed by robot $i$ using camera, and $\sigma_\theta^2=\sigma_b^2+\sigma_g^2$ is the linearized covariance of the angle.
In this way, all bearings observed by different robots can be used to estimate the gravity.
We relax the problem by removing the unit constraint of ${^{F_u}\mathbf{g}}$ and use the cosine error of the angles:
\begin{equation}
    \min_{^{F_u}\mathbf{g}} \sum_{i \in \mathcal{N}} \sum_{j \in \mathcal{B}_i} \left\| {^{F_u}\tilde{\mathbf{p}}_{R_i \to R_j}} \cdot {^{F_u}\mathbf{g}} - \cos(\hat{\theta}_{i,j}) \right\|_{\sigma_{\cos(\hat{\theta}_{i,j})}^2}^2,
\end{equation}
where $\sigma_{\cos(\hat{\theta}_{i,j})}=\sin(\hat{\theta}_{i,j})\sigma_\theta$ is the linearized covariance of the cosine of the angle at $\hat{\theta}_{i,j}$.
For numerical stability, we limit the minimum value of $\sigma_{\cos(\hat{\theta}_{i,j})}$ to $10^{-2}$.
The above problem can be rewritten as a linear least squares problem:
\begin{gather}
    \mathbf{A} = \begin{bmatrix}
        {^{F_u}\tilde{\mathbf{p}}_{R_0 \to R_1}}/\sigma_{\cos(\hat{\theta}_{0,1})} \\
        {^{F_u}\tilde{\mathbf{p}}_{R_0 \to R_2}}/\sigma_{\cos(\hat{\theta}_{0,2})} \\
        \cdots                                                                     \\
    \end{bmatrix},
    \mathbf{b} = \begin{bmatrix}
        \cos(\hat{\theta}_{0,1})  /\sigma_{\cos(\hat{\theta}_{0,1})} \\
        \cos(\hat{\theta}_{0,2})  /\sigma_{\cos(\hat{\theta}_{0,2})} \\
        \cdots                                                       \\
    \end{bmatrix},\notag
    \\
    \min_{^{F_u}\mathbf{g}} \left\| \mathbf{A} {^{F_u}\mathbf{g}}-\mathbf{b} \right\|^2,
\end{gather}
where $\mathbf{A}$ and $\mathbf{b}$ contain all the robot pairs in measured bearings and angles between the bearings and the gravity.

Considering that $\mathbf{A}$ has a much larger number of rows than columns and low row rank, we solve the dual problem with the same optimal solution for best performance:
\begin{equation}
    \min_{^{F_u}\mathbf{g}} \left\| \mathbf{A}^T \mathbf{A} {^{F_u}\mathbf{g}} - \mathbf{A}^T \mathbf{b} \right\|^2.
\end{equation}

In NLOS conditions, where some bearings may be unavailable, the matrix $\mathbf{A}^T \mathbf{A}$ can become singular.
To handle this, we compute the least-squares solution using the pseudoinverse obtained from Singular Value Decomposition (SVD).
\begin{equation}
    \begin{gathered}
        \mathbf{P} = \mathbf{A}^T \mathbf{A} = \mathbf{U} \bm{\Sigma} \mathbf{V}^T,~
        \mathbf{P}^+ = \mathbf{V} \bm{\Sigma}^+ \mathbf{U}^T,
        \\
        ^{F_u}\mathbf{g} = \mathbf{P}^+ \mathbf{A}^T \mathbf{b} + \mathbf{P}^+ \mathbf{P} \bm{\nu},
    \end{gathered}
\end{equation}
where $\bm{\Sigma}^+$ is the pseudoinverse of $\bm{\Sigma}$ by taking the reciprocal of each non-zero element on the diagonal while the others are zero, and $\mathbf{P}^+$ is the Moore-Penrose inverse of $\mathbf{P}$.
Note that $\bm{\nu}$ can be any vector, and by setting $\bm{\nu}$ to $\mathbf{0}$, we can get the minimum norm solution $^{F_u}\mathbf{g}$ of the least square problem.
While the condition $^{F_u}\mathbf{g} \in \mathbb{S}^2$ was not explicitly enforced, when the constraint is sufficient, the error remains negligible in practice.
$^{F_u}\mathbf{g}$ will be normalized to ensure it is a unit vector.

\begin{figure}[t]
    \centering
    \includegraphics[width=\halfwidth]{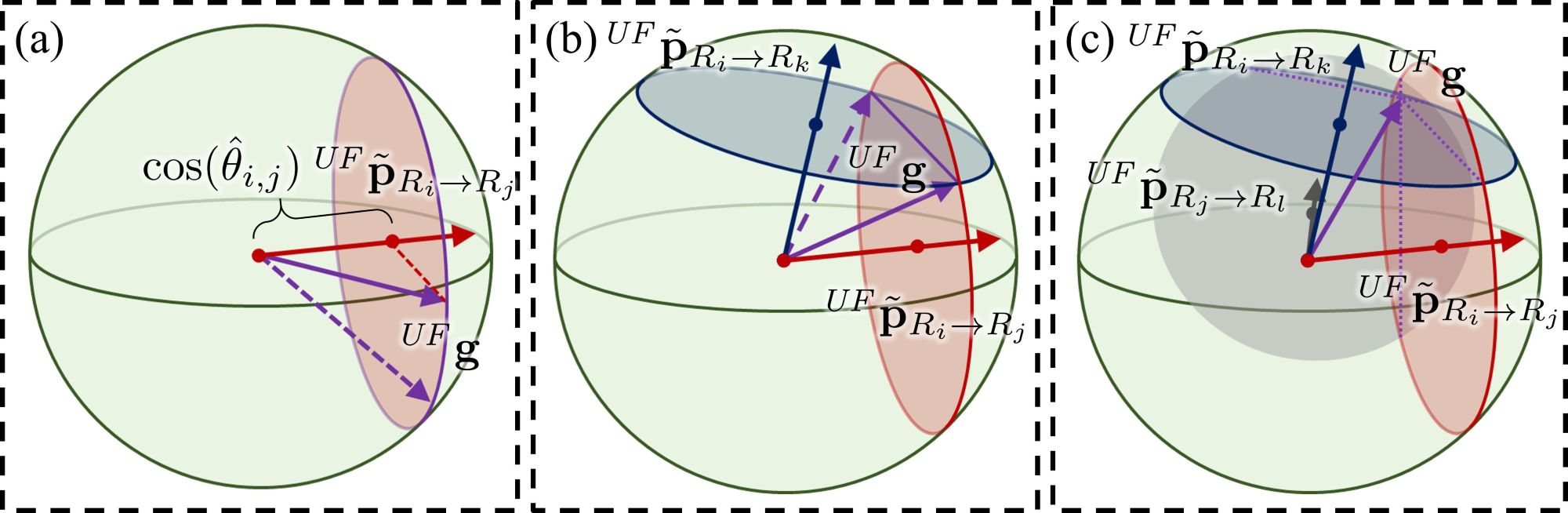}
    \caption{
        Illustration of gravity estimation and gravity constraint matrix $\mathbf{A}$.
        (a) The row rank of $\mathbf{A}$ is 1, the gravity can take any value on a circle.
        (b) The row rank of $\mathbf{A}$ is 2, the gravity has two possible values.
        (c) The row rank of $\mathbf{A}$ is 3 or more, the gravity is fully constrained.
    }
    \label{Fig: SFC: Gravity}
\end{figure}

We observe that when the row rank of matrix $\mathbf{A}$ is less than 3, the constraints are insufficient to uniquely determine the gravity vector.
Considering this property, we refer to $\mathbf{A}$ as the gravity constraint matrix, as illustrated in \autoref{Fig: SFC: Gravity}.
In such cases, the solution from least squares may not lie near the unit sphere.
To address this, we add a vector with magnitude $\sqrt{1 - |^{F_u}\mathbf{g}|^2}$ to the estimated gravity $^{F_u}\mathbf{g}$ to move it onto the unit sphere.

When $rank(\mathbf{A}) = 1$, the estimated gravity lies on a plane, indicating collinear robot positions.
In this case, the null space of $\mathbf{A}$ has dimension two, and any perturbation $\bm{\nu}$ within that plane leads to equivalent estimations.

When $rank(\mathbf{A}) = 2$, the estimated gravity lies on a line, indicating coplanar robot positions.
The ambiguity reduces to a sign ambiguity: both $\bm{\nu}$ and $-\bm{\nu}$ yield valid gravity estimates.
This arises because only directional constraints are used, while the spatial constraints of the measurements are ignored.

If any robot has more than two bearing measurements, the plane spanned by any pair of them can be used to verify the correct orientation of gravity.
Specifically, the projection of the estimated gravity onto the normal of the bearing plane should match that of the measured gravity:
\begin{equation}
    \begin{aligned}
          & ({^{F_u}\tilde{\mathbf{p}}_{R_i \to R_j}} \times {^{F_u}\tilde{\mathbf{p}}_{R_i \to R_k}}) \cdot {^{F_u}\mathbf{g}}
        \\
        = & {^{R_i}\mathbf{R}_{B_i}} ({{\hat{\mathbf{z}}_b}{}_{B_i \to M_j}} \times {{\hat{\mathbf{z}}_b}{}_{B_i \to M_k}}) \cdot {{\hat{\mathbf{z}}_g}{}_{R_i}}.
    \end{aligned}
\end{equation}
By verifying the above condition across all robots, we can disambiguate between the two possible gravity directions and select the one that is consistent with the measurements.

To simplify the following estimation, we rotate $F_u$ to a new frame $F_u'$ so that the estimated gravity vector aligns with the positive $z$-axis.
Let $^{F_u'}\mathbf{R}_{F_u}$ denote the rotation from $F_u'$ to $F_u$, in which the gravity vector and positions become:
\begin{equation}
    \begin{aligned}
        {^{F_u'}\mathbf{g}}       & = {^{F_u'}\mathbf{R}_{F_u}} {^{F_u}\mathbf{g}} = (0,0,1)^T,
        \\
        {^{F_u'}\mathbf{p}_{R_i}} & = {^{F_u'}\mathbf{R}_{F_u}} {^{F_u}\mathbf{p}_{R_i}}.
    \end{aligned}
\end{equation}
For notational consistency, we continue to denote the transformed frame as $F_u$ in the remainder of this paper.

\subsubsection{Position Chirality Determination}
\label{Sec: SFC: Position Chirality Determination}

\begin{figure}[t]
    \centering
    \includegraphics[width=\halfwidth]{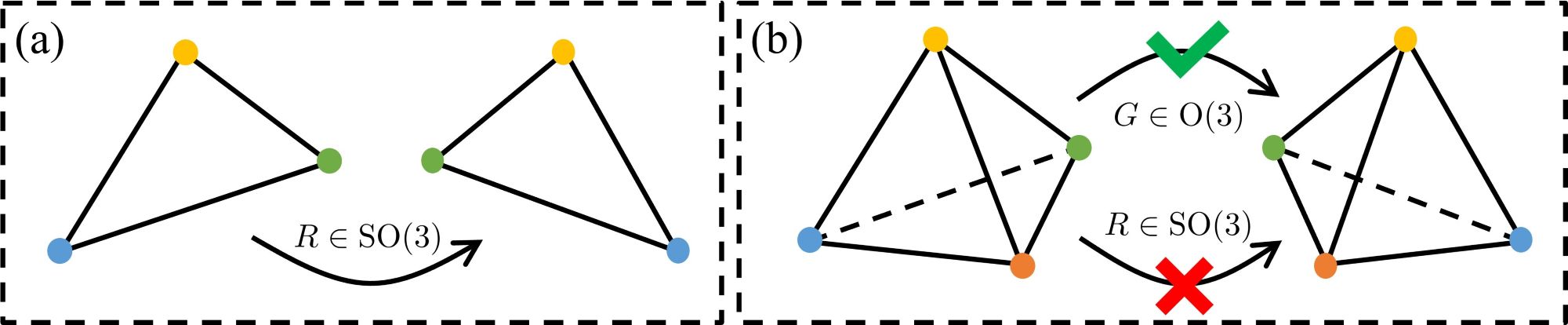}
    \caption{
        Chirality exists when the positions of robots are not coplanar.
        (a) The positions can be transformed by a rotation (in SO(3)).
        (b) The positions can only be transformed by a reflection (in O(3) but not SO(3)).
    }
    \label{Fig: SFC: Chirality}
\end{figure}

Since pairwise distances are symmetric, the MDS solution is ambiguous up to a reflection, i.e., it exhibits chirality: a mirror-image configuration exists that cannot be resolved by any rigid-body rotation (see \autoref{Fig: SFC: Chirality}).
This ambiguity becomes structurally relevant when the robot positions are non-coplanar and the number of robots exceeds three.
As UWB measurements alone are insufficient to resolve chirality, additional spatial cues (such as bearings or gravity) are required to disambiguate the solution.

We use bearing and gravity measurements (if available) to determine the chirality of the positions.
This estimation is formulated as a Wahba's problem \cite{wahba1965least}, which aligns the rotated bearing vectors and local gravity to their corresponding references in a common frame.
Specifically, we solve:
\begin{equation}
    \begin{aligned}
        \min_{{^{F_u}\mathbf{R}_{R_i}}} \sum_{j \in \mathcal{B}_i} \left\| {^{F_u}\tilde{\mathbf{p}}_{R_i \to R_j}} - {^{F_u}\mathbf{R}_{R_i}} {^{R_i}\mathbf{R}_{B_i}} {\hat{\mathbf{z}}_b}{}_{B_i \to M_j} \right\|_{\bm{\Sigma}_b}^2
        \\
        + {w_g} \left\| {^{F_u}\mathbf{g}} - {^{F_u}\mathbf{R}_{R_i}} {{\hat{\mathbf{z}}_g}{}_{R_i}} \right\|_{\bm{\Sigma}_g}^2.
    \end{aligned}
    \label{Equ: SFC: Wahba Problem}
\end{equation}
This problem can be efficiently solved via SVD \cite{markley1988attitude}:
\begin{align}
    \mathbf{K} & = \frac{1}{\sigma_b} \sum_{j \in \mathcal{B}_i} ({^{R_i}\mathbf{R}_{B_i}} {{\hat{\mathbf{z}}_b}{}_{B_i \to M_j}} {^{F_u}\tilde{\mathbf{p}}_{R_i \to R_j}^T}) + \frac{w_g}{\sigma_g} ({{\hat{\mathbf{z}}_g}{}_{R_i}}{^{F_u}\mathbf{g}^T}) \notag
    \\
               & = \mathbf{U} \bm{\Sigma} \mathbf{V}^T, ~
    {^{F_u}\mathbf{G}_{R_i}} = \mathbf{U} \mathbf{V}^T.
    \label{Equ: SFC: Wahba Solution}
\end{align}
However, the SVD-based solution may yield a reflection in O(3) rather than in SO(3) (i.e., $\det({^{F_u}\mathbf{G}_{R_i}}) = -1$), which is not a valid rotation.
We correct the sign to obtain the optimal rotation (which may be a suboptimal solution):
\begin{equation}
    {^{F_u}\mathbf{R}_{R_i}} = \mathbf{U} \mathbf{S} \mathbf{V}^T,~
    \mathbf{S} = diag(1,1,\det(\mathbf{U} \mathbf{V}^T)).
    \label{Equ: SFC: Wahba Solution Correction}
\end{equation}

A negative determinant typically suggests that the estimated positions lie in the wrong chirality, as the optimal solution is a reflection rather than a rotation.
However, in nearly planar configurations with measurement noise, the sign of the determinant may become unreliable.

To determine the correct chirality, we adopt a heuristic method that considers the coplanarity of the bearing and gravity measurements.
For robot $i$, if at least three bearing measurements are available (or two bearings with gravity), we assess coplanarity via principal component analysis:
\begin{gather}
    \mathbf{\overline{p}}_{i} = \frac{1}{|\mathcal{B}_i'|+{w_g}}
    (
    \sum_{j \in \mathcal{B}_i'} ({^{R_i}\mathbf{R}_{B_i}} {{\hat{\mathbf{z}}_b}{}_{B_i \to M_j}})/{\sigma_b} +
    {w_g} ({{\hat{\mathbf{z}}_g}{}_{R_i}})/{\sigma_g}
    ),\notag
    \\
    \mathbf{Q}_i = \begin{bmatrix}
        (({^{R_i}\mathbf{R}_{B_i}} {{\hat{\mathbf{z}}_b}{}_{B_i \to M_0}})/{\sigma_b} - \mathbf{\overline{p}}_{i})^T
        \\
        \cdots
        \\
        (({{\hat{\mathbf{z}}_g}{}_{R_i}})/{\sigma_g} - \mathbf{\overline{p}}_{i})^T,~\text{if}~{w_g} \neq 0
    \end{bmatrix}
    ,~
    \mathbf{C}_{i} = \frac{\mathbf{Q}_i^T \mathbf{Q}_i}{|\mathcal{B}_i'|+{w_g}},\notag
    \\
    \lambda_1, \lambda_2, \lambda_3 = Eigenvalues(\mathbf{C}_{i}),~ \lambda_1 \geq \lambda_2 \geq \lambda_3,
\end{gather}
where $\mathcal{B}_i' = \mathcal{B}_i \cup \{i\}$, and the coplanarity is quantified as:
\begin{equation}
    \mathrm{CP}_{i} = {\lambda_3} / {(\lambda_1+\lambda_2+\lambda_3)}.
\end{equation}
A low $\mathrm{CP}_i$ (close to zero) indicates a nearly planar configuration, in which the sign of $|{^{F_u}\mathbf{G}_{R_i}}|$ may become unreliable.
If fewer than three bearings are available, we set $\mathrm{CP}_i = 0$.

To evaluate each chirality, we define a score:
\begin{equation}
    \mathrm{Score} = \sum_{i\in\mathcal{N}}{\mathrm{CP}_{i} \times | ^{F_u}\mathbf{G}_{R_i} |},
\end{equation}
and compute the scores for both chiralities and select the one with the higher score.
To ensure robustness, the selected chirality is accepted only if its score exceeds the empirical threshold:
$\max (0.5 \times \sum_{i\in\mathcal{N}}{\mathrm{CP}_{i}}, 0.001)$.

\subsubsection{Yaw Estimation}
\label{Sec: SFC: Yaw Estimation}

If the gravity is available, we can estimate the Yaw angle of the robots.
We have the positions ${^{F_u}\mathbf{p}_{R_i}}$ and the gravity $(0,0,1)$.
By aligning the gravity and matching the bearings, we can calculate the Yaw angle of each robot.
The main idea in this part is the same as the Sec. III-C in Xun's work \cite{xun2023crepes}, however, instead of using dual bearing measurements ${{\hat{\mathbf{z}}_b}{}_{B_i \to M_j}}$ and ${{\hat{\mathbf{z}}_b}{}_{B_j \to M_i}}$, we use the bearing measurements ${{\hat{\mathbf{z}}_b}{}_{B_i \to M_j}}$ and the position difference ${^{F_u}\mathbf{p}_{R_i \to R_j}}$ to calculate the Yaw angle.
This approach mitigates the strong dependency on LOS conditions, enabling yaw estimation in the presence of NLOS configurations.

First, we introduce an intermediate coordinate frame $R_i'$, which retains only the Yaw component of ${R_i}$ and aligns the gravity direction to the $z$-axis in $F_u$.
We find a rotation matrix ${^{R_i'}\hat{\mathbf{R}}_{R_i}}$ to perform this alignment:
\begin{equation}
    (0,0,1)^T = {^{R_i'}\hat{\mathbf{R}}_{R_i}} \, {\hat{\mathbf{z}}_g}{}_{R_i}.
\end{equation}

Then we express each bearing measurements in frame $R_i'$, and project the rotated bearing measurements in $R_i'$ and the difference of the positions in $F_u$ to the XY plane, and calculate the angle $\hat{\psi}_{i,j}$ between $R_i'$ and $F_u$:
\begin{equation}
    \begin{aligned}
        \sin(\hat{\psi}_{i,j}) & = ({^{R_i'}\hat{\mathbf{R}}_{R_i}} {^{R_i}\mathbf{R}_{B_i}} {{\hat{\mathbf{z}}_b}{}_{B_i \to M_j}})_{xy} \times ({^{F_u}\tilde{\mathbf{p}}_{R_i \to R_j}})_{xy},
        \\
        \cos(\hat{\psi}_{i,j}) & = ({^{R_i'}\hat{\mathbf{R}}_{R_i}} {^{R_i}\mathbf{R}_{B_i}} {{\hat{\mathbf{z}}_b}{}_{B_i \to M_j}})_{xy} \cdot ({^{F_u}\tilde{\mathbf{p}}_{R_i \to R_j}})_{xy}.
    \end{aligned}
\end{equation}

Then for robot $i$ and its bearings $j \in \mathcal{B}_i$, we can calculate the mean of the angles $\hat{\psi}_{i,j}$ as the Yaw angle of the robot $i$:
\begin{equation}
    \hat{\psi}_{i} = \frac{1}{|\mathcal{B}_i|} \sum_{j \in \mathcal{B}_i} \hat{\psi}_{i,j}.
\end{equation}

\subsubsection{Rotation Estimation}
\label{Sec: SFC: Rotation Estimation}

If gravity is available, the rotation matrix ${^{F_u}\hat{\mathbf{R}}_{R_i}}$ is calculated using the Yaw angle $\hat{\psi}_{i}$:
\begin{equation}
    \begin{aligned}
        {^{F_u}\hat{\mathbf{R}}_{R_i'}} & =    \begin{bmatrix}
                                                   \cos(\hat{\psi}_{i}) & -\sin(\hat{\psi}_{i}) & 0 \\
                                                   \sin(\hat{\psi}_{i}) & \cos(\hat{\psi}_{i})  & 0 \\
                                                   0                    & 0                     & 1
                                               \end{bmatrix},
        \\
        {^{F_u}\hat{\mathbf{R}}_{R_i}}  & = {^{F_u}\hat{\mathbf{R}}_{R_i'}} {^{R_i'}\hat{\mathbf{R}}_{R_i}}.
    \end{aligned}
\end{equation}

If gravity is unavailable, the positions ${^{F_u}\mathbf{p}_{R_j}}$ are directly used to calculate the rotation of robot $i$ in $F_u$ by solving the Wahba's problem \eqref{Equ: SFC: Wahba Solution Correction} to obtain the rotation matrix ${^{F_u}\hat{\mathbf{R}}_{R_i}}$.

\subsubsection{Relative Pose Extraction}
\label{Sec: SFC: Relative Pose Extraction}

We have obtained the estimated poses ${^{F_u}\hat{\mathbf{R}}_{R_i}}$ and ${^{F_u}\mathbf{p}_{R_i}}$ of all robots $i$ in the frame $F_u$.
However, since $F_u$ is an arbitrary frame, these poses are not meaningful for multi-robot applications.
Therefore, we transform them from $F_u$ to the chosen reference robot frame $R_r$ to obtain relative poses.
To enhance robustness, we first evaluate the consistency between the estimated relative poses and the original measurements.
Measurements exhibiting large errors are rejected as outliers.
For each robot $i$, once it accumulates at least two inlier bearing and gravity measurements, its rotation becomes fully constrained and is marked as observable.
The observability of the reference robot's rotation (i.e., $R_r$) is a prerequisite, since the transformation from $F_u$ to $R_r$ requires its rotation.
Additionally, all positions derived from the MDS are marked as observable by default.

A relative pose is extracted only when both the rotation and position estimates of robot $i$ and the reference robot are observable.
The SFC's result relative position ${^{R_r}\hat{\mathbf{p}}_{R_i}}$ and rotation ${^{R_r}\hat{\mathbf{R}}_{R_i}}$ of robot $i$ with respect to the reference robot $R_r$ are computed as:
\begin{equation}
    \begin{aligned}
        {^{R_r}\hat{\mathbf{p}}_{R_i}} & = {^{R_r}\hat{\mathbf{R}}_{F_u}}({^{F_u}\mathbf{p}_{R_i}}-{^{F_u}\mathbf{p}_{R_r}}),
        \\
        {^{R_r}\hat{\mathbf{R}}_{R_i}} & = {^{R_r}\hat{\mathbf{R}}_{F_u}}{^{F_u}\hat{\mathbf{R}}_{R_i}}.
    \end{aligned}
    \label{Equ: SFC: Pose Estimation}
\end{equation}

\subsection{Single-Frame Optimization (SFO)}
\label{Sec: Single-Frame Optimization}

While the SFC provides an instantaneous estimate, it makes simplifying assumptions (e.g., ignoring translational extrinsics) and does not optimally weight measurements according to their noise characteristics.
Following the principle of maximum likelihood estimation, SFO refines the poses by solving a nonlinear least-squares optimization problem that incorporates \eqref{Equ: Distance Residuals}, \eqref{Equ: Bearing Residuals}, and \eqref{Equ: Gravity Residuals}.
The problem is formulated as:

\begin{equation}
    \begin{aligned}
        \min_{\mathfrak{X}^{t}_{R_{\mathcal{N}}}}
        \Big\{
          & \sum_{j \in \mathcal{N}}\sum_{k \in \mathcal{D}^{t}_j} {w_d}^{t}_{j,k} \| \underbrace{{r}_d ({\hat{z}_d}{}^{t}_{D_j \to D_k}, \mathfrak{X}^{t}_{R_{\mathcal{N}}})}_{\text{distance residual}} \|_{\bm{\Sigma}_d}^2
        \\
        + & \sum_{j \in \mathcal{N}}\sum_{k \in \mathcal{B}^{t}_j} {w_b}^{t}_{j,k} \| \underbrace{\mathbf{r}_b ({\hat{\mathbf{z}}_b}{}^{t}_{B_j \to M_k}, \mathfrak{X}^{t}_{R_{\mathcal{N}}})}_{\text{bearing residual}} \|_{\bm{\Sigma}_b}^2
        \\
        + & \; {w_g} \sum_{j \in \mathcal{N}} \| \underbrace{\mathbf{r}_g ({\hat{\mathbf{z}}_g}{}^{t}_{R_j}, \mathfrak{X}^{t}_{R_{\mathcal{N}}}, {^{R_r}\mathbf{g}^{t}})}_{\text{gravity residual}} \|_{\bm{\Sigma}_g}^2 \Big\}.
        \label{Equ: SFO: Pose Optimization}
    \end{aligned}
\end{equation}
All optimization problems (\eqref{Equ: SFO: Pose Optimization}, \eqref{Equ: MFLO: Pose Optimization}, and \eqref{Equ: MFTO: Pose Optimization}) are solved using the Ceres Solver \cite{solver1ceres} with the Levenberg-Marquardt algorithm.
Rotations are parameterized on the quaternion manifold, and the gravity is parameterized on the unit spherical manifold.
Huber loss \cite{huber1992robust} is used to reduce the impact of bearing outliers, while the half-Cauchy model in \eqref{Equ: Distance NLL Asymmetric} is used to handle the heavy-tailed distance errors.
${w_d}^{t}_{j,k}$ are set to $1$, ${w_b}^{t}_{j,k}$ are determined by outlier rejection in \autoref{Sec: Outlier Rejection}.

It is worth noting that the availability of relative poses is determined following SFC's judgement, which is more robust than re-evaluating at SFO based on directional inlier counts.

\section{Multi-Frame Relative Estimator}
\label{Sec: Multi-Frame Relative Estimator}

SFRE provides one-shot estimation of the relative poses.
However, the results are sensitive to noise and outliers, especially in severe NLOS conditions.
To improve the robustness and accuracy, we add IMU measurements to provide inertial constraints for smoothing.

\subsection{Robocentric Relative Kinematics}
\label{Sec: Robocentric Relative Kinematics}

\begin{figure}[t]
    \centering
    \includegraphics[width=.8\halfwidth]{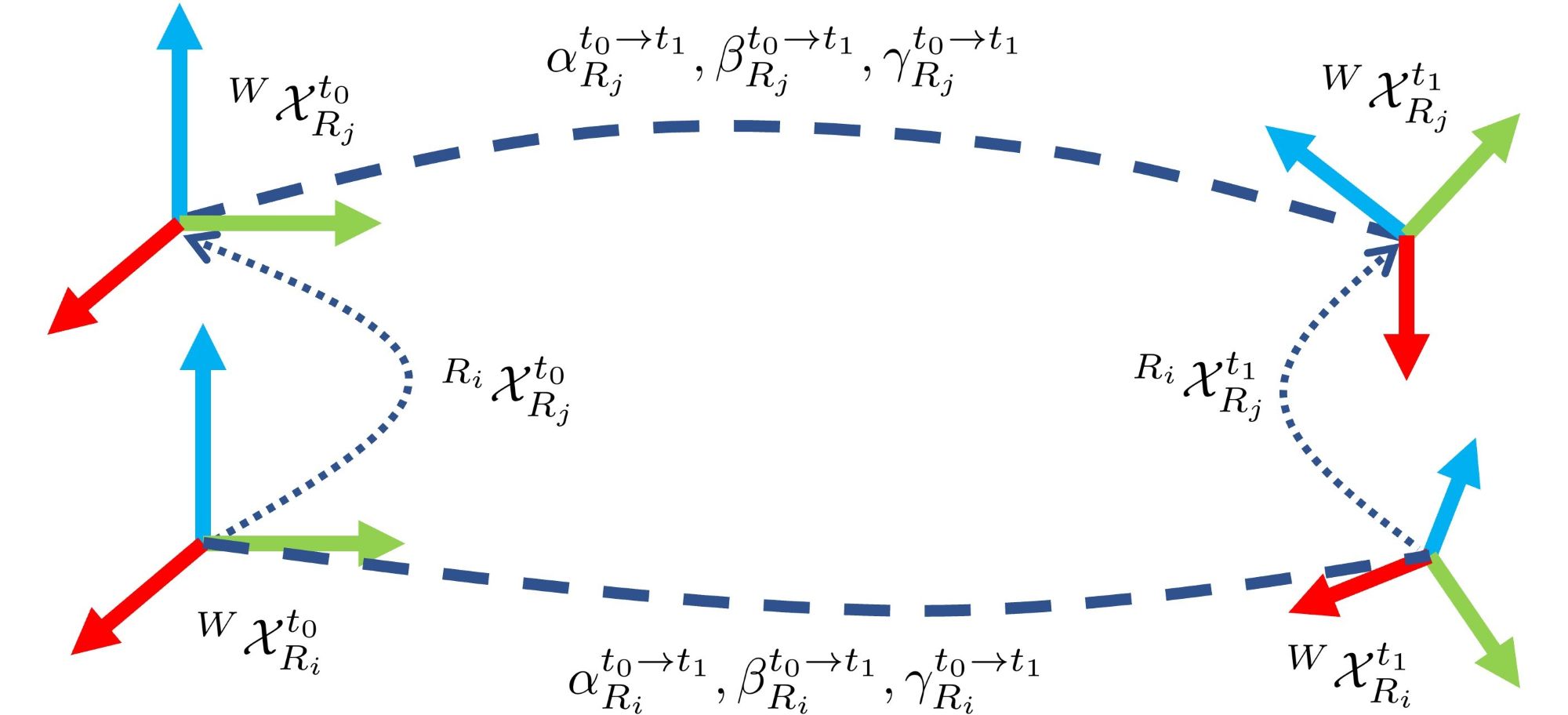}
    \caption{
        Relative kinematics of two robots. Both preintegrations are used to derive the motion of robot $j$ relative to robot $i$ from $t_0$ to $t_1$.
    }
    \label{Fig: IMU Residuals}
\end{figure}

To enable estimation in non-inertial environments and eliminate the need for a globally consistent gravity vector, we derive the kinematics in a robocentric frame.
The ego-motion of any robot $i$ in the world frame (used here for auxiliary derivation and replaceable be any inertial reference frame) can be calculated as:
\begin{equation}
    \begin{aligned}
        {^{W}\mathbf{p}_{R_i}^{t_1}} & =
        {^{W}\mathbf{p}_{R_i}^{t_0}}+{^{W}\mathbf{v}_{R_i}^{t_0}}\Delta t-\textstyle\iint\mathbf{g}~dt^2+{^{W}\mathbf{R}_{R_i}^{t_0}}\bm{\alpha}^{t_0 \to t_1}_{R_i},
        \\
        {^{W}\mathbf{v}_{R_i}^{t_1}} & =
        {^{W}\mathbf{v}_{R_i}^{t_0}}-\textstyle\int\mathbf{g}~dt+{^{W}\mathbf{R}_{R_i}^{t_0}}\bm{\beta}^{t_0 \to t_1}_{R_i},
        \\
        {^{W}\mathbf{q}_{R_i}^{t_1}} & =
        {^{W}\mathbf{q}_{R_i}^{t_0}}\otimes\bm{\gamma}^{t_0 \to t_1}_{R_i}.
    \end{aligned}
\end{equation}
Using the transformation of the states of robot $i$ and robot $j$ at time $t_0$ and $t_1$, the relative kinematics of robot $j$ in robot $i$'s frame can be calculated as:
\begin{equation}
    \begin{aligned}
        {^{R_i}\mathbf{p}_{R_j}^{t_1}} & = ({^{W}\mathbf{R}_{R_i}^{t_1}})^{-1}({^{W}\mathbf{p}_{R_j}^{t_1}}-{^{W}\mathbf{p}_{R_i}^{t_1}})
        \\
                                       & = \mathbf{R}^{-1}\{\bm{\gamma}^{t_0 \to t_1}_{R_i}\}
        \\
                                       & \qquad ({^{R_i}\mathbf{p}_{R_j}^{t_0}}+{^{R_i}\mathbf{v}_{R_j}^{t_0}}\Delta t + {^{R_i}\mathbf{R}_{R_j}^{t_0}}\bm{\alpha}_{R_j}^{t_0 \to t_1}-\bm{\alpha}_{R_i}^{t_0 \to t_1}),
        \\
        {^{R_i}\mathbf{v}_{R_j}^{t_1}} & = ({^{W}\mathbf{R}_{R_i}^{t_1}})^{-1}({^{W}\mathbf{v}_{R_j}^{t_1}}-{^{W}\mathbf{v}_{R_i}^{t_1}})
        \\
                                       & = \mathbf{R}^{-1}\{\bm{\gamma}^{t_0 \to t_1}_{R_i}\} ({^{R_i}\mathbf{v}_{R_j}^{t_0}} + {^{R_i}\mathbf{R}_{R_j}^{t_0}}{\bm{\beta}^{t_0 \to t_1}_{R_j}}-{\bm{\beta}^{t_0 \to t_1}_{R_i}}),
        \\
        {^{R_i}\mathbf{q}_{R_j}^{t_1}} & = ({^{W}\mathbf{q}_{R_i}^{t_1}})^{-1}\otimes{^{W}\mathbf{q}_{R_j}^{t_1}}
        \\
                                       & = (\bm{\gamma}_{R_i}^{t_0 \to t_1})^{-1}\otimes({^{R_i}\mathbf{q}_{R_j}^{t_0}}\otimes\bm{\gamma}_{R_j}^{t_0 \to t_1}).
    \end{aligned}
    \label{Equ: Relative Kinematics}
\end{equation}
This preintegration-based relative kinematics is shown in \autoref{Fig: IMU Residuals}.
Note that ${^{R_i}\mathbf{v}_{R_j}^{t_i}}$ is defined as the rotation-projected relative translational velocity ${^{R_i}\mathbf{v}_{R_j}^{t_i}} \triangleq {^{R_i}\mathbf{R}_{W}^{t_i}}{^{W}\mathbf{v}_{R_i \to R_j}^{t_i}}$, which does not contain the relative velocity generated by the rotation of the reference frame $R_r$.
Also, note that gravity disappears in the calculation, so there is no need to estimate the direction and magnitude of gravity.
Moreover, this allows our system to work in any non-inertial environment, such as space or moving platforms, as long as all robots experience the same gravitational force.

\subsection{Multi-Frame Loosely-Coupled Optimization (MFLO)}
\label{Sec: Multi-Frame Loosely-Coupled Optimization}

The MFLO serves two critical functions:
First, it temporally smooths the per-frame SFRE estimates by incorporating inertial constraints;
Second, it efficiently computes a high-quality initial guess for all states in the sliding window.
This robust initialization is crucial for ensuring the fast and reliable convergence of the final tightly-coupled optimization stage.

To improve computational efficiency, we optimize only the full state of the first frame $\mathfrak{X}^{t_0}_{R_{\mathcal{N}}}$, during this stage.
Referring to \eqref{Equ: Relative Kinematics}, the inertial constraints are formulated as:
\begin{equation}
    \begin{aligned}
         & \mathbf{r}_{i_p} ({\hat{\mathbf{z}}_i}{}^{t_0 \to t_i}_{R_r}, {\hat{\mathbf{z}}_i}{}^{t_0 \to t_i}_{R_j}, {^{R_r}\hat{\mathbf{p}}_{R_j}^{t_i}}, \mathfrak{X}^{t_0}_{R_{\mathcal{N}}})
        = \delta {^{R_r}\mathbf{p}_{R_j}^{t_0 \to t_i}}
        \\
         & \quad = \mathbf{R}\{\hat{\bm{\gamma}}^{t_0 \to t_i}_{R_r}\}{^{R_r}\hat{\mathbf{p}}_{R_j}^{t_i}}-
        \\
         & \qquad\quad ({^{R_r}\mathbf{p}_{R_j}^{t_0}}+{^{R_r}\mathbf{v}_{R_j}^{t_0}}\Delta t-{\hat{\bm{\alpha}}^{t_0 \to t_i}_{R_r}}+{^{R_r}\mathbf{R}_{R_j}^{t_0}}{\hat{\bm{\alpha}}^{t_0 \to t_i}_{R_j}}),
        \\
         & \mathbf{r}_{i_q} ({\hat{\mathbf{z}}_i}{}^{t_0 \to t_i}_{R_r}, {\hat{\mathbf{z}}_i}{}^{t_0 \to t_i}_{R_j}, {^{R_r}\hat{\mathbf{q}}_{R_j}^{t_i}}, \mathfrak{X}^{t_0}_{R_{\mathcal{N}}}) = \delta {^{R_r}\mathbf{q}_{R_j}^{t_0 \to t_i}}
        \\
         & \quad = \mathbf{Log}((\hat{\bm{\gamma}}^{t_0 \to t_i}_{R_j})^{-1}({^{R_r}\mathbf{q}_{R_j}^{t_0}})^{-1}\hat{\bm{\gamma}}^{t_0 \to t_i}_{R_r}{^{R_r}\hat{\mathbf{q}}_{R_j}^{t_i}}).
        \label{Equ: MFLO: Inertial Constraints}
    \end{aligned}
\end{equation}

The target position-rotation-inertial(-gravity) optimization problem with frames $\mathfrak{F}^{t_i}, i \in \mathcal{M}$ using \eqref{Equ: MFLO: Inertial Constraints} is formulated as:
\begin{equation}
    \begin{aligned}
        \min_{\mathfrak{X}^{t_0}_{R_{\mathcal{N}}}}
        \Bigl\{
          & \sum_{i \in \mathcal{M}}\sum_{j \in \mathcal{N}} {w_p} \| \underbrace{\mathbf{r}_{i_p} ({\hat{\mathbf{z}}_i}{}^{t_0 \to t_i}_{R_r}, {\hat{\mathbf{z}}_i}{}^{t_0 \to t_i}_{R_j}, {^{R_r}\hat{\mathbf{p}}_{R_j}^{t_i}}, \mathfrak{X}^{t_0}_{R_{\mathcal{N}}})}_{\text{relative integration to position residual}} \|_{\bm{\Sigma}_p}^2
        \\
        + & \sum_{i \in \mathcal{M}}\sum_{j \in \mathcal{N}} {w_q} \| \underbrace{\mathbf{r}_{i_q} ({\hat{\mathbf{z}}_i}{}^{t_0 \to t_i}_{R_r}, {\hat{\mathbf{z}}_i}{}^{t_0 \to t_i}_{R_j}, {^{R_r}\hat{\mathbf{q}}_{R_j}^{t_i}}, \mathfrak{X}^{t_0}_{R_{\mathcal{N}}})}_{\text{relative integration to rotation residual}} \|_{\bm{\Sigma}_q}^2
        \\
        + & \; {w_g} \sum_{j \in \mathcal{N}} \| \underbrace{\mathbf{r}_g ({\hat{\mathbf{z}}_g}{}^{t_0}_{R_j}, \mathfrak{X}^{t_0}_{R_{\mathcal{N}}}, {^{R_r}\mathbf{g}^{t_0}})}_{\text{gravity residual}} \|_{\bm{\Sigma}_g}^2
        \Bigr\}.
        \label{Equ: MFLO: Pose Optimization}
    \end{aligned}
\end{equation}

Note that the $\mathbf{r}_{i_p}$ is added for all $i,j$ since the positions are always observable, while the $\mathbf{r}_{i_q}$ is added only when the rotation is marked as observable after \autoref{Sec: SFC: Relative Pose Extraction} for robot $R_j$ at frame $i$.
The $\bm{\Sigma}_p$ and $\bm{\Sigma}_q$ are set by practical experience, which are $0.1$ and $0.01$, respectively.
In this way, we can utilize all frames $\mathfrak{F}_i$ in the sliding window, rather than only keyframes $\mathfrak{KF}_i$.
This provides many more constraints for the optimization; even if only a few results are generated by SFRE, the optimization can still be solved.
Moreover, decoupling different robot states yields a sparse Jacobian matrix, enhancing its efficiency.
After optimization, states in the time window are propagated via relative kinematics.

\subsection{Multi-Frame Tightly-Coupled Optimization (MFTO)}
\label{Sec: Multi-Frame Tightly-Coupled Optimization}

MFTO builds the tightly-coupled bearing-distance-inertial(-gravity) optimization problem in the sliding window using the keyframes $\mathfrak{KF}_i, i \in \mathcal{M}_k$ and the latest frame $\mathfrak{F}$.
We directly use SFRE to quantify the informativeness of the current frame.
When a new frame arrives, whether it is a keyframe is decided by the following rules:
\begin{itemize}
    \item {If the time since the last keyframe exceeds $t_{\text{min}}=100\,\text{ms}$ and SFRE produces a result.}
    \item {If the time since the last keyframe exceeds $t_{\text{max}}=200\,\text{ms}$}.
\end{itemize}

\begin{figure*}[t]
    \centering
    \includegraphics[width=\fullwidth]{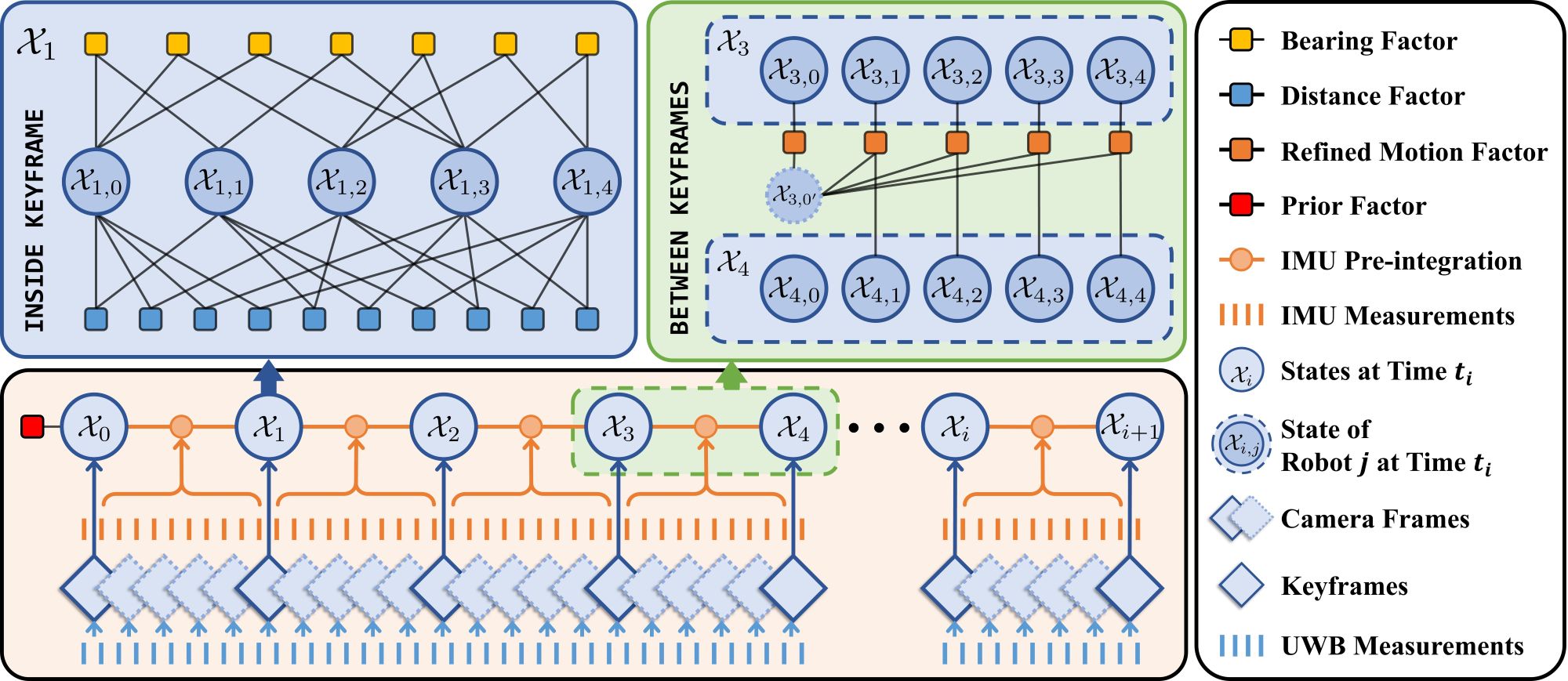}
    \caption{
        Demonstration of the state graph in multi-frame tightly-coupled optimization.
        Inside one Keyframe, the states are connected by bearing and distance factors from multiple neighbors.
        Between Keyframes, the states are connected by relative inertial factors, which are calculated with IMU preintegration.
    }
    \label{Fig: MFTO: State Graph of Optimization}
\end{figure*}

\subsubsection{Refined Inertial Residuals with Auxiliary State}
\label{Sec: MFTO: Refined Inertial Residuals}

In \eqref{Equ: MFLO: Inertial Constraints}, inertial residuals are constructed using two preintegrations to minimize optimization parameters and accelerate multi-frame initialization.
However, this approach neglects the covariance of IMU preintegration, which is crucial for least-squares optimization as it ensures maximum likelihood estimation.
Decoupling the state and preintegration within a residual is challenging, and without this decoupling, the IMU preintegration covariance cannot be accurately incorporated into the optimization's information matrix.
To correctly construct the Mahalanobis distance using the covariance of preintegration, we add an auxiliary state $\mathfrak{X}^{t_i}_{R_r'}$ between $\mathfrak{X}^{t_i}_{R_r}$ and $\mathfrak{X}^{t_j}_{R_r}$, which is defined as:
\begin{equation}
    \begin{aligned}
        \mathfrak{X}^{t_i}_{R_r'}                     & = \left[ {^{R_r}\mathbf{p}_{R_r'}^{t_i}}, {^{R_r}\mathbf{v}_{R_r'}^{t_i}}, {^{R_r}\mathbf{q}_{R_r'}^{t_i}},{^{R_r}\mathbf{b}_{\mathbf{a}}}_{R_r'}^{t_i},{^{R_r}\mathbf{b}_{\bm{\omega}}}_{R_r'}^{t_i} \right],
        \\
        {^{R_r}\mathbf{b}_{\mathbf{a}}}_{R_r'}^{t_i}  & = {^{R_r}\mathbf{b}_{\mathbf{a}}}_{R_r}^{t_j},
        \\
        {^{R_r}\mathbf{b}_{\bm{\omega}}}_{R_r'}^{t_i} & = {^{R_r}\mathbf{b}_{\bm{\omega}}}_{R_r}^{t_j}.
    \end{aligned}
\end{equation}

Then the residuals can be divided into two parts, the reference one using $R_r$'s IMU measurements:
\begin{equation}
    \mathbf{r}_{i_{R_r}} (\hat{\mathbf{z}_i}^{t_i \to t_j}_{R_r}, \mathfrak{X}^{t_{\mathcal{M}_k'}}_{R_{\mathcal{N}}}) =
    \begin{bmatrix}
        \begin{aligned}
             & \delta {\bm{\alpha}_{R_r}^{t_i \to t_j}}              \\
             & \delta {\bm{\beta}_{R_r}^{t_i \to t_j}}               \\
             & \delta {\bm{\gamma}_{R_r}^{t_i \to t_j}}              \\
             & \delta {\mathbf{b}_{\mathbf{a}}}^{t_i \to t_j}_{R_r}  \\
             & \delta {\mathbf{b}_{\bm{\omega}}}^{t_i \to t_j}_{R_r} \\
        \end{aligned}
    \end{bmatrix},
    \label{Equ: MFTO: Inertial Residual RF}
\end{equation}
\begin{equation}
    \begin{aligned}
        \delta {\bm{\alpha}_{R_r}^{t_i \to t_j}}=
         & {^{R_r}\mathbf{p}_{R_r'}^{t_i}}-{\hat{\bm{\alpha}}^{t_i \to t_j}_{R_r}},
        \\
        \delta {\bm{\beta}_{R_r}^{t_i \to t_j}}=
         & {^{R_r}\mathbf{v}_{R_r'}^{t_i}}-{\hat{\bm{\beta}}^{t_i \to t_j}_{R_r}},
        \\
        \delta {\bm{\gamma}_{R_r}^{t_i \to t_j}}=
         & \mathbf{Log}({^{R_r}\mathbf{q}_{R_r'}^{t_i}}(\hat{\bm{\gamma}}^{t_i \to t_j}_{R_r})^{-1}),
        \\
        \delta {\mathbf{b}_{\mathbf{a}}}^{t_i \to t_j}_{R_r}
         & = {^{R_r}\mathbf{b}_{\mathbf{a}}}^{t_j}_{R_r}-{^{R_r}\mathbf{b}_{\mathbf{a}}}^{t_i}_{R_r},
        \\
        \delta {\mathbf{b}_{\bm{\omega}}}^{t_i \to t_j}_{R_r}
         & = {^{R_r}\mathbf{b}_{\bm{\omega}}}^{t_j}_{R_r}-{^{R_r}\mathbf{b}_{\bm{\omega}}}^{t_i}_{R_r},
    \end{aligned}
\end{equation}
and the other one using target robot $j$'s IMU measurements:
\begin{equation}
    \mathbf{r}_{i} (\hat{\mathbf{z}_i}^{t_i \to t_j}_{R_k}, \mathfrak{X}^{t_{\mathcal{M}_k'}}_{R_{\mathcal{N}}}) =
    \begin{bmatrix}
        \begin{aligned}
             & \delta {\bm{\alpha}_{R_k}^{t_i \to t_j}}              \\
             & \delta {\bm{\beta}_{R_k}^{t_i \to t_j}}               \\
             & \delta {\bm{\gamma}_{R_k}^{t_i \to t_j}}              \\
             & \delta {\mathbf{b}_{\mathbf{a}}}^{t_i \to t_j}_{R_k}  \\
             & \delta {\mathbf{b}_{\bm{\omega}}}^{t_i \to t_j}_{R_k} \\
        \end{aligned}
    \end{bmatrix},
    \label{Equ: MFTO: Inertial Residual TF}
\end{equation}
\begin{equation}
    \begin{aligned}
        \delta {\bm{\alpha}_{R_k}^{t_i \to t_j}}=
         & ({^{R_r}\mathbf{R}_{R_k}^{t_i}})^{-1}(\mathbf{R}\{{^{R_r}\mathbf{q}_{R_r'}^{t_i}}\}{^{R_r}\mathbf{p}_{R_k}^{t_j}}-
        \\
         & ({^{R_r}\mathbf{p}_{R_k}^{t_i}}+{^{R_r}\mathbf{v}_{R_k}^{t_i}}\Delta t-{^{R_r}\mathbf{p}_{R_r'}^{t_i}}))-{\hat{\bm{\alpha}}^{t_i \to t_j}_{R_k}},
        \\
        \delta {\bm{\beta}_{R_k}^{t_i \to t_j}}=
         & ({^{R_r}\mathbf{R}_{R_k}^{t_i}})^{-1}(\mathbf{R}\{{^{R_r}\mathbf{q}_{R_r'}^{t_i}}\}{^{R_r}\mathbf{v}_{R_k}^{t_j}}-
        \\
         & ({^{R_r}\mathbf{v}_{R_k}^{t_i}}-{^{R_r}\mathbf{v}_{R_r'}^{t_i}}))-{\hat{\bm{\beta}}^{t_i \to t_j}_{R_k}},
        \\
        \delta {\bm{\gamma}_{R_k}^{t_i \to t_j}}=
         & \mathbf{Log}(({^{R_r}\mathbf{q}_{R_k}^{t_i}})^{-1}{^{R_r}\mathbf{q}_{R_r'}^{t_i}}{^{R_r}\mathbf{q}_{R_k}^{t_j}}(\hat{\bm{\gamma}}^{t_i \to t_j}_{R_k})^{-1}),
        \\
        \delta {\mathbf{b}_{\mathbf{a}}}^{t_i \to t_j}_{R_k}
         & =  {^{R_k}\mathbf{b}_{\mathbf{a}}}^{t_j}_{R_k}-{^{R_k}\mathbf{b}_{\mathbf{a}}}^{t_i}_{R_k},
        \\
        \delta {\mathbf{b}_{\bm{\omega}}}^{t_i \to t_j}_{R_k}
         & = {^{R_k}\mathbf{b}_{\bm{\omega}}}^{t_j}_{R_k}-{^{R_k}\mathbf{b}_{\bm{\omega}}}^{t_i}_{R_k}.
    \end{aligned}
\end{equation}
This process is shown in \autoref{Fig: MFTO: State Graph of Optimization} (between keyframes).

\subsubsection{Marginalization}
\label{Sec: MFTO: Marginalization}

Marginalization is a classical technique in sliding window optimization, used to retain the constraints of specific factors while removing them from the optimization problem.
In our system, as no landmarks are involved, the marginalization process simplifies into forming prior constraints on the earliest states $\mathfrak{X}^{t_0}_{R_{\mathcal{N}}}$ in the sliding window.
This process is implemented using the Schur complement \cite{sibley2010sliding}.
As the window slides, the oldest states and factors in the window are removed during the next optimization.
To preserve their information, we construct the prior information $\{ \mathbf{r}_m, \mathbf{H}_m \}$ by combining marginalized factors associated with the first states and the existing prior from the last optimization.

\subsubsection{Problem Formulation}
\label{Sec: MFTO: Problem Formulation}

Finally, we construct the tightly-coupled bearing-distance-inertial(-gravity) optimization problem with keyframes $\mathfrak{KF}_i, i \in \mathcal{M}_k$ and the latest frame $\mathfrak{F}$ (together as $\mathcal{M}_k'$) using \eqref{Equ: Distance Residuals}, \eqref{Equ: Bearing Residuals}, \eqref{Equ: Gravity Residuals}, \eqref{Equ: MFTO: Inertial Residual RF}, \eqref{Equ: MFTO: Inertial Residual TF}, and the marginalization residual.
\begin{equation}
    \begin{aligned}
        \min_{\mathfrak{X}^{t_{\mathcal{M}_k'}}_{R_{\mathcal{N}}}}
        \Big\{
          & \sum_{i \in \mathcal{M}_k'} \| \underbrace{\mathbf{r}_{i_{R_r}} (\hat{\mathbf{z}_i}^{t_i \to t_{i+1}}_{R_r}, \mathfrak{X}^{t_{\mathcal{M}_k'}}_{R_{\mathcal{N}}})}_{\text{reference inertial residual}} \|_{\bm{\Sigma}_i}^2
        \\
        + & \sum_{i \in \mathcal{M}_k'}\sum_{j \in \mathcal{N}} \| \underbrace{\mathbf{r}_{i} (\hat{\mathbf{z}_i}^{t_i \to t_{i+1}}_{R_j}, \mathfrak{X}^{t_{\mathcal{M}_k'}}_{R_{\mathcal{N}}})}_{\text{target inertial residual}} \|_{\bm{\Sigma}_i}^2
        \\
        + & \sum_{i \in \mathcal{M}_k'}\sum_{j \in \mathcal{N}}\sum_{k \in \mathcal{D}^{t_i}_j} {w_d}^{t_i}_{j,k} \| \underbrace{{r}_d ({\hat{z}_d}{}^{t_i}_{D_j \to D_k}, \mathfrak{X}^{t_{\mathcal{M}_k'}}_{R_{\mathcal{N}}})}_{\text{distance residual}} \|_{\bm{\Sigma}_d}^2
        \\
        + & \sum_{i \in \mathcal{M}_k'}\sum_{j \in \mathcal{N}}\sum_{k \in \mathcal{B}^{t_i}_j} {w_b}^{t_i}_{j,k} \| \underbrace{\mathbf{r}_b ({\hat{\mathbf{z}}_b}{}^{t_i}_{B_j \to M_k}, \mathfrak{X}^{t_{\mathcal{M}_k'}}_{R_{\mathcal{N}}})}_{\text{bearing residual}} \|_{\bm{\Sigma}_b}^2
        \\
        + & {w_g} \sum_{i \in \mathcal{M}_k'}\sum_{j \in \mathcal{N}} \| \underbrace{\mathbf{r}_g ({\hat{\mathbf{z}}_g}{}^{t_i}_{R_j}, \mathfrak{X}^{t_{\mathcal{M}_k'}}_{R_{\mathcal{N}}}, {^{R_r}\mathbf{g}^{t_i}})}_{\text{gravity residual}} \|_{\bm{\Sigma}_g}^2
        \\
        + & \| \underbrace{\mathbf{r}_m - \mathbf{H}_m \mathfrak{X}^{t_0}_{R_{\mathcal{N}}}}_{\text{marginalization}} \|^2
        \Big\}.
        \label{Equ: MFTO: Pose Optimization}
    \end{aligned}
\end{equation}
MFLO and MFTO can run upon each new frame, yielding a maximum output frequency equal to the camera's.

\section{Outlier Rejection}
\label{Sec: Outlier Rejection}

Outliers in observations can greatly reduce estimation accuracy when being fed into closed-form solvers or optimizations.
To build a robust system, it is important to remove outliers from the measurements.
Gravity and distance measurements are generally reliable, but bearings are more likely to have outliers due to infrared interference and reflections.
Effective outlier rejection for bearing measurements presents a circular challenge: the bearings are required to estimate rotation and resolve positional chirality, yet a robust estimation first requires the outliers to be removed.

\begin{figure}[t]
    \centering
    \includegraphics[width=\halfwidth]{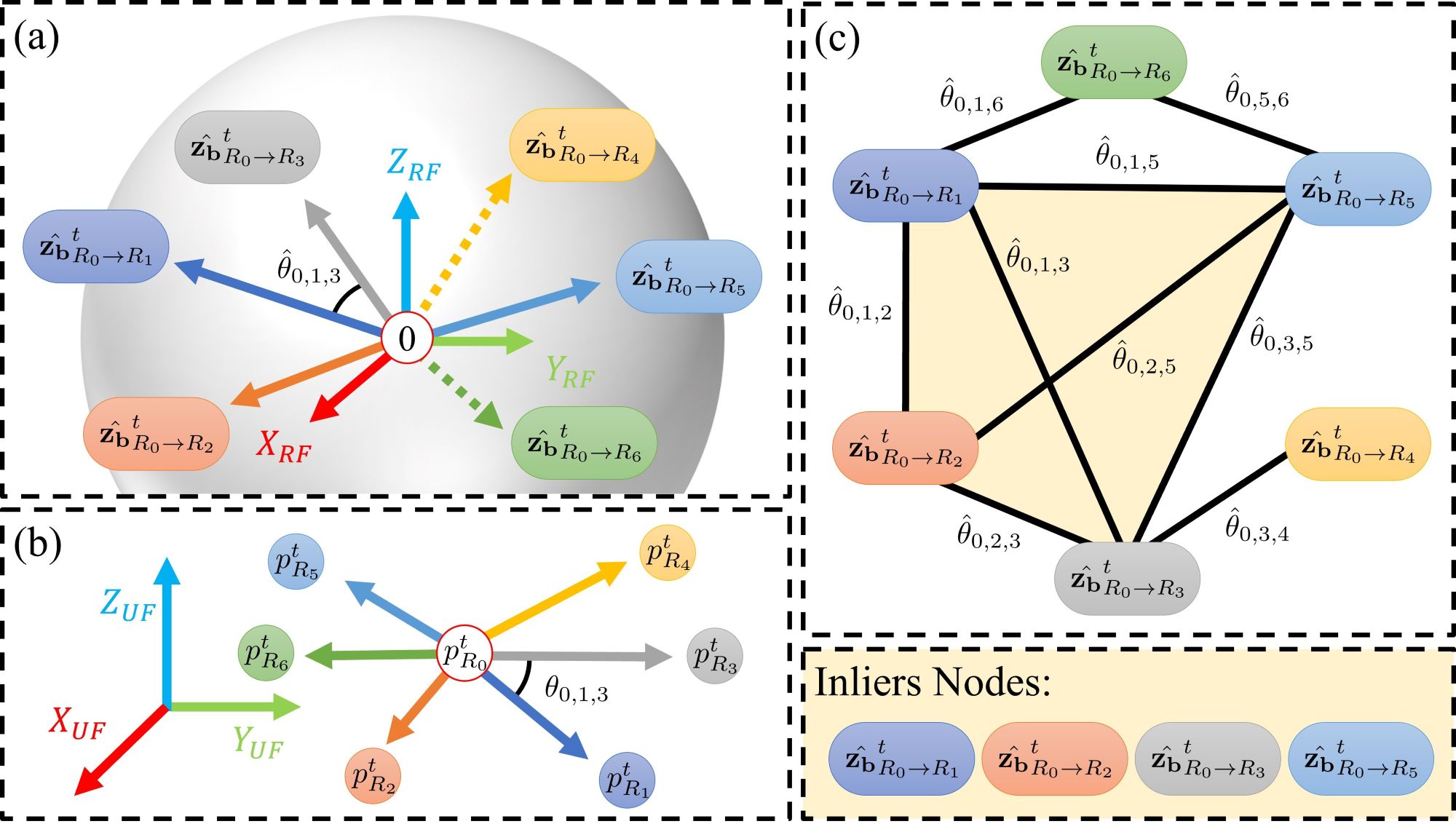}
    \caption{
        Outlier rejection.
        (a) Measured bearings ${{\hat{\mathbf{z}}_b}{}_{B_i \to M_j}}$ yield angles ${\hat{\theta}_{i,j,k}}$.
        (b) Positions ${^{F_u}\mathbf{p}_{R_j}}$ obtained from MDS provide angles ${\theta_{i,j,k}}$, which coincide with (a) in the noise-free case.
        (c) Each measurement is represented as a node, and edges are established between nodes when their angles are sufficiently close.
        Nodes in the maximal clique are identified as inliers.
    }
    \label{Fig: Outlier Rejection}
\end{figure}

To break this dependency, we adopt PCM \cite{mangelson2018pairwise} and propose a novel, state-free outlier rejection scheme that operates before any pose information is known.
Since PCM is originally designed for PGO, applying it to bearing measurements necessitates specific modifications, which we refer to as PCM-B.
Our method hinges on a key geometric invariant: while individual bearing vectors are dependent on the observer's rotation, the angle between two bearing vectors is not.
This angle should remain consistent with the angle formed by the corresponding relative position vectors, regardless of rotation or reflection (chirality), as illustrated in \autoref{Fig: Outlier Rejection} (a).

First, we use MDS positions from \eqref{Equ: SFC: MDS Solution} as the reference, with bearings as the measurements to be verified.
Consider bearings of robot $i$ to robot $j$ and robot $k$, the angle ${\hat{\theta}_{i,j,k}}$ between the bearing $j$ and $k$ is calculated as:
\begin{equation}
    {\hat{\theta}_{i,j,k}} = \cos^{-1} \left( {{\hat{\mathbf{z}}_b}{}_{B_i \to M_j}} \cdot {{\hat{\mathbf{z}}_b}{}_{B_i \to M_k}}\right),
\end{equation}
where $\hat{\theta}_{i,j,k} \sim \mathcal{N} (\bar{\theta}_{i,j,k}, 2\sigma_b^2)$.
And the angle $\theta_{i,j,k}$ between bearings of the positions from MDS is calculated as:
\begin{equation}
    {{\theta}_{i,j,k}} = \cos^{-1}\left({^{F_u}\tilde{\mathbf{p}}_{R_i \to R_j}} \cdot {^{F_u}\tilde{\mathbf{p}}_{R_i \to R_k}}\right).
    \label{Equ: bearing angle}
\end{equation}
For simplicity, we assume ${^{F_u}\mathbf{p}_{R_i}} \sim \mathcal{N} ({^{F_u}\mathbf{p}_{R_i}}, \bm{\Sigma}_{d}=\sigma_d^2 \mathbf{I})$, then we have $\theta_{i,j,k} \sim \mathcal{N} (\bar{\theta}_{i,j,k}, {\sigma}_{{\theta}_{i,j,k}}^2)$, where:
\begin{equation}
    {\sigma}_{{\theta}_{i,j,k}} = \sqrt{\left(\frac{\sigma_d}{\|^{F_u}\mathbf{p}_{R_i \to R_j}\|}\right)^2+\left(\frac{\sigma_d}{\|^{F_u}\mathbf{p}_{R_i \to R_k}\|}\right)^2}.
\end{equation}

Since $({\hat{\theta}_{i,j,k}} - {\theta_{i,j,k}}) \sim \mathcal{N}(0, 2\sigma_b^2 + \sigma_{{\theta}_{i,j,k}}^2)$, we can set a threshold based on probability.
If the difference is smaller than the threshold, we consider the bearings ${{\hat{\mathbf{z}}_b}{}_{B_i \to M_j}}$ and ${{\hat{\mathbf{z}}_b}{}_{B_i \to M_k}}$ to be consistent with each other.

Given the $n-1$ bearings ${{\hat{\mathbf{z}}_b}{}_{B_i \to M_j}}$, we construct an undirected consistency graph $G=(V, E)$, where each vertex $v \in V$ represents a bearing measurement, and each edge $e \in E$ indicates consistency between the two connected vertices.
If robot $j$ is not visible to robot $i$, the corresponding vertex $v_j$ is omitted from the graph, which does not affect other vertices.
When the bearings ${{\hat{\mathbf{z}}_b}{}_{B_i \to M_j}}$ and ${{\hat{\mathbf{z}}_b}{}_{B_i \to M_k}}$ are consistent, an edge $e_{jk}$ is added between $j$ and $k$.
Due to the high error rate observed in heuristic methods \cite{pattabiraman2015fast} for finding the maximum clique, we employ Clipper+ \cite{fathian2024clipper+} to efficiently find the maximal clique in the graph $G$, which contains the most consistent bearings.
The bearings within the maximal clique are assigned ${w_b}_{i,j} = 1$, while others are set to ${w_b}_{i,j} = 0$.
PCM-B is a lightweight preprocessing step, with runtime at ms-level even under 50\% outliers and 100 device, so it does not become a performance bottleneck.

\section{Experiments}
\label{Sec: Experiments}

\subsection{Evaluation Metrics}
\label{Sec: Evaluation Metrics}

We adopt the Absolute Trajectory Error (ATE) \cite{zhang2018tutorial}, which measures the global consistency of an estimated trajectory with respect to the groundtruth.
We compute the ATE in the local frame of device 0 ($R_r = R_0$), with the groundtruth rotation ${^{W}\mathbf{R}_{R_j}}$ and position ${^{W}\mathbf{p}_{R_j}}$ of device $j$.
For a single frame at time $t_i$, the rotation and position ATE are defined as:
\begin{equation}
    \begin{aligned}
        \text{ATE}_{R}^{t_i} & = \sqrt{ \frac{1}{n-1} \sum_{j=1}^{n-1} \left\| \mathbf{Log} \left({^{W}\mathbf{R}_{R_0}^{t_i}}^T {^{W}\mathbf{R}_{R_j}^{t_i}} {^{R_0}\hat{\mathbf{R}}_{R_j}^{t_i}}^T\right) \right\|^2 },
        \\
        \text{ATE}_{p}^{t_i} & = \sqrt{ \frac{1}{n-1} \sum_{j=1}^{n-1} \left\| {^{W}\mathbf{R}_{R_0}^{t_i}}^T ({^{W}\mathbf{p}_{R_j}^{t_i}} - {^{W}\mathbf{p}_{R_0}^{t_i}}) - {^{R_0}\hat{\mathbf{p}}_{R_j}^{t_i}} \right\|^2 }.
    \end{aligned}
    \nonumber
\end{equation}
Over a sequence of $M_f$ frames, they are aggregated as:
\begin{equation}
    \text{ATE}_{R} = \sqrt{ \frac{1}{M_f} \sum_{i=1}^{M_f} \left(\text{ATE}_{R}^{t_i}\right)^2 },
    \text{ATE}_{p} = \sqrt{ \frac{1}{M_f} \sum_{i=1}^{M_f} \left(\text{ATE}_{p}^{t_i}\right)^2 }.
    \nonumber
\end{equation}

Unless otherwise specified, the experiments run on a computer with an Intel i5-1260P CPU and 16\,GB of RAM.
Bias estimation is supported but disabled in our experiments, as it yields limited gains at additional computational cost.
In the real-world experiments, each IMU is kept static for about the first $2\,\mathrm{s}$ after power-on, during which the IMU driver estimates and compensates for the initial accelerometer and gyroscope biases.
For long-horizon operation or noisier IMUs, however, we recommend enabling bias estimation, since bias drift can then become a major error source.

We evaluate \CREPES{} based on the \CHALLENGES{}.
For NGI, the underlying theory guarantees independence from any environmental or global information.
Moreover, Xun's work \cite{xun2023crepes} has shown that this hardware performs reliably in dark and outdoor long-distance scenarios (see Sec. IV-B).
Motivated by these properties and prior results, we design the experiments as follows:
\begin{itemize}
    \item \autoref{Sec: Evaluation in Benchmark}: Simulated benchmarks to thoroughly test the accuracy, scalability, and resilience of \CREPES{}.
    \item \autoref{Sec: Evaluation in Outlier Existence}: Simulated experiments evaluating the robustness of \CREPES{} in the presence of outliers, for OE.
    \item \autoref{Sec: Evaluation in Real-World Datasets}: Accuracy comparison experiments in real-world multi-robot scenarios, for NLOS.
    \item \autoref{Sec: Evaluation in Non-inertial Environments}: Variable gravity simulation and space dataset to test performance in gravity-disabled scenarios, for NI.
\end{itemize}

\subsection{Evaluation in Benchmark}
\label{Sec: Evaluation in Benchmark}

In the benchmark, we generate a random SE(3) B-spline to simulate each device's trajectory.
The position control points are uniformly distributed within a $10 \times 10 \times 10 \, \text{m}^3$ space, while the rotation control points are generated based on \cite{yershova2010generating}.
The bearing, distance, and gravity measurements are generated by sampling positions and rotations along the trajectory and performing numerical calculations as described in \autoref{Sec: Problem Statement and Measurement Model}.
IMU measurements are derived from the B-spline's derivative.
Gaussian noise is added to the measurements, with standard deviations of $0.1\,\text{m}$ for distance, $2.0\degree$ for bearing, $2.0\degree$ for gravity, $0.1\,\text{m/s}^2$ for acceleration and $0.01\,\text{rad/s}$ for angular velocity, with bias drifts of $0.001\,\text{m/s}^3$ and $0.0001\,\text{rad/s}^2$, respectively.
Distance, gravity, and inertial measurements are generated at $100\,\text{Hz}$, while bearing measurements are generated at $50\,\text{Hz}$.
$w_g$ is set to $1$ to enable gravity fusion.

\begin{figure}[t]
    \centering
    \includegraphics[width=\halfwidth]{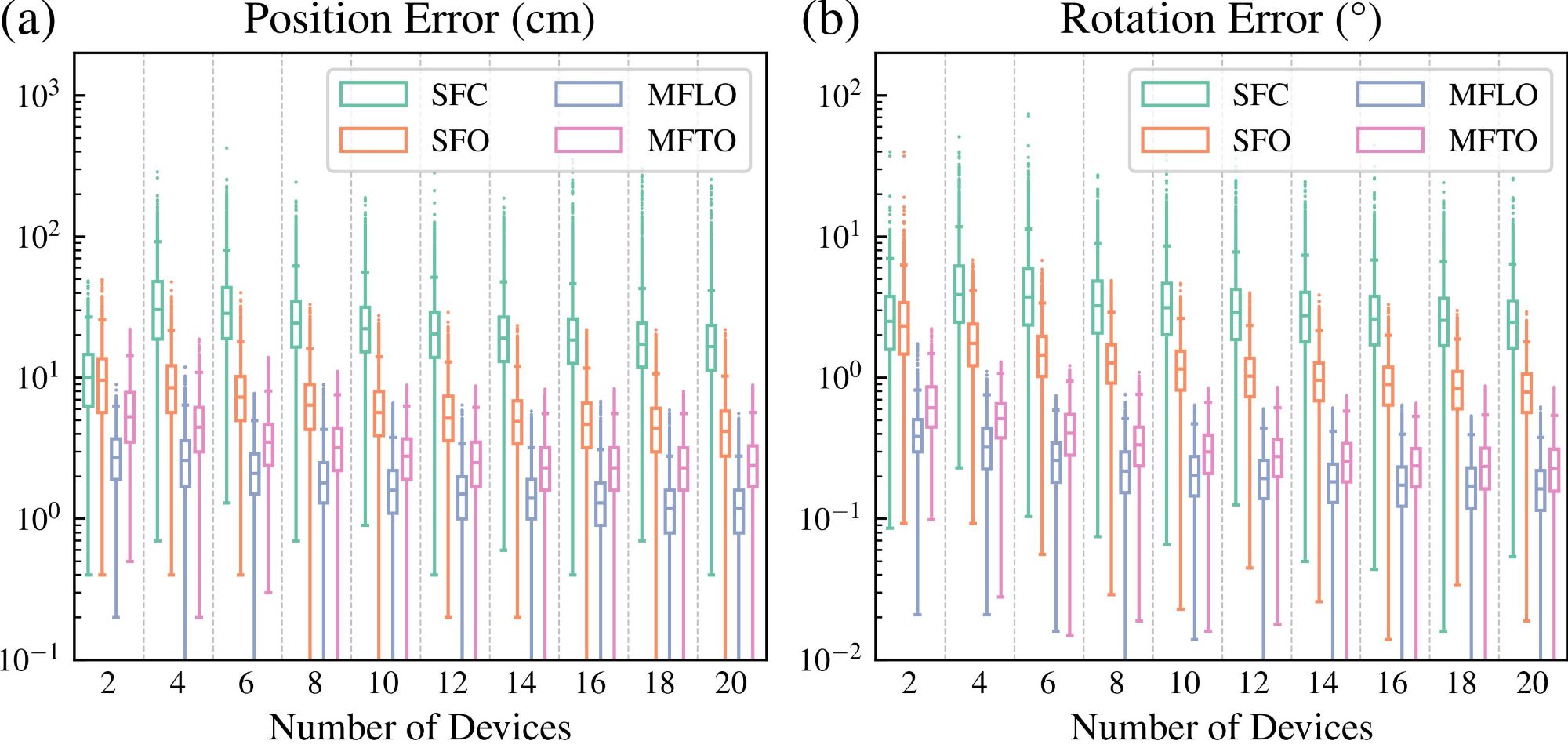}
    \caption{
        Accuracy of the hierarchical outputs of \CREPES{} related to
        (a) position ATE (log scale) and
        (b) rotation ATE (log scale).
    }
    \label{Fig: Exp: Benchmark Results}
\end{figure}
\begin{figure}[t]
    \centering
    \includegraphics[width=\halfwidth]{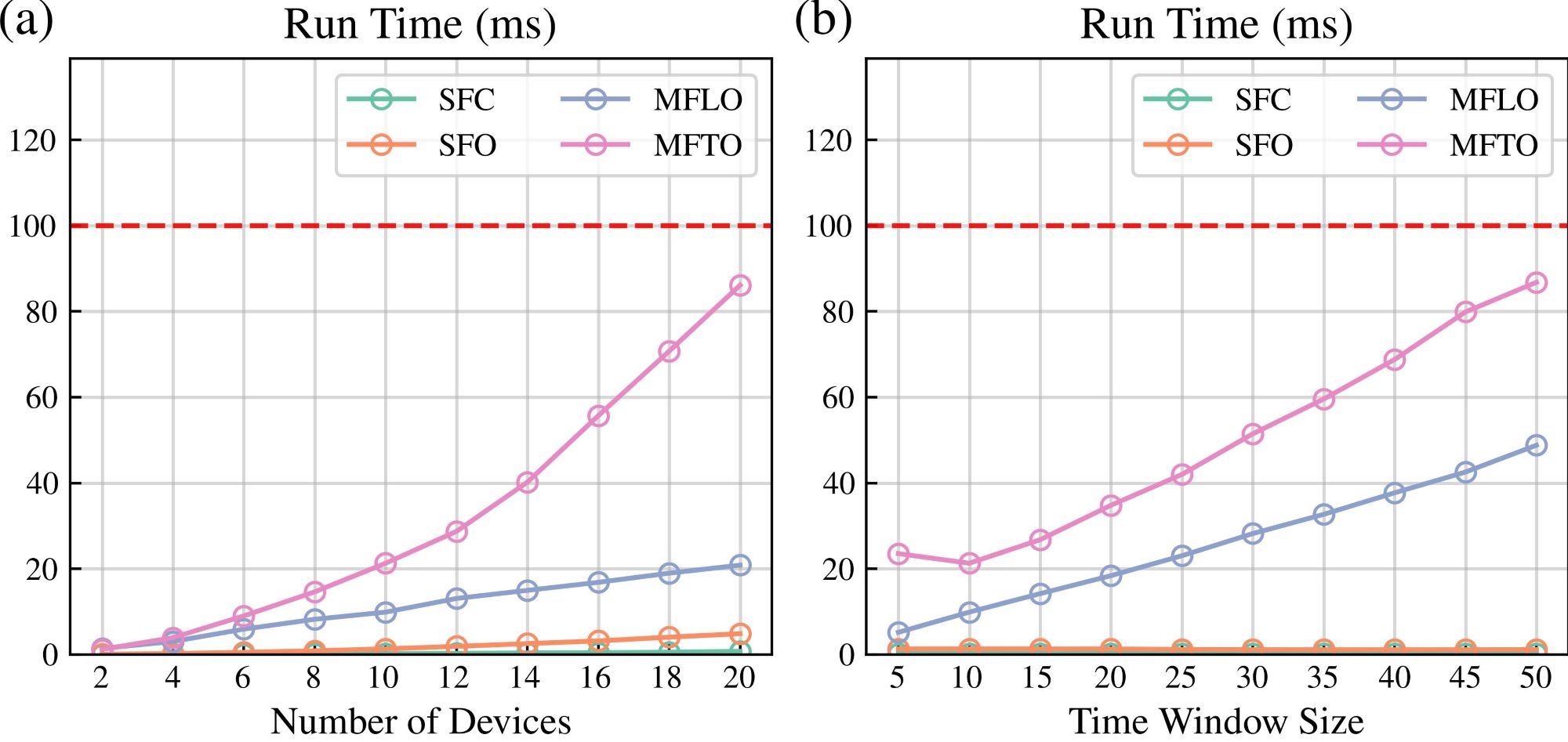}
    \caption{
        Run time of \CREPES{} related to
        (a) the number of devices (time window size of $10$) and
        (b) the size of the time window ($10$ devices).
    }
    \label{Fig: Exp: Scalability Results}
\end{figure}

\subsubsection{Accuracy}
\label{Sec: Accuracy}

We evaluate the accuracy of \CREPES{} under $2$ to $20$ devices.
\autoref{Fig: Exp: Benchmark Results} reports the ATE for all hierarchical outputs.
For instance, with $10$ devices and a time window of $10$ (i.e., number of keyframes), \CREPES{} achieves a position ATE of $3.2\,\text{cm}$ and a rotation ATE of $0.3\degree$.

As expected, estimation errors generally decrease as the number of devices increases, due to the fact that more observations are fused.
Specifically, with $20$ devices, MFTO achieves a $4.1\%$ reduction in position ATE and a $20.0\%$ reduction in rotation ATE, compared to the $10$-device case.

SFC and SFO yield similar errors with $2$ devices, as the closed-form solution is already optimal.
As the number of devices increases, SFC's error grows slightly due to unconsidered coupling between measurements, while SFO's error decreases.
In the $10$ devices setup, SFO reduces position and rotation errors by $76.0\%$ and $69.0\%$ compared to SFC.

Fusing IMU measurements across multiple frames further reduces estimation errors.
Within the hierarchical estimators, MFTO can be regarded as a multi-frame extension of SFO augmented with IMU residuals.
Compared to SFO, MFTO achieves a $54.9\%$ reduction in positional ATE and a $74.6\%$ reduction in rotational ATE when evaluated on $10$ devices.

MFLO outperforms MFTO by $65.1\%$ in position and $42.4\%$ in rotation, as SFRE preserves the bearing and distance information of every frame in its pose estimates.
In contrast, MFTO uses only keyframes, limiting the amount of information it can utilize.
When bearings are partially missing, however, the performance of SFRE degrades, thereby impacting MFLO, while MFTO shows greater resilience, as discussed later.

\subsubsection{Scalability}
\label{Sec: Scalability}

The computational cost of \CREPES{} depends on both the number of devices and the time window size.
\autoref{Fig: Exp: Scalability Results} reports the total time for problem construction and optimization, with MFTO being the most time-consuming component.
With 6 devices and a time window of 10, \CREPES{} runs at $100\,\text{Hz}$ (IMU rate); with 9 devices, it operates at $50\,\text{Hz}$ (camera rate); and with 20 devices, it maintains real-time performance at $10\,\text{Hz}$.
These results show \CREPES{}'s potential for real-world multi-robot applications.

\begin{figure}[t]
    \centering
    \includegraphics[width=\halfwidth]{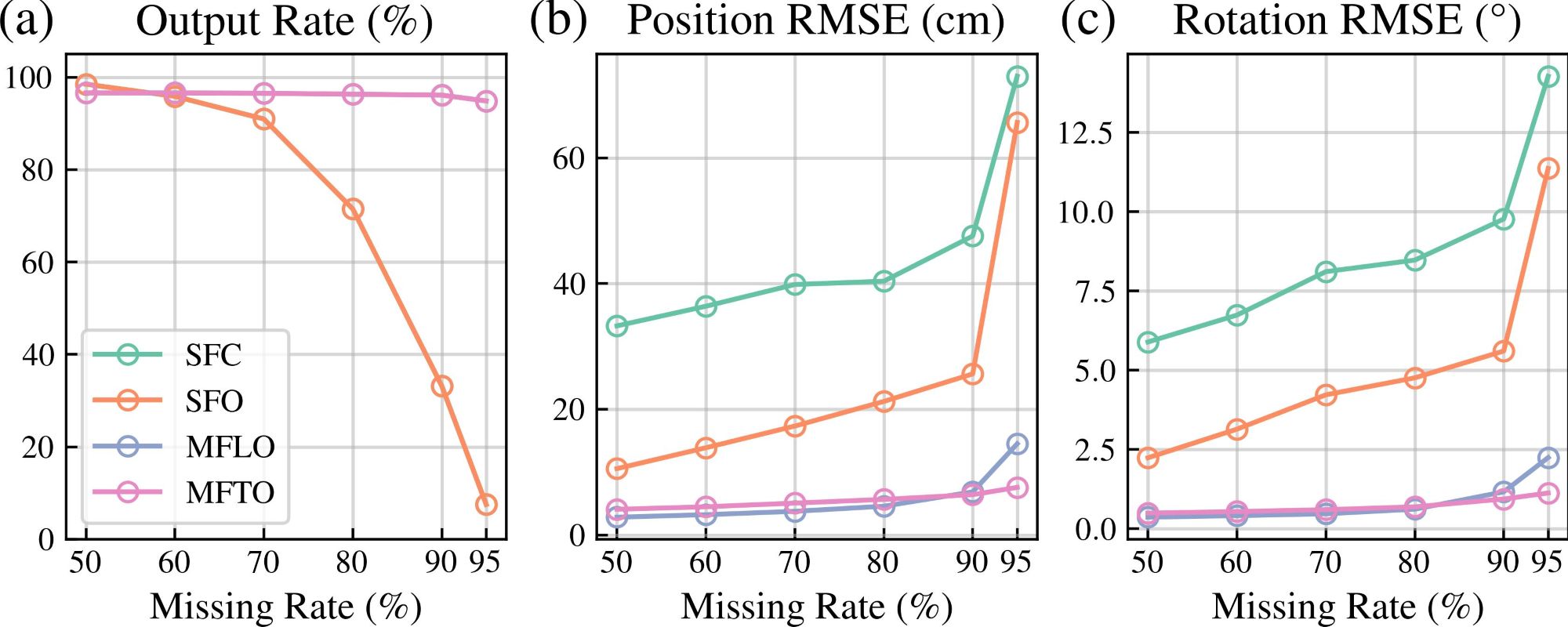}
    \caption{
        Resilience illustration with different missing rates.
        (a) Output rate (SFC and SFO are the same, MFLO and MFTO are the same).
        (b) Position error.
        (c) Rotation error.
    }
    \label{Fig: Benchmark: Missing Results}
\end{figure}

\subsubsection{Resilience}
\label{Sec: Resilience}

We evaluate the resilience under varying levels of perceptual degradation.
Resilience is assessed by the bearing missing rate, defined as $1-(|\mathcal{B}_i| / (n-1))$.
For example, with $10$ devices and $3$ bearings per device on average, the missing rate is $33\%$.
Resilience is quantified using position error, rotation error, and output rate, which is defined as the ratio of valid outputs to the total number of frames.

\autoref{Fig: Benchmark: Missing Results} shows the results of the estimators under bearing missing rates from $50\%$ to $95\%$.
At $50\%$, all methods produce outputs with MFLO achieving the lowest errors ($3.0\,\text{cm}$, $0.4\degree$), followed by MFTO, SFO, and SFC.
As the missing rate increases, SFC and SFO show steady error growth, while MFLO and MFTO remain stable.
At $90\%$, the output rate of SFC and SFO drops below $40\%$, while MFLO and MFTO retain $90\%$ output rate.
At $95\%$, SFC and SFO only produce a few results with large errors ($66.0\,\text{cm}$, $11.4\degree$), MFLO shows moderate degradation, while MFTO maintains full output with minimal errors ($8.0\,\text{cm}$, $1.1\degree$).
These results demonstrate that MFTO achieves the highest resilience, validating the tightly-coupled design under extreme perceptual degradation.

\subsection{Evaluation in Outlier Existence}
\label{Sec: Evaluation in Outlier Existence}

\subsubsection{Robustness}
\label{Sec: Outlier: Robustness}

\begin{figure}[t]
    \centering
    \includegraphics[width=\halfwidth]{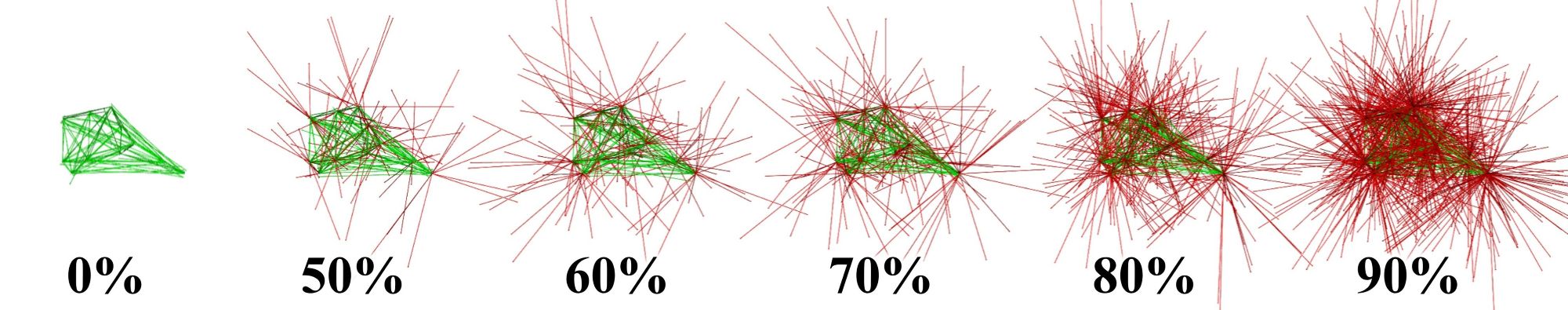}
    \caption{
        Input bearing measurements fed into \CREPES{} at different outlier rates.
        The red lines are outliers, and the green lines are inliers.
    }
    \label{Fig: Exp: Outlier Visualization}
\end{figure}

We evaluate the robustness of the outlier rejection algorithm in \CREPES{}.
The outlier rate is defined as $(|\mathcal{B}_i| - |\mathcal{B}_i'|) / |\mathcal{B}_i|$, where $|\mathcal{B}_i'|$ denotes the cardinality of the expected bearing set.
Outliers are simulated by adding randomly generated bearings with random IDs to the expected bearing set, see \autoref{Fig: Exp: Outlier Visualization}.
For example, with 10 devices, each providing 9 bearings, a 50\% outlier rate corresponds to the addition of 9 outlier bearings to the original set.
We assess the performance of outlier rejection using precision and recall across varying outlier rates.
Accuracy is not an appropriate metric in this context, for instance, if 90\% of the bearings are outliers, labeling all as outliers would yield 90\% accuracy.

\begin{figure}[t]
    \centering
    \includegraphics[width=\halfwidth]{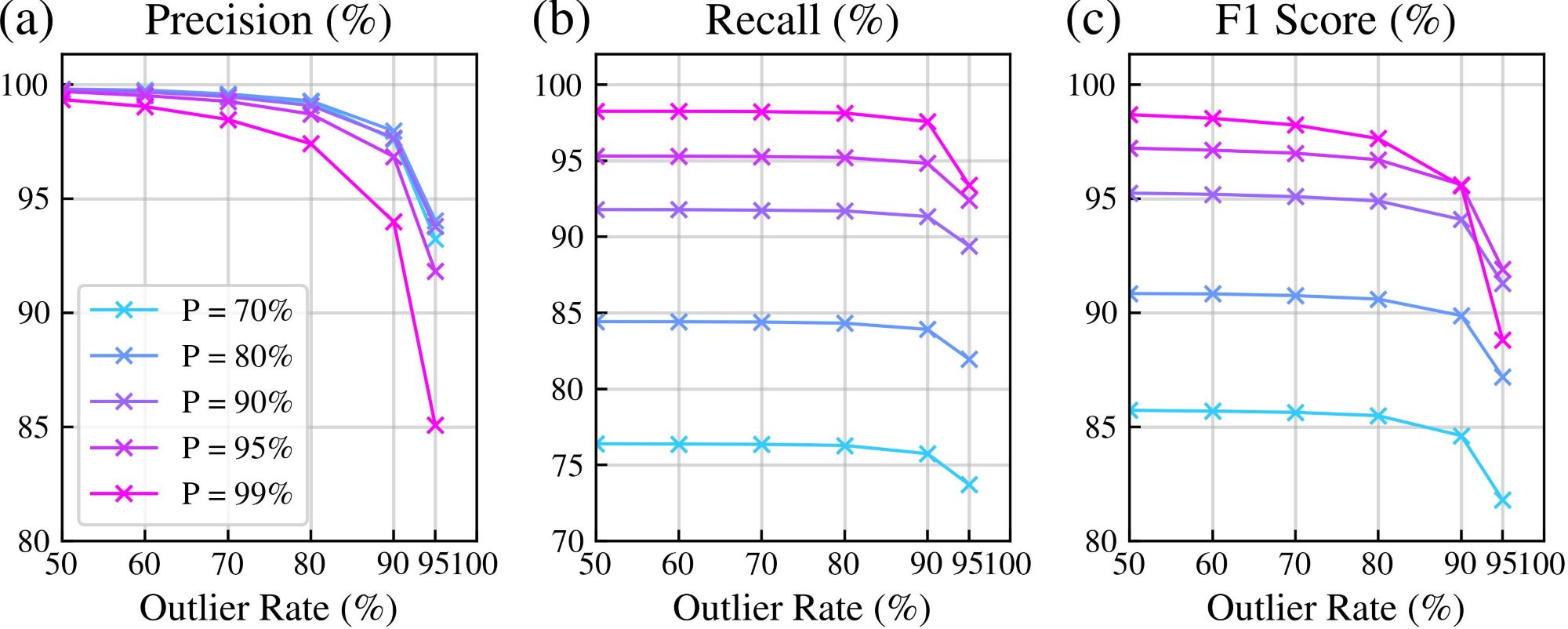}
    \caption{
        Results of precision and recall rate in different outlier rates, 95\% is a suitable probability threshold.
        (a) Precision.
        (b) Recall rate.
        (c) F1 score.
    }
    \label{Fig: Exp: Outlier Rates}
\end{figure}

The evaluation results are shown in \autoref{Fig: Exp: Outlier Rates}.
We assess precision and recall under different angular thresholds for \eqref{Equ: bearing angle}.
PCM-B remains effective with up to 90\% outliers: at 70\% threshold, precision reaches 97.6\% but recall drops to 75.7\%; at 99\%, recall is 97.6\% but precision falls to 94.0\%.
A 95\% threshold provides a good balance, achieving 96.8\% precision and 94.8\% recall at 90\% outliers.
The processing time of PCM-B remains below $1\,\text{ms}$ when the outlier rate is under 90\%.

\begin{figure}[t]
    \centering
    \includegraphics[width=\halfwidth]{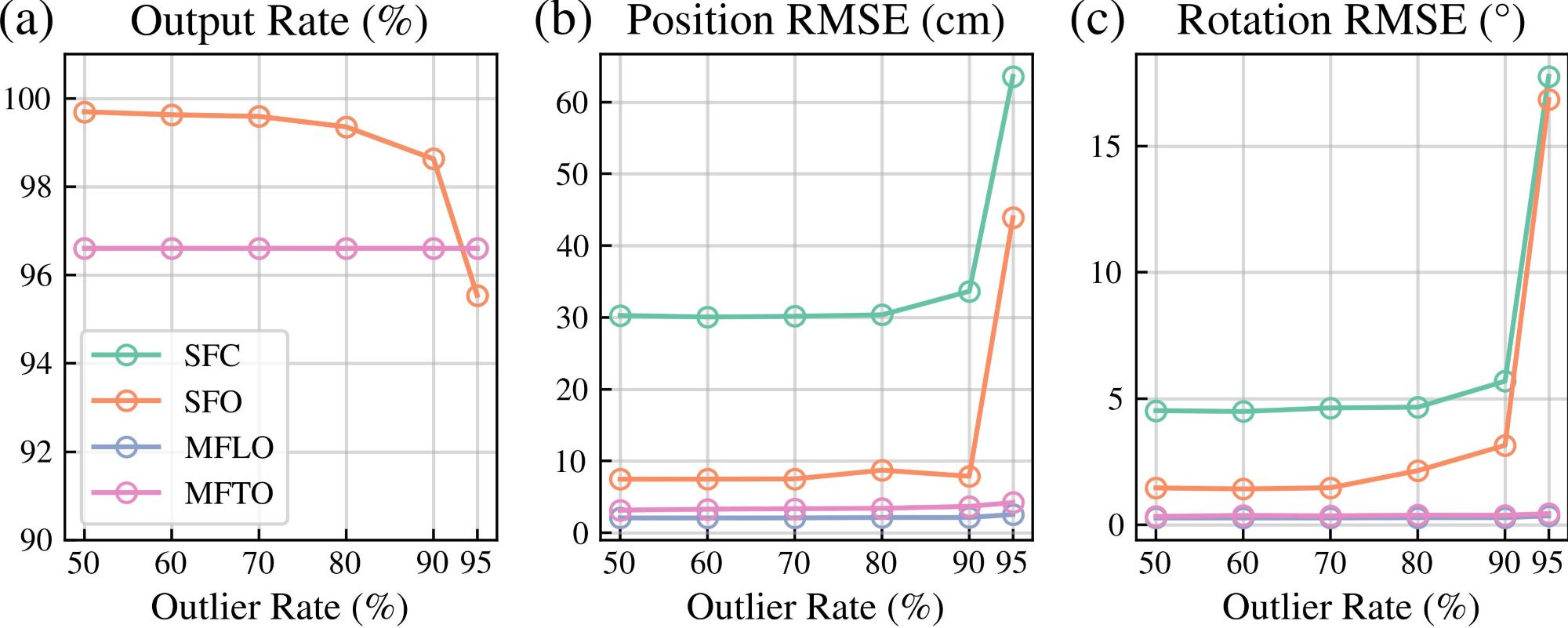}
    \caption{
        Robustness results.
        (a) Output rate (SFC and SFO are the same, MFLO and MFTO are the same).
        (b) Position error.
        (c) Rotation error.
    }
    \label{Fig: Exp: Outlier Results}
\end{figure}

We further evaluate the accuracy of \CREPES{} under varying outlier rates, fixing the PCM-B threshold to 95\%.
As shown in \autoref{Fig: Exp: Outlier Results}, when the outlier rate is below 90\%, the ATEs of MFLO and MFTO remain below $4.0\,\text{cm}$ and $0.4\degree$.
Even at a 95\% outlier rate, MFLO and MFTO maintain stable accuracy with ATE under $5.0\,\text{cm}$ and $0.5\degree$.
By contrast, SFC and SFO show rapid degradation beyond 90\% outliers.
These results validate the effectiveness of PCM-B and the robustness of MFLO and MFTO in extreme outlier conditions.

\begin{table}[t]
    \centering
    \caption{Outlier Rejection Results under $90\%$ Outliers}
    \label{Tab: Exp: Outlier Rejection Results under Outliers}
    \resizebox{.8\halfwidth}{!}{
        \begin{tabular}{@{\hskip 1pt}c@{\hskip 5pt}c@{\hskip 5pt}c@{\hskip 5pt}c@{\hskip 5pt}c@{\hskip 5pt}c@{\hskip 5pt}c@{\hskip 5pt}c@{\hskip 3pt}c@{\hskip 1pt}}
            \toprule
            \multirow{3}{*}{\#}                              &
            \multirow{3}{*}{\rotatebox[origin=c]{90}{PCM-B}} &
            \multirow{3}{*}{\rotatebox[origin=c]{90}{GNC-S}} &
            \multirow{3}{*}{\rotatebox[origin=c]{90}{GNC-M}} &
            SFC                                              &
            SFO                                              &
            MFLO                                             &
            \multicolumn{2}{c}{MFTO}
            \\
            \cmidrule(r){5-5}
            \cmidrule(r){6-6}
            \cmidrule(r){7-7}
            \cmidrule{8-9}
                                                             &
                                                             &
                                                             &
                                                             &
            Pos / Rot
                                                             &
            Pos / Rot
                                                             &
            Pos / Rot
                                                             &
            Pos / Rot
                                                             &
            Runtime
            \\
                                                             &
                                                             &
                                                             &
                                                             &
            ($\mathrm{cm}$) / ($\degree$)
                                                             &
            ($\mathrm{cm}$) / ($\degree$)
                                                             &
            ($\mathrm{cm}$) / ($\degree$)
                                                             &
            ($\mathrm{cm}$) / ($\degree$)
                                                             &
            ($\mathrm{ms}$)
            \\
            \midrule
            1                                                &           &           &           & 410.2 / 70.3        & 328.1 / 65.7       & 219.1 / 36.4       & 14.5 / 1.2         & 80.3          \\
            2                                                &           & \ding{51} &           & 410.2 / 70.3        & 315.7 / 64.3       & 183.4 / 34.8       & 13.4 / 1.2         & \textbf{71.3} \\
            3                                                &           &           & \ding{51} & 410.2 / 70.3        & 328.1 / 65.7       & 219.1 / 36.4       & 3.6 / 0.4          & 232.5         \\
            4                                                &           & \ding{51} & \ding{51} & 410.2 / 70.3        & 315.7 / 64.3       & 183.4 / 34.8       & 3.6 / 0.4          & 217.3         \\
            \rowcolor{gray!15}[3pt][3pt]
            5                                                & \ding{51} &           &           & \textbf{33.6 / 5.7} & \textbf{7.9 / 3.1} & \textbf{2.1 / 0.3} & 3.7 / 0.4          & 86.7          \\
            6                                                & \ding{51} & \ding{51} &           & \textbf{33.6 / 5.7} & \textbf{7.9 / 3.1} & \textbf{2.1 / 0.3} & 3.6 / 0.4          & 87.4          \\
            7                                                & \ding{51} &           & \ding{51} & \textbf{33.6 / 5.7} & \textbf{7.9 / 3.1} & \textbf{2.1 / 0.3} & \textbf{3.4 / 0.4} & 220.7         \\
            8                                                & \ding{51} & \ding{51} & \ding{51} & \textbf{33.6 / 5.7} & \textbf{7.9 / 3.1} & \textbf{2.1 / 0.3} & 3.5 / 0.4          & 222.2         \\
            \bottomrule
        \end{tabular}
    }
\end{table}

\subsubsection{Combination and Comparison with GNC}
\label{Sec: Outlier: Combination and Comparison with GNC}

We also evaluate the integration of PCM-B with GNC \cite{yang2020graduated}, which rejects outliers by iteratively adjusting the weights $w_b$ and $w_d$ during optimization.
Based on the single-frame and multi-frame formulations in \eqref{Equ: SFO: Pose Optimization} and \eqref{Equ: MFTO: Pose Optimization}, we denote the methods as GNC-S and GNC-M, respectively.

As shown in \autoref{Tab: Exp: Outlier Rejection Results under Outliers}, PCM-B by itself (row 2) achieves consistently robust results across all modules.
In contrast, the GNC variants are less effective: GNC-S alone (row 3) proves insufficient, leading to significantly higher errors, while GNC-M alone (row 4) only improves the MFTO module to a level competitive with PCM-B.
Although combining PCM-B with GNC (rows 6-8) yields slight improvements, this comes at a steep computational cost: the iterative nature of GNC increases the MFTO runtime by $2.1\times$.
Therefore, PCM-B alone offers the optimal trade-off between accuracy and efficiency.

\subsection{Evaluation in Real-World Datasets}
\label{Sec: Evaluation in Real-World Datasets}

We evaluate \CREPES{} with 5 devices in real-world, including both hand-held and mobile robots experiments.
The real-world datasets are collected on Intel N100, Intel i5-1260P CPU , and NVIDIA Jetson Xavier NX CPU.
The groundtruth poses are provided by NOKOV MCS.
The groundtruth of device $i$ is expressed in a frame $\tilde{R}_i$ that differs from the device frame $R_i$ by an extrinsic transformation.
While the positional extrinsic ${^{{\tilde{R}}_i}\mathbf{p}_{R_i}}$ can be manually measured with high accuracy, the rotational extrinsic ${^{{\tilde{R}}_i}\mathbf{R}_{R_i}}$ is hard to determine.
To estimate ${^{{\tilde{R}}_i}\mathbf{R}_{R_i}}$ for all devices, we minimize the overall bearing error:
\begin{equation}
    \begin{aligned}
         & {^{W}\mathbf{p}^{t_i}_{R_k}} = {^{W}\mathbf{p}_{{\tilde{R}}_k}^{t_i}}+{^{W}\mathbf{R}_{{\tilde{R}}_k}^{t_i}}{^{{\tilde{R}}_k}\mathbf{p}_{R_k}},~
        {^{W}\mathbf{R}^{t_i}_{R_j}} = {^{W}\mathbf{R}_{\tilde{R}_j}^{t_i}}{{^{\tilde{R}_j}}\mathbf{R}_{R_j}}
        \\
         & \min_{^{{\tilde{R}}_j}\mathbf{R}_{R_j}} \sum_{i,j}^{m,n} \sum_{k \in \mathcal{B}^{t_i}_j}
        \left\|{^{W}\mathbf{R}^{t_i}_{R_j}}^T {^{W}\tilde{\mathbf{p}}^{t_i}_{R_j \to R_k}} - {^{R_j}\mathbf{R}_{B_j}^T} {\hat{\mathbf{z}}_b}{}^{t_i}_{B_j \to M_k}\right\|^2.
    \end{aligned}
    \nonumber
\end{equation}

\begin{table}[t]
    \centering
    \caption{Experiment Configuration ($\sigma$ denotes RMSE)}
    \label{Tab: Experiment Configuration}
    \resizebox{\halfwidth}{!}{
        \begin{tabular}{@{\hskip 1pt}l@{\hskip 3pt}c@{\hskip 2pt}c@{\hskip 2pt}c@{\hskip 3pt}c@{\hskip 2pt}c@{\hskip 2pt}c@{\hskip 3pt}c@{\hskip 2pt}c@{\hskip 2pt}c@{\hskip 2pt}c@{\hskip 2pt}c@{\hskip 1pt}}
            \toprule
            \multirow{3}{*}{Dataset}
             &
            \multicolumn{3}{c}{Camera}
             &
            \multicolumn{3}{c}{UWB}
             &
            \multicolumn{5}{c}{IMU}
            \\
            \cmidrule(r){2-4}
            \cmidrule(r){5-7}
            \cmidrule{8-12}
             & Freq.           & BW.               & Bear. $\sigma$ & Freq.           & BW.               & Dist. $\sigma$  & Freq.           & BW.               & Acce. $\sigma$     & Gyro. $\sigma$        & Grav. $\sigma$ \\
             & ($\mathrm{Hz}$) & ($\mathrm{KB/s}$) & ($^\circ$)     & ($\mathrm{Hz}$) & ($\mathrm{KB/s}$) & ($\mathrm{cm}$) & ($\mathrm{Hz}$) & ($\mathrm{KB/s}$) & ($\mathrm{m/s^2}$) & ($^\circ/\mathrm{s}$) & ($^\circ$)     \\
            \midrule
            \texttt{\textbf{LOS}}
             & 50              & 2.32              & 3.7            & 100             & 8.24              & 5.5             & 100             & 11.44             & 1.9                & 12.2                  & 1.5            \\
            \texttt{\textbf{NLOS}}
             & 50              & 2.32              & 3.3            & 100             & 8.24              & 5.7             & 100             & 11.44             & 2.1                & 12.5                  & 1.7            \\
            \texttt{\textbf{HDM}}
             & 50              & 2.32              & 2.9            & 100             & 8.24              & 10.7            & 100             & 11.44             & 0.9                & 10.4                  & 1.9            \\
            \texttt{\textbf{MRP}}
             & 50              & 2.32              & 24.5           & 100             & 8.24              & 16.2            & 200             & 22.88             & 7.7                & 14.4                  & 3.6            \\
            \bottomrule
        \end{tabular}
    }
\end{table}

\begin{figure}[t]
    \centering
    \includegraphics[width=\halfwidth]{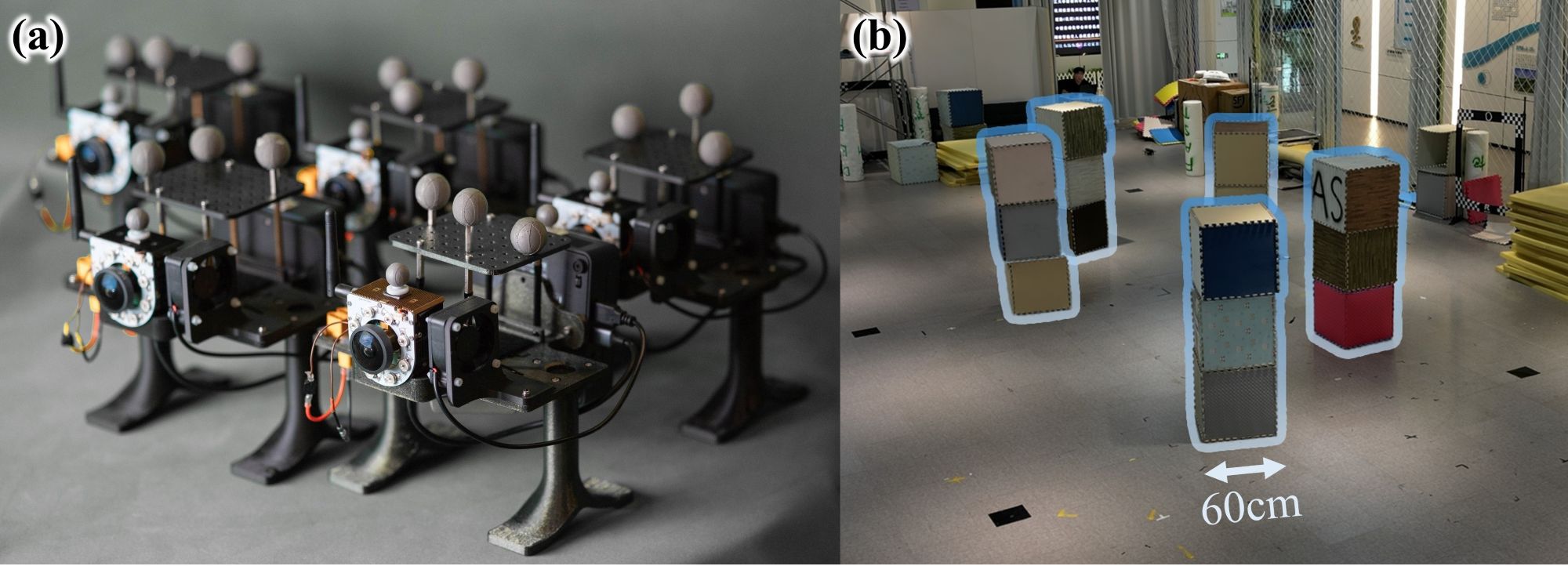}
    \caption{
        Handheld experimental platform of \CREPES{}.
        (a) The platform of five \CREPES{} devices.
        (b) The environment of NLOS experiments.
    }
    \label{Fig: Exp: Real-World Environment}
\end{figure}

\begin{figure}[t]
    \centering
    \includegraphics[width=\halfwidth]{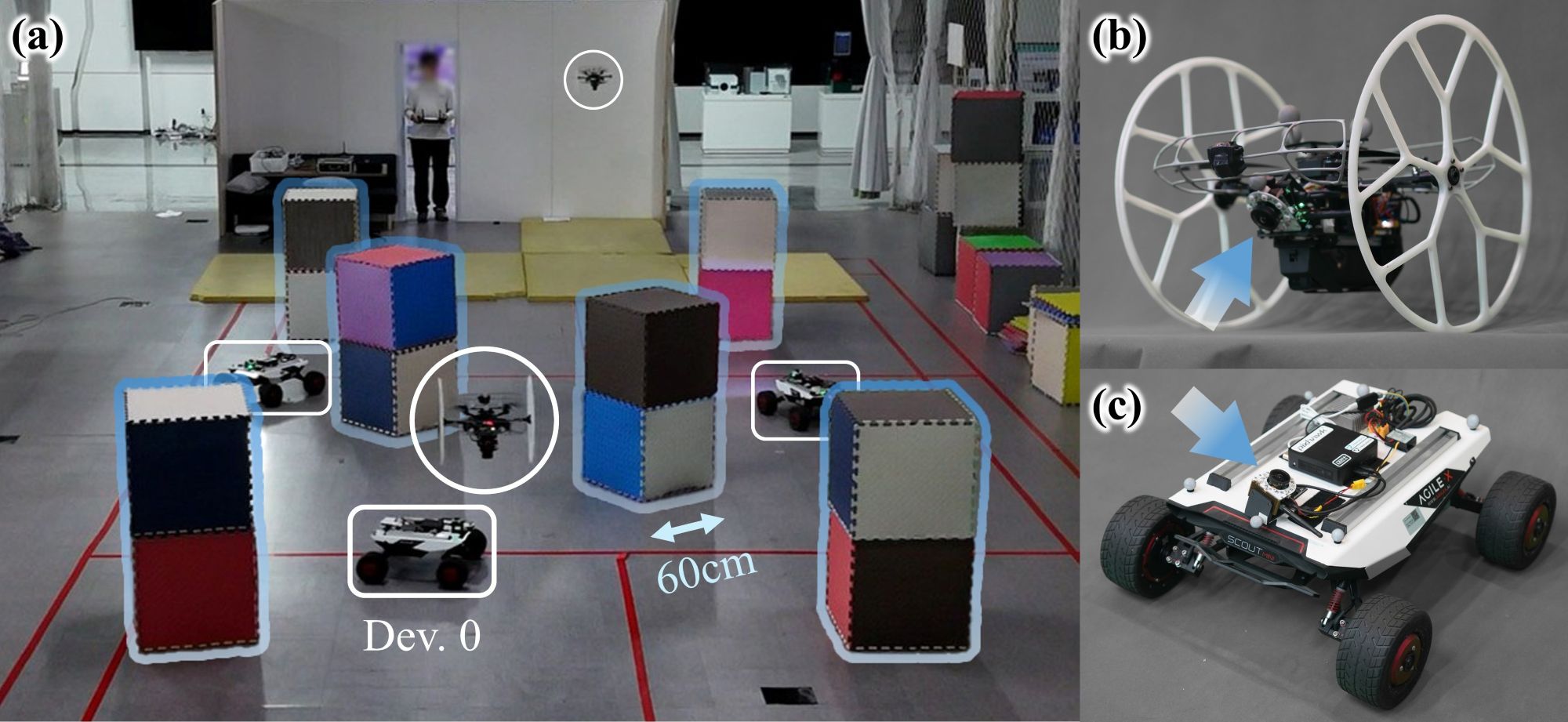}
    \caption{
        Multi-robot experimental platform of \CREPES{}.
        (a) Experimental environment for \texttt{MRP\_3} and \texttt{MRP\_4}, where circles and squares represent UAV and UGV positions, respectively.
        (b) The UAV platform, with the on-board \CREPES{} device highlighted by a blue arrow.
        (c) The UGV platform, with the on-board \CREPES{} device highlighted by a blue arrow.
    }
    \label{Fig: Exp: Real-World Environment 2}
\end{figure}

\autoref{Tab: Experiment Configuration} summarizes the sensor frequencies, bandwidths, and Root Mean Square Errors (RMSEs).
The IMU RMSE refers to the gravity-direction error, computed from roll-pitch estimates provided by the complementary filter \cite{mahony2008nonlinear} and the groundtruth.
The specific hardware implementation (Time Synchronization, Swarm Bridge, and ID Extraction) in \autoref{Sec: Hardware Design and Implementation} incurs minimal computational cost and can be neglected.
The IMUs are calibrated for scale factors and axis misalignments via imu-tk~\cite{tedaldi2014robust}; the time window is fixed to $30$ keyframes and $w_g$ is set to $1$.
Four scenario families are collected within a $10 \times 20 \times 3\,\text{m}^3$ indoor space, including:
\begin{itemize}
    \item \textbf{\texttt{LOS}}: five hand-held devices (\autoref{Fig: Exp: Real-World Environment}(a)) traverse an open space, with average linear and angular speeds increasing from $(33.2\,\mathrm{cm/s},16.2^\circ/\mathrm{s})$ in \texttt{LOS\_1} to $(96.1\,\mathrm{cm/s},25.4^\circ/\mathrm{s})$ in \texttt{LOS\_4};
    \item \textbf{\texttt{NLOS}}: the same setting as \textbf{\texttt{LOS}}, but with arbitrarily placed obstacles (\autoref{Fig: Exp: Real-World Environment}(b)), with average linear and angular speeds ranging from $(38.9\,\mathrm{cm/s},14.9^\circ/\mathrm{s})$ in \texttt{NLOS\_1} to $(91.0\,\mathrm{cm/s},24.2^\circ/\mathrm{s})$ in \texttt{NLOS\_4};
    \item \textbf{\texttt{HDM}}: two devices among the five are hand-swung at high speed while the remaining remain static, with circular and figure-eight motions under LOS/NLOS (\autoref{Fig: Exp: Real-World HDM Results}(a)-(d)). Across \texttt{HDM\_1}--\texttt{HDM\_4}, the average linear and angular speeds range from $(141.8\,\mathrm{cm/s},94.8^\circ/\mathrm{s})$ to $(216.4\,\mathrm{cm/s},186.8^\circ/\mathrm{s})$;
    \item \textbf{\texttt{MRP}}: a heterogeneous deployment with two UAVs and three UGVs under LOS/NLOS (\autoref{Fig: Exp: Real-World Environment 2}), with average linear and angular speeds ranging from $(89.3\,\mathrm{cm/s},21.4^\circ/\mathrm{s})$ to $(107.4\,\mathrm{cm/s},24.8^\circ/\mathrm{s})$.
\end{itemize}

The dataset details and results are summarized in \autoref{Tab: RMSE of CREPES in Real-World Datasets}.
Position and rotation errors are aggregated over the four non-reference devices (i.e., excluding Device 0); for the \textbf{\texttt{HDM}} rows only Device 1 (w.r.t.\ Device 0) is reported because the other three devices are static.
Xun's \cite{xun2023crepes} with normalized-innovation squared (NIS) outlier rejection is included as a baseline.

\begin{table*}[t]
    \centering
    \caption{
        RMSE of \CREPES{} in Real-World Datasets.
        (The values in parentheses denote the median errors.)
    }
    \label{Tab: RMSE of CREPES in Real-World Datasets}
    \resizebox{\fullwidth}{!}{
        \begin{tabular}{c@{\hskip 3pt}r@{\hskip 7pt}r@{\hskip 3pt}r@{\hskip 3pt}r@{\hskip 3pt}r@{\hskip 3pt}rccccc@{\hskip 1pt}}
            \toprule
            \multirow{5}{*}{Dataset}
                             &
            \multirow{5}{*}{\begin{tabular}[c]{@{}c@{}}Avg.\\Length\\($\text{m}$)\end{tabular}}
                             &
            \multirow{5}{*}{\begin{tabular}[c]{@{}c@{}}Avg.\\Time\\($\text{s}$)\end{tabular}}
                             &
            \multirow{5}{*}{\begin{tabular}[c]{@{}c@{}}Avg.\\Speed\\($\text{cm/s}$)\end{tabular}}
                             &
            \multirow{5}{*}{\begin{tabular}[c]{@{}c@{}}Avg.\\Angular\\Speed\\($\degree/\text{s}$)\end{tabular}}
                             &
            \multirow{5}{*}{\begin{tabular}[c]{@{}c@{}}Avg.\\Relative\\Speed\\($\text{cm/s}$)\end{tabular}}
                             &
            \multirow{5}{*}{\begin{tabular}[c]{@{}c@{}}Avg.\\Relative\\Angular\\Speed\\($\degree/\text{s}$)\end{tabular}}
                             &
            \multirow{3}{*}{Xun's\cite{xun2023crepes}}
                             &
            \multirow{3}{*}{SFC}
                             &
            \multirow{3}{*}{SFO}
                             &
            \multirow{3}{*}{MFLO}
                             &
            \multirow{3}{*}{MFTO}
            \\
            \\
            \\
            \cmidrule(r){8-8}
            \cmidrule(r){9-9}
            \cmidrule(r){10-10}
            \cmidrule(r){11-11}
            \cmidrule(r){12-12}
                             &
                             &
                             &
                             &
                             &
                             &
                             &
            Pos / Rot
                             &
            Pos / Rot
                             &
            Pos / Rot
                             &
            Pos / Rot
                             &
            Pos / Rot
            \\
                             &
                             &
                             &
                             &
                             &
                             &
                             &
            ($\mathrm{cm}$) / ($\degree$)
                             &
            ($\mathrm{cm}$) / ($\degree$)
                             &
            ($\mathrm{cm}$) / ($\degree$)
                             &
            ($\mathrm{cm}$) / ($\degree$)
                             &
            ($\mathrm{cm}$) / ($\degree$)
            \\
            \midrule
            \texttt{LOS\_1}  & 20.5 & 61.6 & 33.2  & 16.2  & 99.7   & 24.2  & \textbf{6.4 (5.7) / 2.1 (1.8)} & 34.6 (23.9) / 2.6 (2.3)  & 7.1 (5.8) / 2.1 (1.8)           & 7.8 (6.9) / 1.9 (1.7)   & 6.5 (5.6) / 2.0 (1.8)            \\
            \texttt{LOS\_2}  & 19.6 & 36.7 & 53.5  & 19.8  & 120.2  & 29.5  & 8.6 (7.3) / 2.4 (2.2)          & 36.3 (24.4) / 3.0 (2.6)  & 8.7 (6.4) / 2.3 (2.0)           & 8.5 (7.0) / 2.1 (1.8)   & \textbf{8.0 (6.4) / 2.2 (1.9)}   \\
            \texttt{LOS\_3}  & 19.6 & 35.1 & 55.9  & 20.6  & 98.0   & 26.6  & 8.5 (7.2) / 2.8 (2.1)          & 36.3 (24.1) / 3.3 (2.6)  & 8.6 (6.4) / 2.5 (2.0)           & 7.3 (6.4) / 2.2 (1.8)   & \textbf{7.1 (6.0) / 2.2 (1.8)}   \\
            \texttt{LOS\_4}  & 20.8 & 21.7 & 96.1  & 25.4  & 100.9  & 31.4  & \textbf{7.2 (6.2) / 2.1 (1.7)} & 36.8 (26.2) / 2.8 (2.2)  & 8.7 (6.4) / 2.6 (2.1)           & 8.3 (6.9) / 2.5 (2.0)   & 8.1 (6.7) / 2.6 (2.1)            \\
            \texttt{NLOS\_1} & 21.3 & 54.9 & 38.9  & 14.9  & 50.8   & 16.9  & 6.5 (5.4) / 2.2 (1.9)          & 38.7 (23.7) / 2.7 (2.2)  & 6.2 (4.6) / 2.1 (1.7)           & 6.1 (4.9) / 1.8 (1.5)   & \textbf{5.3 (4.3) / 1.9 (1.6)}   \\
            \texttt{NLOS\_2} & 21.6 & 34.8 & 62.0  & 19.0  & 64.3   & 21.5  & 8.4 (7.2) / 2.4 (2.2)          & 35.1 (21.0) / 3.0 (2.5)  & 8.0 (6.0) / 2.4 (2.0)           & 7.1 (5.8) / 2.1 (1.9)   & \textbf{7.0 (6.2) / 2.1 (1.8)}   \\
            \texttt{NLOS\_3} & 21.2 & 32.6 & 65.2  & 20.6  & 77.0   & 24.5  & 6.8 (5.8) / 2.3 (2.0)          & 33.3 (20.7) / 2.7 (2.3)  & 7.1 (5.4) / 2.2 (1.9)           & 6.5 (5.1) / 2.1 (1.8)   & \textbf{6.3 (5.2) / 2.1 (1.8)}   \\
            \texttt{NLOS\_4} & 21.1 & 23.2 & 91.0  & 24.2  & 83.5   & 27.0  & 8.0 (6.9) / 2.5 (2.2)          & 38.1 (25.0) / 2.9 (2.6)  & 8.0 (6.0) / 2.4 (2.1)           & 7.6 (6.3) / 2.2 (1.8)   & \textbf{7.5 (6.3) / 2.2 (1.9)}   \\
            \texttt{HDM\_1}  & 56.2 & 26.0 & 216.4 & 186.8 & 1000.6 & 213.4 & 25.9 (22.9) / 5.5 (4.7)        & 51.6 (24.6) / 10.1 (3.0) & 40.5 (11.3) / 9.6 (2.1)         & 28.3 (18.8) / 3.1 (2.7) & \textbf{21.8 (17.0) / 3.0 (2.6)} \\
            \texttt{HDM\_2}  & 35.4 & 25.0 & 141.8 & 94.8  & 814.7  & 118.2 & 20.2 (16.5) / 3.4 (3.1)        & 68.4 (22.3) / 8.1 (2.8)  & 59.8 (6.4) / 8.2 (2.1)          & 22.7 (17.7) / 2.7 (2.6) & \textbf{15.6 (12.6) / 2.6 (2.4)} \\
            \texttt{HDM\_3}  & 29.0 & 19.0 & 153.0 & 116.5 & 796.2  & 150.4 & 43.7 (19.4) / 3.9 (3.5)        & 36.4 (21.7) / 3.1 (2.6)  & \textbf{12.1 (8.6) / 2.6 (2.1)} & 20.0 (14.4) / 3.2 (2.9) & 14.2 (12.3) / 3.6 (3.2)          \\
            \texttt{HDM\_4}  & 44.1 & 29.0 & 152.3 & 108.9 & 735.1  & 135.2 & 22.5 (16.3) / 6.6 (6.2)        & 54.6 (24.5) / 9.2 (6.5)  & 38.8 (7.9) / 7.5 (3.5)          & 38.7 (16.9) / 3.5 (3.0) & \textbf{16.0 (12.7) / 3.2 (2.9)} \\
            \texttt{MRP\_1}  & 58.1 & 65.1 & 89.3  & 21.4  & 78.5   & 13.9  & 21.7 (17.1) / 4.9 (4.2)        & 40.3 (26.9) / 6.4 (4.8)  & 21.9 (10.2) / 6.9 (3.2)         & 16.0 (13.2) / 3.9 (2.7) & \textbf{10.7 (9.2) / 3.6 (2.7)}  \\
            \texttt{MRP\_2}  & 71.2 & 66.3 & 107.4 & 22.3  & 88.2   & 14.6  & 23.1 (18.4) / 4.8 (4.7)        & 44.9 (26.5) / 6.5 (5.0)  & 23.9 (9.7) / 8.3 (3.1)          & 16.1 (13.3) / 3.5 (2.5) & \textbf{10.5 (8.9) / 3.3 (2.6)}  \\
            \texttt{MRP\_3}  & 66.5 & 66.7 & 99.6  & 23.4  & 94.2   & 15.4  & 48.3 (26.7) / 4.7 (4.2)        & 80.0 (39.0) / 13.1 (5.0) & 60.8 (13.9) / 15.0 (4.3)        & 32.2 (19.8) / 3.4 (2.5) & \textbf{16.3 (13.1) / 3.1 (2.5)} \\
            \texttt{MRP\_4}  & 69.2 & 66.7 & 103.7 & 24.8  & 95.1   & 16.5  & 44.1 (25.2) / 5.3 (4.8)        & 84.7 (40.3) / 15.0 (5.4) & 66.7 (14.5) / 17.9 (4.6)        & 29.0 (19.9) / 4.8 (2.4) & \textbf{16.1 (12.0) / 3.2 (2.5)} \\
            \bottomrule
        \end{tabular}
    }
\end{table*}

\subsubsection{Accuracy under LOS and NLOS Conditions}
\label{Sec: Accuracy under LOS and NLOS Conditions}

In the 5-device \textbf{\texttt{LOS}}/\textbf{\texttt{NLOS}} datasets, the error generally decreases from SFC to SFO, MFLO, and MFTO, demonstrating the benefit of the layered estimation pipeline.
In the \textbf{\texttt{LOS}} datasets, where the observation constraints are sufficient, MFTO and Xun's method achieve comparable accuracy.
In the \textbf{\texttt{NLOS}} datasets, the advantage of MFTO becomes clear.
MFTO reduces the average error from $6.5-8.4\,\mathrm{cm}$ to $5.3-7.5\,\mathrm{cm}$.
This indicates that the tightly-coupled estimator better exploits the remaining inter-device constraints under degraded visibility conditions.

\autoref{Tab: RMSE of CREPES MFRE in NLOS Datasets} further separates the results into unoccluded and occluded intervals.
For MFTO, the error increases from $6.0\,\mathrm{cm}/2.0^\circ$ to $7.4\,\mathrm{cm}/2.1^\circ$ under occlusion.
By contrast, MFLO increases from $6.1\,\mathrm{cm}/2.0^\circ$ to $8.4\,\mathrm{cm}/2.1^\circ$, and Xun's method increases from $6.8\,\mathrm{cm}/2.3^\circ$ to $8.6\,\mathrm{cm}/2.4^\circ$.
Thus, MFTO achieves the lowest occluded error and a smaller relative increase in position RMSE than MFLO ($23\%$ vs. $38\%$), confirming its stronger robustness to occlusion.

The \texttt{NLOS\_2} example in \autoref{Fig: Exp: Real-World NLOS Results} shows the same trend.
MFTO does not exhibit a pronounced error increase during occluded intervals.
During occlusion, MFTO remains at $7.9\,\mathrm{cm}/1.9^\circ$, whereas MFLO and Xun's method increase to $9.3\,\mathrm{cm}/1.9^\circ$ and $9.6\,\mathrm{cm}/2.3^\circ$, respectively.
Overall, \CREPES{} keeps the occluded MFTO error below $9\,\mathrm{cm}$ and close to $2^\circ$ in the 5-device NLOS datasets.

\begin{table}[t]
    \centering
    \caption{
        RMSE of \CREPES{} MFRE in NLOS datasets.
    }
    \label{Tab: RMSE of CREPES MFRE in NLOS Datasets}
    \resizebox{0.9\halfwidth}{!}{
        \begin{tabular}{@{\hskip 1pt}c@{\hskip 5pt}c@{\hskip 5pt}c@{\hskip 5pt}c@{\hskip 5pt}c@{\hskip 5pt}c@{\hskip 5pt}c@{\hskip 1pt}}
            \toprule
            \multirow{4}{*}{Dataset}
                             &
            \multicolumn{3}{c}{Unoccluded Error}
                             &
            \multicolumn{3}{c}{Occluded Error}
            \\
            \cmidrule(r){2-4}\cmidrule{5-7}
                             &
            Xun's\cite{xun2023crepes}
                             &
            MFLO
                             &
            MFTO
                             &
            Xun's\cite{xun2023crepes}
                             &
            MFLO
                             &
            MFTO
            \\
            \cmidrule(r){2-2}
            \cmidrule(r){3-3}
            \cmidrule(r){4-4}
            \cmidrule(r){5-5}
            \cmidrule(r){6-6}
            \cmidrule{7-7}
                             &
            Pos / Rot
                             &
            Pos / Rot
                             &
            Pos / Rot
                             &
            Pos / Rot
                             &
            Pos / Rot
                             &
            Pos / Rot
            \\
                             &
            ($\mathrm{cm}$) / ($\degree$)
                             &
            ($\mathrm{cm}$) / ($\degree$)
                             &
            ($\mathrm{cm}$) / ($\degree$)
                             &
            ($\mathrm{cm}$) / ($\degree$)
                             &
            ($\mathrm{cm}$) / ($\degree$)
                             &
            ($\mathrm{cm}$) / ($\degree$)
            \\
            \midrule
            \texttt{NLOS\_1} & 6.0 / 2.1 & 5.6 / 1.7 & \textbf{4.8 / 1.8} & 8.1 / 2.3 & 7.6 / 2.0 & \textbf{6.6 / 2.0} \\
            \texttt{NLOS\_2} & 7.9 / 2.4 & \textbf{6.3 / 2.2} & 6.6 / 2.1 & 9.6 / 2.3 & 9.3 / 1.9 & \textbf{7.9 / 1.9} \\
            \texttt{NLOS\_3} & 6.3 / 2.2 & \textbf{5.9 / 2.1} & 6.0 / 2.1 & 8.2 / 2.6 & 8.1 / 2.1 & \textbf{7.2 / 2.1} \\
            \texttt{NLOS\_4} & 7.8 / 2.4 & \textbf{7.1 / 2.1} & 7.2 / 2.1 & 8.9 / 2.5 & 9.4 / 2.4 & \textbf{8.7 / 2.4} \\
            \midrule
            Overall          & 6.8 / 2.3 & 6.1 / 2.0 & \textbf{6.0 / 2.0} & 8.6 / 2.4 & 8.4 / 2.1 & \textbf{7.4 / 2.1} \\
\bottomrule
        \end{tabular}
    }
\end{table}

\subsubsection{Accuracy under Dynamic Motion}
\label{Sec: Accuracy under Dynamic Motion}

\texttt{HDM\_1} and \texttt{HDM\_2} use circular and figure-eight motions under LOS, respectively.
\texttt{HDM\_3} and \texttt{HDM\_4} repeat the same two motion patterns under NLOS.
The results are shown in \autoref{Fig: Exp: Real-World HDM Results} and \autoref{Tab: RMSE of CREPES in Real-World Datasets}, where the values in parentheses denote the median errors.

The four \textbf{\texttt{HDM}} sequences involve substantially stronger dynamics than the previous datasets, with average relative speeds of $735.1$--$1000.6\,\mathrm{cm/s}$ and average relative angular speeds of $118.2$--$213.4^\circ/\mathrm{s}$.
Despite these aggressive motions, MFTO remains the most accurate or competitive method across the four sequences.
In \texttt{HDM\_1}, where the average relative speed reaches $1000.6\,\mathrm{cm/s}$ and the average relative angular speed reaches $213.4^\circ/\mathrm{s}$, MFTO achieves $21.8\,(17.0)\,\mathrm{cm}$ in position and $3.0^\circ\,(2.6^\circ)$ in rotation.

The median errors further show that MFTO maintains stable frame-level accuracy under high dynamics.
For example, MFTO obtains median position errors of $17.0$, $12.6$, $12.3$, and $12.7\,\mathrm{cm}$ on \texttt{HDM\_1}--\texttt{HDM\_4}, respectively.
Several single-frame results exhibit a much larger gap between RMSE and median error, such as SFO on \texttt{HDM\_2} with $59.8\,(6.4)\,\mathrm{cm}$ and SFO on \texttt{HDM\_4} with $38.8\,(7.9)\,\mathrm{cm}$.
This indicates that SFC and SFO are often accurate on well-constrained frames, but can produce occasional outliers under high-speed or degraded observations.

We also observe that SFO can be lower than MFLO or MFTO in some cases, such as \texttt{HDM\_3}.
However, SFC and SFO only output estimates when the single-frame observations are sufficiently constrained, whereas MFLO and MFTO continue estimating through less favorable intervals.
Therefore, the slightly higher RMSE of MFLO or MFTO in such cases should be interpreted together with their more continuous sequence-level estimation.

\subsubsection{Accuracy on Mobile-Robot Platforms}
\label{Sec: Accuracy on Mobile-Robot Platforms}

\begin{figure*}[p]

    \begin{minipage}{\fullwidth}
        \centering
        \includegraphics[width=\fullwidth]{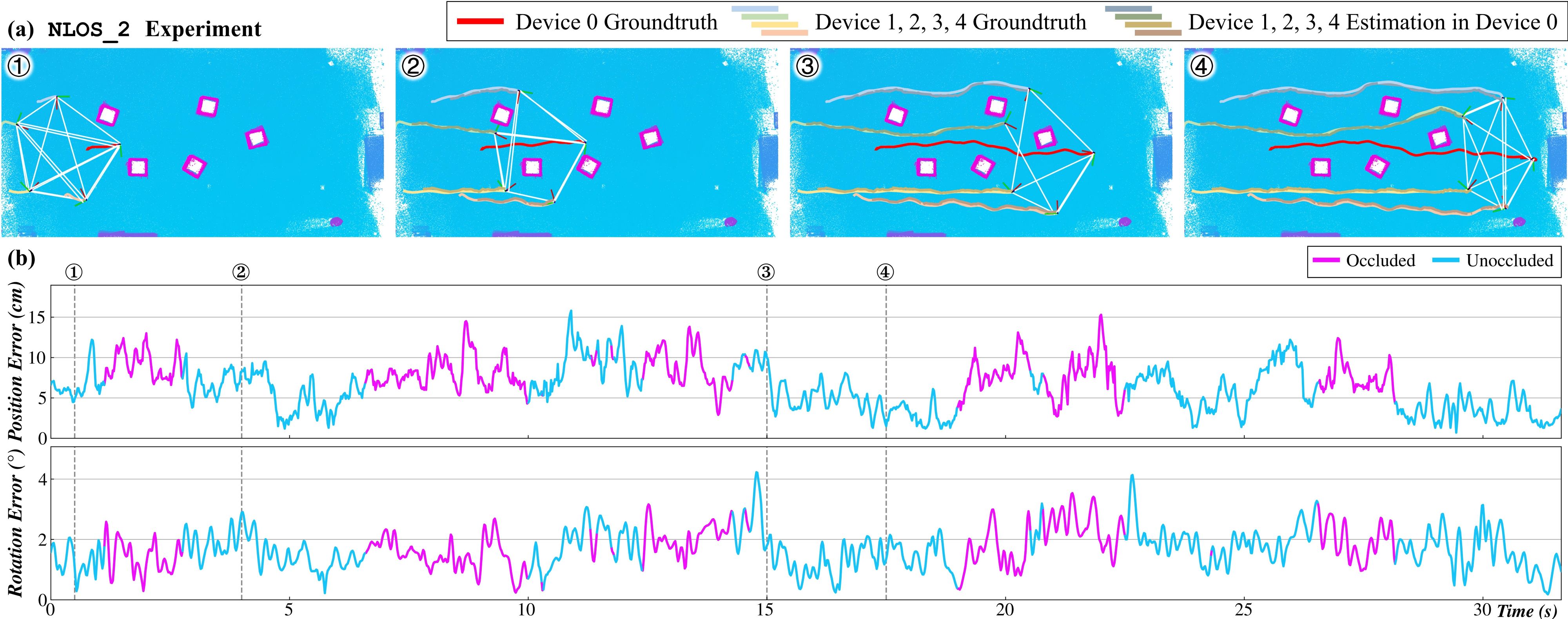}
        \caption{
            Experiment of five moving devices in \texttt{NLOS\_2}, chosen as the median-performance NLOS dataset.
            Blue and purple regions denote the ground and obstacles, respectively.
            (a) The trajectory of five devices and their visibility (white connection).
            The circled markers \ding{172}--\ding{175} indicate four representative trajectory snapshots, which correspond to the same markers in (b).
            (b) The position error and rotation error of Device 1 (reference frame is Device 0).
            The error when occluded is nearly the same as that in unoccluded conditions.
        }
        \label{Fig: Exp: Real-World NLOS Results}
    \end{minipage}

    \begin{minipage}{\fullwidth}
        \centering
        \includegraphics[width=\fullwidth]{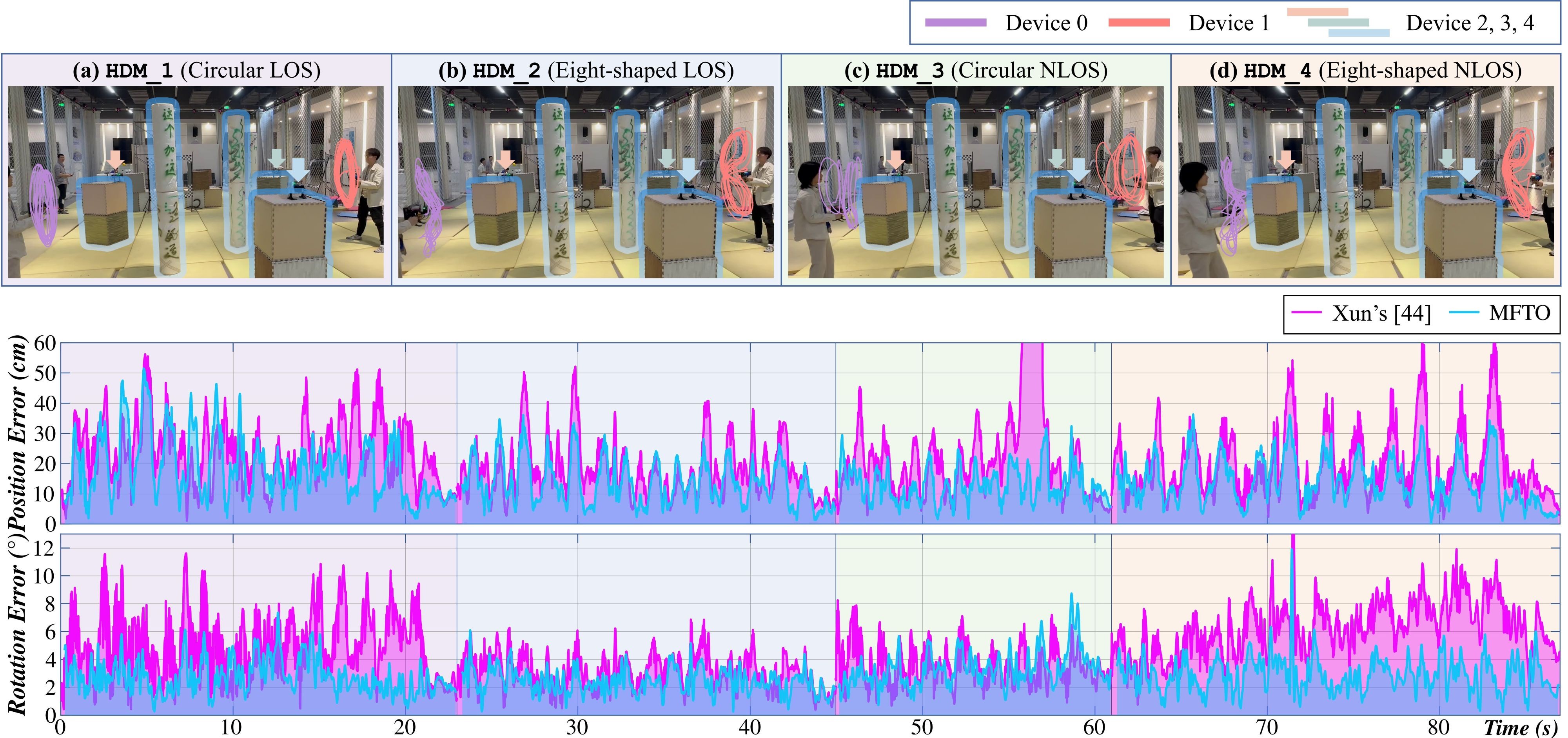}
        \caption{
            High dynamic motion (\textbf{\texttt{HDM}}) results.
            Upper: Trajectories in the world frame, with each recorded segment shown in a distinct color.
            Lower: Position and rotation error of Device~1, expressed in the reference frame of Device~0.
            The background shading along the time axis delineates the same shading dataset segments as in the upper panel.
        }
        \label{Fig: Exp: Real-World HDM Results}
    \end{minipage}

    \begin{minipage}{\fullwidth}
        \centering
        \includegraphics[width=\fullwidth]{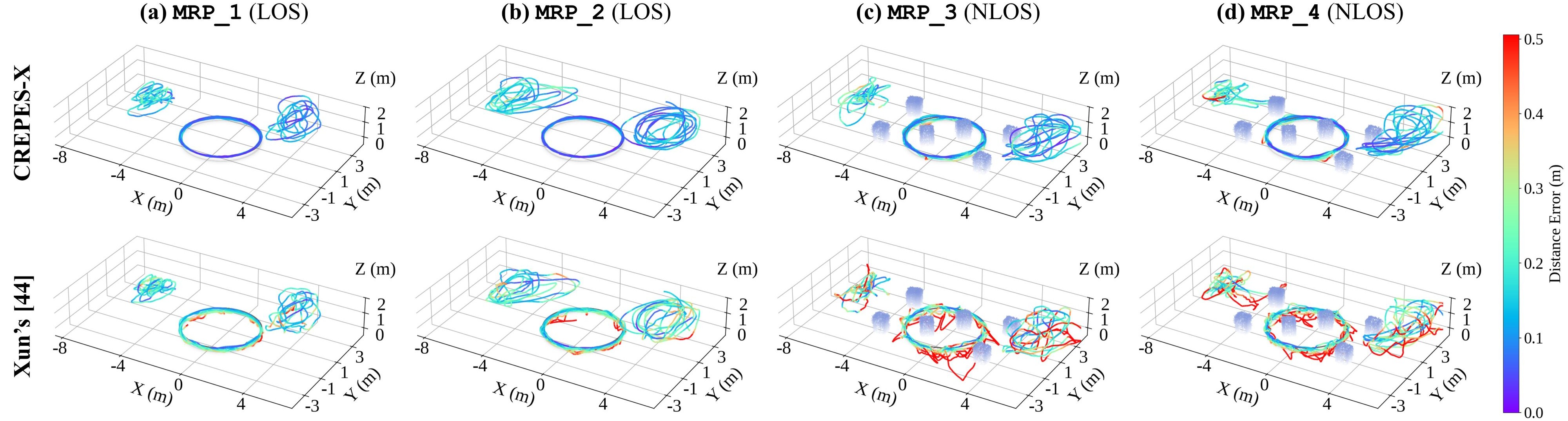}
        \caption{
            Mobile-robot platform (\textbf{\texttt{MRP}}) results transformed into world frame.
            Upper: Estimated trajectories from MFTO.
            Lower: Estimated trajectories from Xun's~\cite{xun2023crepes}.
            The trajectories are colored by position error.
            MFTO achieves lower position errors than Xun's baseline, especially under NLOS conditions.
        }
        \label{Fig: Exp: Real-World UGV Results}
    \end{minipage}

\end{figure*}

Device~0 is mounted on a UGV and serves as the reference frame.
The three UGVs autonomously track a pre-generated circular trajectory with a radius of $2\,\mathrm{m}$ from three equally spaced starting positions using model predictive control.
Meanwhile, two human operators manually fly the UAVs in front of and behind the UGV formation.
\texttt{MRP\_1} and \texttt{MRP\_2} are collected under LOS, while \texttt{MRP\_3} and \texttt{MRP\_4} are collected under NLOS with obstacles placed in the operating area, as shown in \autoref{Fig: Exp: Real-World Environment 2}(a).
The results are reported in \autoref{Fig: Exp: Real-World UGV Results} and \autoref{Tab: RMSE of CREPES in Real-World Datasets}.

MFTO achieves the best accuracy on all \textbf{\texttt{MRP}} sequences.
On the LOS sequences, MFTO obtains $10.7\,(9.2)\,\mathrm{cm}/3.6^\circ\,(2.7^\circ)$ on \texttt{MRP\_1} and $10.5\,(8.9)\,\mathrm{cm}/3.3^\circ\,(2.6^\circ)$ on \texttt{MRP\_2}.
Compared with Xun's baseline~\cite{xun2023crepes}, this reduces the position RMSE by $51\%$ and $55\%$, respectively.
Under NLOS, MFTO maintains position errors of $16.3\,(13.1)\,\mathrm{cm}$ on \texttt{MRP\_3} and $16.1\,(12.0)\,\mathrm{cm}$ on \texttt{MRP\_4}, while Xun's baseline increases to $48.3\,(26.7)\,\mathrm{cm}$ and $44.1\,(25.2)\,\mathrm{cm}$.

The median errors provide a complementary view of the results.
MFTO maintains median position errors of $9.2$, $8.9$, $13.1$, and $12.0\,\mathrm{cm}$ on \texttt{MRP\_1}--\texttt{MRP\_4}, respectively.
Similar to the \textbf{\texttt{HDM}} sequences, the large RMSE--median gaps of some single-frame results, such as SFO on \texttt{MRP\_3} and \texttt{MRP\_4}, indicate occasional outliers under degraded observations.

Compared with MFLO, MFTO also provides a clear advantage under occlusion.
On \texttt{MRP\_3} and \texttt{MRP\_4}, MFTO reduces the position RMSE from $32.2\,\mathrm{cm}$ and $29.0\,\mathrm{cm}$ to $16.3\,\mathrm{cm}$ and $16.1\,\mathrm{cm}$, respectively.
This improvement supports the benefit of the tightly coupled formulation, which better uses the remaining distance, bearing, and inertial constraints when visual observations are partially unavailable.
Overall, across the multi-robot UAV/UGV deployments, MFTO achieves $10.5$--$16.3\,\mathrm{cm}$ position RMSE and $3.1^\circ$--$3.6^\circ$ rotation RMSE under both LOS and NLOS conditions.

\subsubsection{Impact of Number of Devices}
\label{Sec: Impact of Number of Devices}

We vary the number of devices from $2$ to $5$ and report the aggregated results in \autoref{Tab: RMSE of CREPES in Real-World Datasets n}.
With only two devices, position errors increase under \textbf{\texttt{NLOS}} because occlusion leaves few redundant inter-device constraints.
This is a structural limitation of the two-device setting and is also reflected in Xun's formulation~\cite{xun2023crepes}.
In this regime, Xun's filter uses every measurement, whereas MFTO only consumes keyframe observations and therefore operates on a sparser measurement set.
This denser measurement budget allows Xun's filter to slightly outperform MFTO in the 2-device \textbf{\texttt{LOS}} case, with $8.4\,\mathrm{cm}$ versus $9.0\,\mathrm{cm}$.

\begin{table}[t]
    \centering
    \caption{
        RMSE of \CREPES{} in Real-World Datasets with Different Numbers of Devices.
        (The values in parentheses denote the median errors.)
        \textit{failed} marks runs in which Xun's filter diverged before the sequence completed.
    }
    \label{Tab: RMSE of CREPES in Real-World Datasets n}
    \resizebox{0.8\halfwidth}{!}{
        \begin{tabular}{@{\hskip 1pt}c@{\hskip 5pt}c@{\hskip 1pt}c@{\hskip 3pt}c@{\hskip 3pt}c@{\hskip 3pt}c@{\hskip 3pt}c@{\hskip 1pt}}
            \toprule
            \multirow{3}{*}{\rotatebox[origin=c]{90}{Dataset}}
             &
            \multirow{3}{*}{\begin{tabular}[c]{@{}c@{}}Num.\\ Dev.\end{tabular}}
             &
            Xun's \cite{xun2023crepes}
             &
            SFC
             &
            SFO
             &
            MFLO
             &
            MFTO
            \\
            \cmidrule(r){3-3}
            \cmidrule(r){4-4}
            \cmidrule(r){5-5}
            \cmidrule(r){6-6}
            \cmidrule(r){7-7}
             &
             &
            Pos
             &
            Pos
             &
            Pos
             &
            Pos
             &
            Pos
            \\
             &
             &
            ($\mathrm{cm}$)
             &
            ($\mathrm{cm}$)
             &
            ($\mathrm{cm}$)
             &
            ($\mathrm{cm}$)
             &
            ($\mathrm{cm}$)
            \\
            \midrule
            \multirow{4}{*}{\rotatebox[origin=c]{90}{\textbf{\texttt{LOS}}}}
             & 2 & 8.4 (7.1)   & 8.4 (7.1)   & \textbf{8.3 (6.9)}   & 48.7 (10.4)  & 9.0 (7.4)            \\
             & 3 & 9.1 (7.8)   & 10.2 (8.5)  & \textbf{8.9 (7.3)}   & 10.1 (9.3)   & 9.1 (7.8)            \\
             & 4 & 8.0 (6.8)   & 17.9 (11.2) & 8.5 (6.8)            & 8.6 (7.6)    & \textbf{7.8 (6.6)}   \\
             & 5 & 7.5 (6.4)   & 35.7 (24.3) & 8.1 (6.2)            & 7.9 (6.8)    & \textbf{7.2 (6.0)}   \\
            \midrule
            \multirow{4}{*}{\rotatebox[origin=c]{90}{\textbf{\texttt{NLOS}}}}
             & 2 & 48.5 (9.2)  & 8.8 (7.8)   & \textbf{7.7 (6.2)}   & 41.3 (10.4)  & 13.7 (7.6)           \\
             & 3 & 9.9 (6.4)   & 8.7 (7.3)   & \textbf{6.7 (5.3)}   & 39.2 (6.7)   & 7.2 (5.7)            \\
             & 4 & 8.4 (6.3)   & 21.0 (10.3) & 8.0 (5.6)            & 7.4 (5.7)    & \textbf{6.8 (5.4)}   \\
             & 5 & 7.3 (6.1)   & 36.7 (22.5) & 7.2 (5.3)            & 6.7 (5.4)    & \textbf{6.3 (5.1)}   \\
            \midrule
            \multirow{4}{*}{\rotatebox[origin=c]{90}{\textbf{\texttt{HDM}}}}
             & 2 & failed      & 38.5 (12.3) & 37.2 (8.0)           & 52.3 (21.9)  & \textbf{28.0 (17.3)} \\
             & 3 & failed      & 25.5 (13.6) & \textbf{20.7 (8.4)}  & 106.4 (21.3) & 20.9 (16.3)          \\
             & 4 & failed      & 43.4 (18.5) & 30.5 (8.8)           & 30.0 (19.0)  & \textbf{20.0 (14.8)} \\
             & 5 & 28.1 (18.5) & 55.2 (23.4) & 42.9 (8.0)           & 29.4 (17.2)  & \textbf{17.3 (13.3)} \\
            \midrule
            \multirow{4}{*}{\rotatebox[origin=c]{90}{\textbf{\texttt{MRP}}}}
             & 2 & failed      & 21.0 (14.6) & \textbf{17.0 (12.5)} & 112.2 (20.4) & 28.7 (11.8)          \\
             & 3 & 36.6 (19.5) & 49.3 (22.1) & 29.6 (12.7)          & 35.5 (18.5)  & \textbf{16.1 (12.4)} \\
             & 4 & 40.7 (20.5) & 78.7 (39.3) & 56.2 (12.8)          & 38.5 (18.9)  & \textbf{22.1 (13.6)} \\
             & 5 & 31.7 (18.8) & 61.9 (30.9) & 44.3 (11.5)          & 24.4 (15.9)  & \textbf{13.7 (10.5)} \\
            \bottomrule
        \end{tabular}
    }
\end{table}

With three or more devices, the additional inter-device constraints largely compensate for MFTO's keyframe sparsity.
The advantage is more evident in challenging \textbf{\texttt{HDM}} and \textbf{\texttt{MRP}} sequences, where Xun's filter often produces much larger errors.
Furthermore, Xun's filter sometimes diverges before the sequence is completed.
For example, it fails in the 3--4-device \textbf{\texttt{HDM}} cases, whereas MFTO remains stable.

We also note that the SFC error does not necessarily decrease monotonically with the number of devices.
SFC solves an MDS-type geometric reconstruction problem, whose accuracy depends on both the number of pairwise constraints and their geometry and consistency.
Thus, a mild RMSE increase with more devices mainly reflects the geometric nature of the underlying MDS problem, rather than a failure of the SFC formulation.

For MFLO, the occasional large RMSE values should be interpreted together with the corresponding median errors.
The relatively small medians indicate that these large RMSE values are mainly caused by a few short unfavorable intervals rather than persistent errors over the entire sequence.
In these intervals, the single-frame observations are weakly constrained, so SFC and SFO cannot provide estimates, whereas MFLO continues to propagate and estimate through the sequence.
As a result, MFLO can accumulate temporary drift, but it still provides more continuous sequence-level estimation.

\subsubsection{Ablation Study}
\label{Sec: Ablation Study}

To evaluate the contribution of each component and validate the design of the proposed multilayer estimation architecture, we conduct an ablation study by removing SFC, SFO, and MFLO and replacing their outputs with random values.
The results are summarized in \autoref{Tab: Exp: Ablation Study Results}.

Row~1 uses MFTO alone from random initialization, achieving $7.5\,\mathrm{cm}$ / $2.3^{\circ}$ with a first optimization runtime of $44.3\,\mathrm{ms}$.
Rows~2--4 show that no single front-end module is sufficient by itself.
SFC alone slightly improves the final MFTO result to $6.9\,\mathrm{cm}$ / $2.1^{\circ}$, while SFO alone converges to the wrong chirality and MFLO alone cannot output a valid estimate.

Rows~5--8 further show the role of each layer.
SFC provides the chirality prior required by SFO.
Without SFC, SFO fails with $240.6\,\mathrm{cm}$ / $66.4^{\circ}$ error, and this incorrect estimate also corrupts MFLO, whose error grows to $349.8\,\mathrm{cm}$ / $5.2^{\circ}$.
SFO provides the refined single-frame estimate needed by MFLO.
Without SFO, MFLO must operate directly on the raw SFC output, and its position error increases from $7.1\,\mathrm{cm}$ to $22.2\,\mathrm{cm}$.

MFLO has two roles in the architecture.
First, it is an independent lightweight multi-frame estimator; in the full system, it already achieves $7.1\,\mathrm{cm}$ / $2.1^{\circ}$, close to the final MFTO accuracy.
Second, it provides an efficient warm start for MFTO.
Removing MFLO leaves the rounded final MFTO accuracy almost unchanged at $7.0\,\mathrm{cm}$ / $2.1^{\circ}$, but increases the first MFTO runtime from $17.9\,\mathrm{ms}$ to $45.4\,\mathrm{ms}$.
Thus, MFLO reduces the first optimization time by $2.5\times$ with negligible accuracy loss.

Overall, the ablation validates the multilayer design.
SFC resolves the chirality ambiguity, SFO refines the single-frame estimate, and MFLO provides both a lightweight standalone estimate and a fast initialization for MFTO.

\begin{table}[t]
    \centering
    \caption{Ablation Study Results}
    \label{Tab: Exp: Ablation Study Results}
    \resizebox{0.8\halfwidth}{!}{
        \begin{tabular}{@{\hskip 1pt}c@{\hskip 5pt}c@{\hskip 5pt}c@{\hskip 5pt}c@{\hskip 5pt}c@{\hskip 5pt}c@{\hskip 5pt}c@{\hskip 5pt}c@{\hskip 3pt}c@{\hskip 1pt}}
            \toprule
            \multirow{3}{*}{\#}                             &
            \multirow{3}{*}{\rotatebox[origin=c]{90}{SFC}}  &
            \multirow{3}{*}{\rotatebox[origin=c]{90}{SFO}}  &
            \multirow{3}{*}{\rotatebox[origin=c]{90}{MFLO}} &
            SFC                                             &
            SFO                                             &
            MFLO                                            &
            \multicolumn{2}{c}{MFTO}
            \\
            \cmidrule(r){5-5}
            \cmidrule(r){6-6}
            \cmidrule(r){7-7}
            \cmidrule{8-9}
                                                            &
                                                            &
                                                            &
                                                            &
            Pos / Rot
                                                            &
            Pos / Rot
                                                            &
            Pos / Rot
                                                            &
            Pos / Rot
                                                            &
            First\;Runtime
            \\
                                                            &
                                                            &
                                                            &
                                                            &
            ($\mathrm{cm}$) / ($\degree$)
                                                            &
            ($\mathrm{cm}$) / ($\degree$)
                                                            &
            ($\mathrm{cm}$) / ($\degree$)
                                                            &
            ($\mathrm{cm}$) / ($\degree$)
                                                            &
            ($\mathrm{ms}$)
            \\
            \midrule
            1 &  &  &  & - & - & - & 7.5 / 2.3 & 44.3 \\
            2 & \ding{51} &  &  & 35.2 / 3.0 & - & - & \textbf{6.9 / 2.1} & 54.5 \\
            3 &  & \ding{51} &  & - & 240.6 / 66.4 & - & 7.5 / 2.3 & 42.1 \\
            4 &  &  & \ding{51} & - & - & No output & 7.5 / 2.3 & 44.3 \\
            5 &  & \ding{51} & \ding{51} & - & 240.6 / 66.4 & 349.8 / 5.2 & 7.5 / 2.3 & 42.1 \\
            6 & \ding{51} &  & \ding{51} & \textbf{35.1 / 3.0} & - & 22.2 / 2.1 & \textbf{6.9 / 2.1} & 21.3 \\
            7 & \ding{51} & \ding{51} &  & 35.2 / 3.0 & 8.1 / 2.4 & - & 7.0 / 2.1 & 45.4 \\
            \rowcolor{gray!15}[3pt][3pt]
            8 & \ding{51} & \ding{51} & \ding{51} & \textbf{35.1 / 3.0} & \textbf{8.0 / 2.4} & \textbf{7.1 / 2.1} & 7.0 / 2.1 & \textbf{17.9} \\
\bottomrule
        \end{tabular}
    }
\end{table}

\subsection{Evaluation in Non-inertial Environments}
\label{Sec: Evaluation in Non-inertial Environments}

A non-inertial environment introduces apparent linear acceleration that is indistinguishable from gravity.
For simplicity, we use \textbf{gravity} to denote both true gravity and equivalent linear-acceleration effects.
We evaluate three cases: A) no gravity, B) constant gravity, and C) time-varying gravity in both direction and magnitude.

In all simulations, prior gravity information is disabled by setting $w_g=0$.
For case C, we generate a random B-spline trajectory and define gravity at each timestamp to point toward the sampled position, with magnitude $9.8(d/5)$, where $d$ is the distance to the origin.
The number of devices and the window size are set to $10$ and $10$, respectively.
As shown in \autoref{Tab: RMSE of CREPES in Datasets g} (\texttt{Bench\_A}, \texttt{Bench\_B}, and \texttt{Bench\_C}), \CREPES{} maintains high relative-state estimation accuracy under all three non-inertial settings.

\begin{table}[t]
    \centering
    \caption{
        RMSE of \CREPES{} in Datasets ($w_g = 0$)
    }
    \label{Tab: RMSE of CREPES in Datasets g}
    \resizebox{.93\halfwidth}{!}{
        \begin{tabular}{@{\hskip 1pt}c@{\hskip 3pt}c@{\hskip 5pt}c@{\hskip 5pt}c@{\hskip 5pt}c@{\hskip 5pt}c@{\hskip 5pt}c@{\hskip 1pt}}
            \toprule
             &
            \multirow{3}{*}{Dataset}
             &
            Xun's \cite{xun2023crepes}
             &
            SFC
             &
            SFO
             &
            MFLO
             &
            MFTO
            \\
            \cmidrule(r){3-3}\cmidrule(r){4-4}\cmidrule(r){5-5}\cmidrule(r){6-6}\cmidrule{7-7}
             &
             &
            Pos / Rot
             &
            Pos / Rot
             &
            Pos / Rot
             &
            Pos / Rot
             &
            Pos / Rot
            \\
             &
             &
            ($\mathrm{cm}$) / ($\degree$)
             &
            ($\mathrm{cm}$) / ($\degree$)
             &
            ($\mathrm{cm}$) / ($\degree$)
             &
            ($\mathrm{cm}$) / ($\degree$)
             &
            ($\mathrm{cm}$) / ($\degree$)
            \\
            \midrule
            \multirow{6}{*}{\rotatebox[origin=c]{90}{Simulation}}
             & \texttt{Bench\_A} & failed & 25.4 / 4.5  & 7.4 / 1.5           & \textbf{2.0 / 0.3}  & 3.5 / 0.4          \\
             & \texttt{Bench\_B} & failed & 25.4 / 4.5  & 7.4 / 1.5           & \textbf{2.1 / 0.3}  & 3.7 / 0.4          \\
             & \texttt{Bench\_C} & failed & 25.4 / 4.5  & 7.4 / 1.5           & \textbf{2.0 / 0.3}  & 3.6 / 0.4          \\
             & \texttt{ISS\_ff}  & failed & 52.5 / 15.0 & 5.0 / 1.8           & \textbf{1.7 / 0.5}  & 2.3 / 0.7          \\
             & \texttt{ISS\_iva} & failed & 50.4 / 24.6 & 7.3 / 5.1           & 3.3 / 3.9           & \textbf{2.8 / 1.4} \\
             & \texttt{ISS\_td}  & failed & 15.2 / 7.6  & 5.9 / 3.7           & \textbf{2.2 / 36.7} & 2.5 / 0.9          \\
            \midrule
            \multirow{16}{*}{\rotatebox[origin=c]{90}{Real World}}
 & \texttt{LOS\_1} & failed & 34.6 / 12.3 & 7.8 / 2.5 & 8.3 / 2.1 & \textbf{6.6 / 2.1} \\
             & \texttt{LOS\_2} & failed & 34.7 / 12.0 & 8.7 / 2.4 & 9.1 / 2.2 & \textbf{8.0 / 2.3} \\
             & \texttt{LOS\_3} & failed & 37.7 / 13.1 & 7.9 / 2.6 & 8.0 / 2.3 & \textbf{7.0 / 2.3} \\
             & \texttt{LOS\_4} & failed & 34.9 / 12.7 & 9.0 / 2.9 & 9.2 / 2.6 & \textbf{8.1 / 2.6} \\
             & \texttt{NLOS\_1} & failed & 38.9 / 13.0 & 13.1 / 4.3 & 8.5 / 1.9 & \textbf{5.4 / 1.9} \\
             & \texttt{NLOS\_2} & failed & 36.6 / 12.8 & 12.5 / 4.1 & 8.8 / 2.2 & \textbf{7.0 / 2.1} \\
             & \texttt{NLOS\_3} & failed & 37.1 / 12.6 & 13.9 / 4.7 & 7.5 / 2.2 & \textbf{6.5 / 2.1} \\
             & \texttt{NLOS\_4} & failed & 41.6 / 13.5 & 10.2 / 3.2 & 8.4 / 2.2 & \textbf{7.6 / 2.2} \\
             & \texttt{HDM\_1} & failed & 37.9 / 15.7 & \textbf{10.8 / 2.8} & 28.7 / 3.5 & 24.0 / 3.0 \\
             & \texttt{HDM\_2} & failed & 27.3 / 14.3 & \textbf{8.2 / 2.2} & 28.0 / 2.7 & 16.9 / 2.6 \\
             & \texttt{HDM\_3} & failed & 38.8 / 12.8 & \textbf{9.4 / 2.8} & 20.1 / 3.2 & 13.9 / 3.8 \\
             & \texttt{HDM\_4} & failed & 27.9 / 12.8 & \textbf{7.0 / 2.4} & 22.6 / 3.0 & 16.0 / 2.8 \\
             & \texttt{MRP\_1} & failed & 39.0 / 13.7 & 21.8 / 12.3 & 16.7 / 3.9 & \textbf{10.6 / 3.6} \\
             & \texttt{MRP\_2} & failed & 44.9 / 15.4 & 24.1 / 13.7 & 17.5 / 3.5 & \textbf{10.4 / 3.3} \\
             & \texttt{MRP\_3} & failed & 141.9 / 32.5 & 134.4 / 31.7 & 75.1 / 3.8 & \textbf{15.5 / 3.1} \\
             & \texttt{MRP\_4} & failed & 151.6 / 39.8 & 145.9 / 39.2 & 82.5 / 3.7 & \textbf{15.8 / 3.2} \\
\bottomrule
        \end{tabular}
    }
\end{table}

We further test $w_g=0$ on real-world datasets, which correspond to the constant-gravity case.
Compared with \autoref{Tab: RMSE of CREPES in Real-World Datasets}, MFLO and MFTO show only mild degradation, indicating that inertial and relative measurements still provide sufficient constraints without a gravity prior.
The main exception is SFC rotation accuracy, which degrades significantly because the mostly planar device layout introduces chirality ambiguity when gravity is unavailable.

Finally, we evaluate \CREPES{} using the Astrobee dataset~\cite{kang2024astrobee}, collected aboard the International Space Station.
Since the dataset contains only one robot, we construct simulated multi-robot sequences by combining trajectory segments and generating bearing and distance measurements the same as in the benchmark.
The resulting datasets, \texttt{ISS\_ff}, \texttt{ISS\_iva}, and \texttt{ISS\_td}, each last $30$\,s.
The results in \autoref{Tab: RMSE of CREPES in Datasets g} show that \CREPES{} also achieves accurate estimation under real microgravity conditions.

We note that \CREPES{} has not been validated for operational use in space and assumes all devices share the same non-inertial field.
Therefore, it does not model large-scale gravity-field gradients.
Nevertheless, for local relative estimation, the method is hardware-agnostic and only requires bearing, distance, and inertial measurements, suggesting its potential applicability to space robotics.

\section{Conclusion}
\label{Sec: Conclusion}

In this paper, we present \CREPES{}, a complete and hierarchical hardware and software solution for cooperative relative localization that overcomes the \CHALLENGES{} in real-world multi-robot scenarios.
Active infrared LEDs eliminate environmental dependence, while a time-synchronized coding scheme associates bearing measurements with robot IDs.
By integrating multi-robot bearing, distance, and optional gravity measurements, \CREPES{} estimates instantaneous relative poses in a single frame with efficient closed-form solutions followed by optimization refinement.
It then performs loosely- and tightly-coupled optimization via IMU-based robocentric relative kinematics over multiple frames.

The hierarchical estimators provide four output streams, each tailored to a specific application need:
SFC: Instantaneous poses for large-scale swarms.
SFO: Refined instantaneous poses with higher accuracy.
MFLO: Accurate, smoothed real-time state for general use.
MFTO: Robust, accurate real-time state, designed to operate in observation-deficient scenarios.

Despite its advantages, \CREPES{} still has limitations.
First, bearing observations are confined to the IR camera's field of view, so peers are implicitly assumed to remain mutually visible; omnidirectional sensing, such as multi-camera rigs or panoramic IR optics, combined with estimation-aware planning could actively maintain the line-of-sight visibility graph.
Second, data association currently relies on cameras, infrared LEDs, and time-synchronized encoding, imposing hardware constraints; anonymous bearing-distance fusion, closed-loop ID feedback, and learning-based association are promising alternatives.
Third, the current real-hardware experiments do not fully cover extreme SE(3) configurations, and the potential increase of UWB ranging errors under such conditions remains insufficiently evaluated.

\section*{Acknowledgment}

The authors would like to thank
Chice Xuan,
Zhihao Tian,
and
Tienan Zhang
for their assistance in hardware design and implementation.
The authors would like to thank
Zhenjun Ying
and
Baozhe Zhang
for their assistance in image processing algorithms.
The authors would like to thank
Juncheng Chen,
Nanhe Chen,
Zhenyu Hou,
Mingwei Lai,
Tiancheng Lai,
Xiangyu Li,
Wentao Liu,
Ruitian Pang,
Pengfei Wang,
Shuo Wang,
Xingpeng Wang,
Ge Wan,
Wenkai Xiao,
Miao Xu,
Jiajun Yu,
Mengke Zhang,
and
Mingxuan Zhang
for their assistance in conducting real-world experiments.
The authors also thank
Kanyu Xu
for her assistance in video production.

\bibliographystyle{IEEEtran}
\bibliography{root}

\end{document}